\documentclass[a4paper,english,round]{article}
\usepackage[T1]{fontenc}
\usepackage[utf8]{inputenc}
\usepackage{array}
\usepackage{float}
\usepackage{wrapfig}
\usepackage{booktabs}
\usepackage{url}
\usepackage{multirow}
\usepackage{amsmath}
\usepackage{amssymb}
\usepackage{graphicx}
\usepackage{rotating}
\usepackage[authoryear,round]{natbib}

\makeatletter

\pdfpageheight\paperheight
\pdfpagewidth\paperwidth

\providecommand{\tabularnewline}{\\}
\floatstyle{ruled}
\newfloat{algorithm}{tbp}{loa}
\providecommand{\algorithmname}{Algorithm}
\floatname{algorithm}{\protect\algorithmname}

\newtheorem{thm}{Theorem}[section]
\newtheorem{definitn}{Definition}[section]
\newtheorem{lemma}{Lemma}[section]
\newtheorem{cor}{Corollary}[section]

\usepackage{iclr2023_conference,times}

\usepackage{amsmath,amsfonts,bm}

\def\eqref#1{equation~\ref{#1}}
\def\1{\bm{1}}

\DeclareMathAlphabet{\mathsfit}{\encodingdefault}{\sfdefault}{m}{sl}
\SetMathAlphabet{\mathsfit}{bold}{\encodingdefault}{\sfdefault}{bx}{n}

\usepackage{hyperref}
\usepackage{url}

\usepackage[algo2e]{algorithm2e}
\iclrfinalcopy

\makeatother

\usepackage{babel}
\begin{document}
\title{Single-level Adversarial Data Synthesis based on Neural Tangent Kernels}
\author{Yu-Rong Zhang\footnotemark[1]\hspace{1.5cm}Ruei-Yang Su\footnotemark[1]\hspace{1.5cm}Sheng-Yen
Chou\footnotemark[1]\hspace{1.5cm}Shan-Hung Wu\thanks{Department of Computer Science, National Tsing Hua University, Taiwan,
R.O.C.. Correspondence to: Shan-Hung Wu <{\tt\small shwu@cs.nthu.edu.tw}>.}}
\maketitle
\begin{abstract}
Generative adversarial networks (GANs) have achieved impressive performance
in data synthesis and have driven the development of many applications.
However, GANs are known to be hard to train due to their bilevel objective,
which leads to the problems of convergence, mode collapse, and gradient
vanishing. In this paper, we propose a new generative model called
the generative adversarial NTK (GA-NTK) that has a single-level objective.
The GA-NTK keeps the spirit of adversarial learning (which helps generate
plausible data) while avoiding the training difficulties of GANs.
This is done by modeling the discriminator as a Gaussian process with
a neural tangent kernel (NTK-GP) whose training dynamics can be completely
described by a closed-form formula. We analyze the convergence behavior
of GA-NTK trained by gradient descent and give some sufficient conditions
for convergence. We also conduct extensive experiments to study the
advantages and limitations of GA-NTK and propose some techniques that
make GA-NTK more practical.\footnote{Our code is available on GitHub at \url{https://github.com/ga-ntk/ga-ntk}. }
\end{abstract}

\section{Introduction}

Generative adversarial networks (GANs) \citep{goodfellow2014gan,Radford2015dcgan},
a branch of deep generative models based on adversarial learning,
have received much attention due to their novel problem formulation
and impressive performance in data synthesis. Variants of GANs have
also driven recent developments of many applications, such as super-resolution
\citep{ledig2017super-resolution}, image inpainting \citep{xu2014inpainting},
and video generation \citep{vondrick16vgan}. 

A GANs framework consists of a discriminator network $\mathcal{D}$
and a generator network $\mathcal{G}$ parametrized by $\boldsymbol{\theta}_{\mathcal{D}}$
and $\boldsymbol{\theta}_{\mathcal{G}}$, respectively. Given a $d$-dimensional
data distribution $\mathcal{P}_{\text{data}}$ and a $c$-dimensional
noise distribution $\mathcal{P}_{\text{noise}}$, the generator $\mathcal{G}$
maps a random noise $\boldsymbol{z}\in\mathbb{R}^{c}$ to a point
$\mathcal{G}(\boldsymbol{z})\in\mathbb{R}^{d}$ in the data space,
while the discriminator $\mathcal{D}$ takes a point $\boldsymbol{x}'\in\mathbb{R}^{d}$
as the input and tells whether $\boldsymbol{x}'$ is real or fake,
i.e., $\mathcal{D}(\boldsymbol{x}')=1$ if $\boldsymbol{x}'\sim\mathcal{P}_{\text{data}}$
and $\mathcal{D}(\boldsymbol{x}')=0$ if $\boldsymbol{x}'\sim\mathcal{P}_{\text{gen}}$,
where $\mathcal{P}_{\text{gen}}$ is the distribution of $\mathcal{G}(\boldsymbol{z})$
and $\boldsymbol{z}\sim\mathcal{P}_{\text{noise}}$. The objective
of GANs is typically formulated as a bilevel optimization problem:
\begin{equation}
\arg\min_{\boldsymbol{\theta}_{\mathcal{G}}}\max_{\boldsymbol{\theta}_{\mathcal{D}}}\mathbb{E}_{\boldsymbol{x}\sim\mathcal{P}_{\text{data}}}[\log\mathcal{D}(\boldsymbol{x})]+\mathbb{E}_{\boldsymbol{z}\sim\mathcal{P}_{\text{noise}}}[\log(1-\mathcal{D}(\mathcal{G}(\boldsymbol{z})))].\label{eq:obj:gan}
\end{equation}
The discriminator $\mathcal{D}$ and generator $\mathcal{G}$ aim
to break each other through the inner $\max$ and outer $\min$ objectives,
respectively. The studies by \citet{goodfellow2014gan,Radford2015dcgan}
show that this adversarial formulation can lead to a better generator
that produces plausible data points/images.

However, GANs are known to be hard to train due to the following issues
\citep{goodfellow2016nips}. \textbf{Failure to converge.} In practice,
Eq. (\ref{eq:obj:gan}) is usually only approximately solved by an
alternating first-order method such as the alternating stochastic
gradient descent (SGD). The alternating updates for $\boldsymbol{\theta}_{\mathcal{D}}$
and $\boldsymbol{\theta}_{\mathcal{G}}$ may cancel each other's progress.
During each alternating training step, it is also tricky to balance
the number of SGD updates for $\boldsymbol{\theta}_{\mathcal{D}}$
and that for $\boldsymbol{\theta}_{\mathcal{G}}$, as a too small
or large number for $\boldsymbol{\theta}_{\mathcal{D}}$ leads to
low-quality gradients for $\boldsymbol{\theta}_{\mathcal{G}}$. 
\textbf{Mode collapse.} The alternating SGD is attracted by stationary
points and therefore is not good at distinguishing between a $\min_{\boldsymbol{\theta}_{\mathcal{G}}}\max_{\boldsymbol{\theta}_{\mathcal{D}}}$
problem and a $\max_{\boldsymbol{\theta}_{\mathcal{D}}}\min_{\boldsymbol{\theta}_{\mathcal{G}}}$
problem. When the solution to the latter is returned, the generator
tends to always produce the points at modes that best deceive the
discriminator, making $\mathcal{P}_{\text{gen}}$ of low diversity.\footnote{Mode collapse can be caused by other reasons, such as the structure
of $\mathcal{G}$. This paper only solves the problem due to alternating
SGD.} \textbf{Vanishing gradients.} At the beginning of a training process,
the finite real and fake training data may not overlap with each other
in the data space, and thus the discriminator may be able to perfectly
separate the real from fake data. Given the cross-entropy loss (or
more generally, any $f$-divergence measure \citep{renyi1961measures}
between $\mathcal{P}_{\text{data}}$ and $\mathcal{P}_{\text{gen}}$),
the value of the discriminator becomes saturated on both sides of
the decision boundary, resulting in zero gradients for $\boldsymbol{\theta}_{\mathcal{G}}$.

In this paper, we argue that the above issues are rooted in the modeling
of  $\mathcal{D}$. In most existing variants of GANs, the discriminator
is  a deep neural network with explicit weights $\boldsymbol{\theta}_{\mathcal{D}}$.
Under gradient descent, the gradients of $\boldsymbol{\theta}_{\mathcal{G}}$
in Eq. (\ref{eq:obj:gan}) cannot be back-propagated through the inner
$\max_{\boldsymbol{\theta}_{\mathcal{D}}}$ problem because otherwise
it requires the computation of high-order derivatives of $\boldsymbol{\theta}_{\mathcal{D}}$.
 This motivates the use of alternating SGD, which in turn causes
the convergence issues and mode collapse. Furthermore, the $\mathcal{D}$
is a single network whose particularity may cause a catastrophic effect,
such as the vanishing gradients, during training.

We instead model the discriminator $\mathcal{D}$ as a Gaussian process
whose mean and covariance are governed by a kernel function called
the neural tangent kernel (NTK-GP) \citep{jacot2018ntk,lee2019ntk,chizat2018ntk}.
The $\mathcal{D}$ approximates an infinite ensemble of infinitely
wide neural networks in a nonparametric manner and has no explicit
weights. In particular, its training dynamics can be completely described
by a closed-form formula. This allows us to simplify adversarial data
synthesis into a single-level optimization problem, which we call
the generative adversarial NTK (GA-NTK). Moreover, since $\mathcal{D}$
is an infinite ensemble of networks, the particularity of a single
element network does not drastically change the training process.
This makes GA-NTK less prone to vanishing gradients and stabilizes
training even when an $f$-divergence measure between $\mathcal{P}_{\text{data}}$
and $\mathcal{P}_{\text{gen}}$ is used as the loss of $\mathcal{D}$.
The following summarizes our contributions:
\begin{itemize}
\item We propose a single-level optimization method, named GA-NTK, for adversarial
data synthesis. It can be solved by ordinary gradient descent, avoiding
the difficulties of bi-level optimization in GANs.
\item We prove the convergence of GA-NTK training under mild conditions.
We also show that $\mathcal{D}$ being an infinite ensemble of networks
can provide smooth gradients for $\mathcal{G}$, which stabilizes
GA-NTK training and helps fight vanishing gradients.
\item We propose some practical techniques to reduce the memory consumption
of GA-NTK during training and improve the quality of images synthesized
by GA-NTK.
\item We conduct extensive experiments on real-world datasets to study the
advantages and limitations of GA-NTK. In particular, we find that
GA-NTK has much lower sample complexity as compared to GANs, and the
presence of a generator is not necessary to generate images under
the adversarial setting. 
\end{itemize}
Note that the goal of this paper is not to replace existing GANs nor
advance the state-of-the-art performance, but to show that adversarial
data synthesis can be done via a single-level modeling. Our work has
implications for future research. In particular, the low sample complexity
makes GA-NTK suitable for applications, such as medical imaging, where
data are personalized or not easily collectible. In addition, GA-NTK
bridges the gap between kernel methods and adversarial data/image
synthesis and thus enables future studies on the relationship between
kernels and generated data.

\section{Related Work}

\subsection{GANs and Improvements\label{subsec:GANs-and-Improvements}}

\citet{goodfellow2014gan} proposes GANs and gives a theoretical convergence
guarantee in the function space. However, in practice, one can only
optimize the generator and discriminator in Eq. (\ref{eq:obj:gan})
in the parameter/weight space. Many techniques have been proposed
to make the bilevel optimization easier. \textbf{Failure to convergence.}
To solve this problem, studies devise new training algorithms for
GANs \citep{nagarajan2017gan-gd-locally-stable,daskalakisISZ2018gans-optimism}
or more general minimax problems \citep{kumparampi2019smooth-minimax,mokhtari2020minimax-eg-ogda}.
But recent works by \citet{meschederGN18which-gan-do-actually-converge,farnia2020nash-equilibria}
show that there may not be a Nash equilibrium solution in GANs. \textbf{Mode
collapse.} \citet{luke2017unroll-gan} alleviates this issue by back-propagating
the computation of $\boldsymbol{\theta}_{\mathcal{G}}$ through the
discriminators trained with several steps to strengthen the $\min_{\boldsymbol{\theta}_{\mathcal{G}}}\max_{\boldsymbol{\theta}_{\mathcal{D}}}$
property. Other works mitigate mode collapse by diversifying the modes
of $\mathcal{D}$ through regularization \citep{che2017modereg,mao2019mode},
modeling $\mathcal{D}$ as an ensemble of multiple neural networks
\citep{durugkar2017multigan,ghosh18multiagentgan}, or using additional
auxiliary networks\citep{srivastava2017veegan,bang2021mggan,li2021tackling}.
\textbf{Vanishing gradients.} \citet{mao2017lsgan} tries to solve
this problem by using the Pearson $\chi^{2}$-divergence between $\mathcal{P}_{\text{data}}$
and $\mathcal{P}_{\text{gen}}$ as the loss to penalize data points
that are far away from the decision boundary. However, it still suffers
from vanishing gradients as any $f$-divergence measure, including
the cross-entropy loss and Pearson $\chi^{2}$-divergence, cannot
measure the difference between disjoint distributions \citep{sajjadi18tempered-gan}.
Later studies replace the loss with either the Wasserstein distance
\citep{arjovsky2017wasserstein-gan,ishaan2017improved-wgan} or maximum
mean discrepancy \citep{gretton2012mmd,li2015gmmn,Li2017mmd-gan}
that can measure the divergence of disjoint $\mathcal{P}_{\text{data}}$
and $\mathcal{P}_{\text{gen}}$. In addition, the works by \citet{miyato2018spec-norm,qi2020lipschitz-densities}
aim to constrain the Lipschitz continuity of the discriminator to
prevent its value from being saturated.

Despite that many efforts have been made to improve the training of
GANs, most existing approaches address only one or two issues at a
time with different assumptions, and in the meanwhile, they introduce
new hyperparameters or side effects. For example, in the Wasserstein
GANs \citep{arjovsky2017wasserstein-gan,ishaan2017improved-wgan}
mentioned above, efficient computation of Wasserstein distance requires
the discriminator to be Lipschitz continuous. However, realizing Lipschitz
continuity introduces new hyperparameters and could limit the expressiveness
of the discriminator \citep{anil2019sorting}. Until now, training
GANs is still not an easy task because one has to 1) tune many hyperparameters
and 2) strike a balance between the benefits and costs of different
training techniques to generate satisfactory data points/images.

\subsection{Gaussian Processes and Neural Tangent Kernels}

\label{subsec:Neural-Tangent-Kernels}Consider an infinite ensemble
of infinitely wide networks that use the mean square error (MSE) as
the loss and are trained by gradient descent. Recent developments
in deep learning theory show that the prediction of the ensemble can
be approximated by a special instance of Gaussian process called NTK-GP
\citep{jacot2018ntk,lee2019ntk,chizat2018ntk}. The NTK-GP is a Bayesian
method, so it outputs a distribution of possible values for an input
point. The mean and covariance of the NTK-GP prediction are governed
by a kernel function $k(\cdot,\cdot)$ called the neural tangent kernel
(NTK). Given two data points $\boldsymbol{x}^{i}$ and $\boldsymbol{x}^{j}$,
the $k(\boldsymbol{x}^{i},\boldsymbol{x}^{j})$ represents the similarity
score of the two points in a kernel space, which is fixed once the
hyperparameters of the initial weights, activation function, and architecture
of the networks in the target ensemble are determined.

Here, we focus on the mean prediction of NTK-GP as it is relevant
to our study. Consider a supervised learning task given $\mathbb{D}^{n}=(\boldsymbol{X}^{n}\in\mathbb{R}^{n\times d},\boldsymbol{Y}^{n}\in\mathbb{R}^{n\times c})$
as the training set, where there are $n$ examples and each example
consists of a pair of $d$-dimensional input and $c$-dimensional
output. Let $\boldsymbol{K}^{n,n}\in\mathbb{R}^{n\times n}$ be the
kernel matrix for $\boldsymbol{X}^{n}$, i.e., $K_{i,j}^{n,n}=k(\boldsymbol{X}_{i,:}^{n},\boldsymbol{X}_{j,:}^{n})$.
 Then, at time step $t$ during gradient descent, the mean prediction
of NTK-GP for $\boldsymbol{X}^{n}$ evolve as 
\begin{equation}
(\boldsymbol{I}^{n}-e^{-\eta\boldsymbol{K}^{n,n}t})\boldsymbol{Y}^{n}\in\mathbb{R}^{n\times c},\label{eq:obj:ntk-gp}
\end{equation}
where $\boldsymbol{I}^{n}\in\mathbb{R}^{n\times n}$ is an identity
matrix and $\eta$ is a sufficiently small learning rate \citep{jacot2018ntk,lee2019ntk}.

The NTK used in Eq. (\ref{eq:obj:ntk-gp}) can be extended to support
different network architectures, including convolutional neural networks
(CNNs) \citep{arora2019cntk,novak2020bayesian}, recurrent neural
networks (RNNs) \citep{alemohammad2021rntk,greg2019tensor_program_i},
networks with the attention mechanism \citep{hron2020infinite_attention},
and other architectures \citep{greg2019tensor_program_i,arora2019cntk}.
Furthermore, studies \citep{novak2019neural-tangents,lee2020fin-vs-inf,Arora2020ntksmalldatset,geifman2020similarity}
show that NTK-GPs perform similarly to their finite-width counterparts
(neural networks) in many situations and sometimes even better on
small-data tasks.

A recent study by \citet{franceschi2021ntk-perspetive-of-gans} analyzes
the behavior of GANs from the NTK perspective by taking into account
the alternating optimization. It shows that, in theory, the discriminator
can provide a well-defined gradient flow for the generator, which
is opposite to previous theoretical interpretations \citep{arjovsky2017towards}.
Our work, on the other hand, focuses on adversarial data synthesis
\emph{without} alternating optimization.\footnote{From GAN perspective, our work can be regarded as a special case of
the framework proposed by \citet{franceschi2021ntk-perspetive-of-gans},
where the discriminator neglects the effect of historical generator
updates and only distinguish between the true and currently generated
data at each alternating step.} We make contributions in this direction by (1) formally proving the
convergence of the proposed single-level optimization, (2) showing
that a generator network is not necessary to generate plausible images
(although it might be desirable), and (3) proposing the batch-wise
and multi-resolutional extensions that respectively improve the memory
efficiency of training and global coherency of generated image patterns.

\section{GA-NTK}

\label{sec:GA-NTK}We present a new adversarial data synthesis method,
called the generative adversarial NTK (GA-NTK), based on the NTK theory
\citep{jacot2018ntk,lee2019ntk,chizat2018ntk}. For simplicity of
presentation, we let $\mathcal{G}(\boldsymbol{z})=\boldsymbol{z}\in\mathbb{R}^{d}$
and focus on the discriminator for now. We will discuss the case where
$\mathcal{G}(\cdot)$ is a generator network in Section \ref{subsec:GA-NTK-in-Practice}.
Given an unlabeled, $d$-dimensional dataset $\boldsymbol{X}^{n}\in\mathbb{R}^{n\times d}$
of $n$ points, we first augment $\boldsymbol{X}^{n}$ to obtain a
labeled training set $\mathbb{D}^{2n}=(\boldsymbol{X}^{n}\oplus\boldsymbol{Z}^{n}\in\mathbb{R}^{2n\times d},\boldsymbol{1}^{n}\oplus\boldsymbol{0}^{n}\in\mathbb{R}^{2n})$,
where $\boldsymbol{Z}^{n}\in\mathbb{R}^{n\times d}$ contains $n$
generated points, $\boldsymbol{1}^{n}\in\mathbb{R}^{n}$ and $\boldsymbol{0}^{n}\in\mathbb{R}^{n}$
are label vectors of ones and zeros, respectively, and $\oplus$ is
the vertical stack operator. Then, we model a discriminator trained
on $\mathbb{D}^{2n}$ as an NTK-GP. Let $\boldsymbol{K}^{2n,2n}\in\mathbb{R}^{2n\times2n}$
be the kernel matrix for $\boldsymbol{X}^{n}\oplus\boldsymbol{Z}^{n}$,
where the value of each element $K_{i,j}^{2n,2n}=k((\boldsymbol{X}^{n}\oplus\boldsymbol{Z}^{n})_{i,:},(\boldsymbol{X}^{n}\oplus\boldsymbol{Z}^{n})_{j,:})$
can be computed once we decide the initialization, activation function,
and architecture of the element networks in the target infinite ensemble,
i.e., the discriminator. By Eq. (\ref{eq:obj:ntk-gp}) and let $\lambda=\eta\cdot t$,
the mean predictions of the discriminator can be written as 
\begin{equation}
\mathcal{D}(\boldsymbol{X}^{n},\boldsymbol{Z}^{n};k,\lambda)=(\boldsymbol{I}^{2n}-e^{-\lambda\boldsymbol{K}^{2n,2n}})(\boldsymbol{1}^{n}\oplus\boldsymbol{0}^{n})\in\mathbb{R}^{2n},\label{eq:ga-ntk:D}
\end{equation}
where $\boldsymbol{I}^{2n}\in\mathbb{R}^{2n\times2n}$ is an identity
matrix. We formulate the objective of GA-NTK as follows:

\begin{equation}
\arg\min_{\boldsymbol{Z}^{n}}\mathcal{L}(\boldsymbol{Z}^{n}),\text{ {where} }\mathcal{L}(\boldsymbol{Z}^{n})=\Vert\boldsymbol{1}^{2n}-\mathcal{D}(\boldsymbol{X}^{n},\boldsymbol{Z}^{n};k,\lambda)\Vert.\label{eq:obj:ga-ntk}
\end{equation}
$\mathcal{L}(\cdot)$ is the loss function and $\boldsymbol{1}^{2n}\in\mathbb{R}^{2n}$
is a vector of ones. Statistically, Eq. (\ref{eq:obj:ga-ntk}) aims
to minimize the Pearson $\chi^{2}$-divergence \citep{jeffreys1946invariant},
a case of $f$-divergence, between $\mathcal{P}_{\text{data}}+\mathcal{P}_{\text{gen}}$
and $2\mathcal{P}_{\text{gen}}$, where $\mathcal{P}_{\text{gen}}$
is the distribution of generated points. Please see Section \ref{sec:stats}
in Appendix for more details. 

GA-NTK formulates an adversarial data synthesis task as a single-level
optimization problem. On one hand, GA-NTK aims to find points $\boldsymbol{Z}^{n}$
that best deceive the discriminator such that it outputs wrong labels
$\boldsymbol{1}^{2n}$ for these points. On the other hand, the discriminator
is trained on $\mathbb{D}^{2n}$ with the correct labels $\boldsymbol{1}^{n}\oplus\boldsymbol{0}^{n}$
and therefore has the opposite goal of distinguishing between the
real and generated points. Such an adversarial setting can be made
single-level because the training dynamics of the discriminator $\mathcal{D}$
by gradient descent can be completely described by a closed-form formula
in Eq. (\ref{eq:ga-ntk:D})---any change of $\boldsymbol{Z}^{n}$
causes $\mathcal{D}$ to be ``retrained'' instantly. Therefore,
one can easily solve Eq. (\ref{eq:obj:ga-ntk}) by ordinary SGD.

\textbf{Training.} Before running SGD, one needs to tune the hyperparameter
$\lambda$. We show in the next section that the value of $\lambda$
should be large enough but finite. Therefore, the complete training
process of GA-NTK is to 1) find the minimal $\lambda$ that allows
the discriminator to separate real data from pure noises in an auxiliary
task, and 2) solve $\boldsymbol{Z}^{n}$ in Eq. (\ref{eq:obj:ga-ntk})
by ordinary SGD with the fixed $\lambda$. Please see Section 7.3
in Appendix for more details.

\subsection{Merits}

\label{subsec:Merits}As compared to GANs, GA-NTK offers the following
advantages:\textbf{ Convergence.} The GA-NTK can be trained by ordinary
gradient descent. This gives much nicer convergence properties:
\begin{thm}
\label{thm:convergence}Let $s$ be the number of the gradient descent
iterations solving Eq. (\ref{eq:obj:ga-ntk}), and let $\boldsymbol{Z}^{n,(s)}$
be the solution at the $s$-th iteration. Suppose the following values
are bounded: (a) $\boldsymbol{X}_{i,j}^{n}$ and $\boldsymbol{Z}_{i,j}^{n,(0)}$,
$\forall i,j$, (b) $t$ and $\eta$, and (c) $\sigma$ and $L$.
Also, assume that (d) $\boldsymbol{X}^{n}$ contains finite, non-identical,
normalized rows. Then, for a sufficiently large $t$, we have 
\[
\min_{j\leq s}\Vert\nabla_{\boldsymbol{Z}^{n}}\mathcal{L}(\boldsymbol{Z}^{n,(j)})\Vert^{2}\leq O(\frac{1}{s-1}).
\]
\end{thm}

\noindent We prove the above theorem by showing that, with a large
enough $\lambda$, $\nabla_{\boldsymbol{Z}^{n}}\mathcal{L}(\boldsymbol{Z}^{n,(s)})$
is smooth enough to lead to the convergence of gradient descent. For
more details, please see Section 6 in Appendix. \textbf{Diversity.}
GA-NTK avoids  mode collapse due to the confusion between the min-max
and max-min problems in alternating SGD. Given different initial values,
the generated points in $\boldsymbol{Z}^{n}$ can be very different
from each other. \textbf{No vanishing gradients, no side effects.}
The hyperparameter $\lambda$ controls how much $\mathcal{D}$ should
learn from the true and fake data during each iteration. Figure \ref{fig:Comparison-of-gradients}
shows the gradients of $\mathcal{D}$ with a finite $\lambda$, which
do not saturate. This avoids the necessity of using a loss that imposes
side effects, such as the Wasserstein distance \citep{arjovsky2017wasserstein-gan,ishaan2017improved-wgan}
whose efficient evaluation requires Lipschitz continuity of $\mathcal{D}$.

\subsection{GA-NTK in Practice}

\label{subsec:GA-NTK-in-Practice}\textbf{Scalability.} To generate
a large number of points, we can parallelly solve multiple $\boldsymbol{Z}^{n}$'s
in Eq. (\ref{eq:obj:ga-ntk}) on different machines.  On a single
machine, the gradients of $\boldsymbol{Z}^{n}$ need to be back-propagated
through the computation of $\boldsymbol{K}^{2n,2n}$, which has $O(n^{2})$
space complexity. This may incur scalability issues for large datasets.
Although recent efforts by \citet{arora2019cntk,bietti19inductive-bais-ntk,han21random-feature-ntk,zandieh2021scaling}
have been made to reduce the time and space complexity of the evaluation
of NTK and its variants, they are still at an early stage of development
and the consumed space in practice may still be too large. To alleviate
this problem, we propose the batch-wise GA-NTK with the objective
\begin{equation}
\arg\min_{\boldsymbol{Z}^{n}}\mathbb{E}_{\boldsymbol{X}^{b/2}\subset\boldsymbol{X}^{n},\boldsymbol{Z}^{b/2}\subset\boldsymbol{Z}^{n}}\Vert\boldsymbol{1}^{b}-\mathcal{D}(\boldsymbol{X}^{b/2},\boldsymbol{Z}^{b/2};k,\lambda)\Vert,\label{eq:obj:bga-ntk}
\end{equation}
that can be solved using mini-batches: during each gradient descent
iteration, we 1) randomly sample a batch of $b$ rows in $\boldsymbol{X}^{n}\oplus\boldsymbol{Z}^{n}$
and their corresponding labels, and 2) update $\boldsymbol{Z}^{n}$
based on $\boldsymbol{K}^{b,b}$. Although the batch-wise GA-NTK is
cosmetically similar to the original GA-NTK, it solves a different
problem. In the original GA-NTK, the $\boldsymbol{Z}^{n}$ aims to
fool a single discriminator $\mathcal{D}$ trained on $2n$ examples,
while in the batch-wise GA-NTK, the $\boldsymbol{Z}^{n}$'s goal is
to deceive \emph{many} discriminators, each trained on $b$ examples
only. Fortunately, \citet{Shankar2020kernelwithouttangent,Arora2020ntksmalldatset}
have shown that NTK-based methods perform well on small datasets.
We will conduct experiments to verify this later. 

\textbf{Generator Network.} So far, we let $\mathcal{G}(\boldsymbol{z})=\boldsymbol{z}$
and show that a generator is \emph{not} necessary in adversarial data
synthesis.\footnote{In GANs, solving $\boldsymbol{Z}$ directly against a finite-width
discriminator is infeasible because it amounts to finding adversarial
examples \citep{goodfellow2014explaining} whose gradients are known
to be very noisy \citep{ilyas2019adversarial}.} Nevertheless, the presence of a generator network may be favorable
in some applications to save time and memory at inference time. This
can be done by extending the batch-wise GA-NTK as follows:
\begin{equation}
\arg\min_{\boldsymbol{\theta}_{\mathcal{G}}}\mathbb{E}_{\boldsymbol{X}^{b/2}\subset\boldsymbol{X}^{n},\boldsymbol{Z}^{b/2}\sim\mathcal{N}(\mathbf{0},\mathbf{I})}\Vert\boldsymbol{1}^{b}-\mathcal{D}(\boldsymbol{X}^{b/2},\mathcal{G}(\boldsymbol{Z}^{b/2};\boldsymbol{\theta}_{\mathcal{G}});k,\lambda)\Vert,\label{eq:obj:bga-ntk-gen}
\end{equation}
where $\mathcal{G}(\cdot\,;\boldsymbol{\theta}_{\mathcal{G}})$ is
a generator network parametrized by $\boldsymbol{\theta}_{\mathcal{G}}$,
and $\boldsymbol{Z}\in\mathbb{R}^{l}$ where $l\leq d$. Note that
this is still a single-level objective, and $\boldsymbol{\theta}_{\mathcal{G}}$
can be solved by gradient descent. We denote this variant GA-NTKg.

\textbf{Image Quality.} To generate images, one can pair up GA-NTK
with a convolutional neural tangent kernel (CNTK) \citep{arora2019cntk,novak2020bayesian,Garriga-Alonso2019cnngp,greg2019scalinglimit}
that approximates a CNN with infinite channels. This allows the NTK-GP
(discriminator) to distinguish between real and fake points based
on local patterns in the pixel space. However, the images synthesized
by this GA-NTK variant may lack global coherency, just like the images
generated by the CNN-based GANs \citep{Radford2015dcgan,tim2016training-gan}.
Many efforts have been made to improve the image quality of CNN-based
GANs, and this paper opens up opportunities for them to be adapted
to the kernel regime. In particular, we propose the multi-resolutional
GA-CNTK based on the work by \citet{wang2018high}, whose objective
is formulated as:
\begin{equation}
\arg\min_{\boldsymbol{Z}^{n}}\sum_{m}\Vert\boldsymbol{1}^{2n}-\mathcal{D}^{m}(\text{pool}^{m}(\boldsymbol{X}^{n}),\text{pool}^{m}(\boldsymbol{Z}^{n});k^{m},\lambda^{m})\Vert,\label{eq:obj:ga-ntk-multi-resolution}
\end{equation}
where $\mathcal{D}^{m}$ is an NTK-GP taking input at a particular
pixel resolution and $\text{pool}^{m}(\cdot)$ is a downsample operation
(average pooling) applied to each row of $\boldsymbol{X}^{n}$ and
$\boldsymbol{Z}^{n}$. The generated points in $\boldsymbol{Z}^{n}$
aim to simultaneously fool multiple NTK-GPs (discriminators), each
classifying real and fake images at a distinct pixel resolution. The
NTK-GPs working at low and high resolutions encourage global coherency
and details, respectively, and together they lead to more plausible
points in $\boldsymbol{Z}^{n}$.

\section{Experiments}

\label{sec:Experiments}We conduct experiments to study how GA-NTK
works in image generation.

\textbf{Datasets.} We consider the unsupervised/unconditional image
synthesis tasks over real-world datasets, including MNIST \citep{lecun2010mnist},
CIFAR-10 \citep{Krizhevsky09learningmultiple}, CelebA \citep{liu2015faceattributes},
CelebA-HQ \citep{liu2015faceattributes}, and ImageNet \citep{deng2009imagenet}.
To improve training efficiency, we resize CelebA images to 64$\times$64
and ImageNet images to 128$\times$128 pixels, respectively. We also
create a 2D toy dataset consisting of 25-modal Gaussian mixtures of
points to visualize the behavior of different image synthesis methods.\textbf{
GA-NTK implementations.} GA-NTK works with different NTK-GPs. For
the image synthesis tasks, we consider the NTK-GPs that model the
ensembles of fully-connected networks \citep{jacot2018ntk,lee2019ntk,chizat2018ntk}
and convolutional networks \citep{arora2019cntk,novak2020bayesian,Garriga-Alonso2019cnngp,greg2019scalinglimit},
respectively. We implement GA-NTK using the Neural Tangents library
\citep{novak2019neural-tangents} and call the variants based on the
former and latter NTK-GPs the GA-FNTK and GA-CNTK, respectively. In
GA-FNTK, an element network of the discriminator has 3 infinitely
wide, fully-connected layers with ReLU non-linearity, while in GA-CNTK,
an element network follows the architecture of InfoGAN \citep{Chen16InfoGAN}
except for having infinite filters at each layer.  We tune the hyperparameters
of GA-FNTK and GA-CNTK following the method proposed in \citet{Poole16transient-chaos,schoenholz17deep-information-prop,raghus17exp-of-dnn}.
We also implement their batch-wise, generator, and multi-resolutional
variants described in Section \ref{subsec:GA-NTK-in-Practice}. See
Section 7 in Appendix for more details.\textbf{ Baselines.} We compare
GA-NTK with some popular variants of GANs, including vanilla GANs
\citep{goodfellow2014gan}, DCGAN \citep{Radford2015dcgan}, LSGAN
\citep{mao2017lsgan}, WGAN \citep{arjovsky2017wasserstein-gan},
WGAN-GP \citep{ishaan2017improved-wgan}, SNGAN \citep{miyato2018spec-norm}
and StyleGAN2 \citep{karras2020analyzing}. To give a fair comparison,
we let the discriminator of each baseline follow the architecture
of InfoGAN \citep{Chen16InfoGAN} and tune the hyperparameters using
grid search. \textbf{Metrics.} We evaluate the quality of a set of
generated images using the Fréchet Inception Distance (FID) \citep{Heusel17FID}.
The lower the FID score the better. We find that an image synthesis
method may produce downgrade images that look almost identical to
some images in the training set. Therefore, we also use a metric called
the average max-SSIM (AM-SSIM) that calculates the average of the
maximum SSIM score \citep{wang2004ssim} between $\mathcal{P}_{\text{gen}}$
and $\mathcal{P}_{\text{data}}$: 
\[
\text{AM-SSIM}(\mathcal{P}_{\text{gen}},\mathcal{P}_{\text{data}})=\mathbb{E}_{\boldsymbol{x}'\sim\mathcal{P}_{\text{gen}}}[\max_{\boldsymbol{x}\sim\mathcal{P}_{\text{data}}}\text{SSIM}(\boldsymbol{x}',\boldsymbol{x})].
\]
 A generated image set will have a higher AM-SSIM score if it contains
downgrade images.\textbf{ Environment and limitations.} We conduct
all experiments on a cluster of machines having 80 NVIDIA Tesla V100
GPUs. As discussed in \ref{subsec:GA-NTK-in-Practice}, GA-NTK consumes
a significant amount of memory on each machine due to the computations
involved in the kernel matrix $\boldsymbol{K}^{2n,2n}$. With the
current version of Neural Tangents library \citep{novak2019neural-tangents}
and a V100 GPU of 32GB RAM, the maximum sizes of the training set
from MNIST, CIFAR-10, CelebA, and ImageNet are 1024, 512, 256, and
128, respectively (where the computation graph and backprop operations
of $\boldsymbol{K}^{2n,2n}$ consume about 27.5 GB RAM excluding other
necessary operations). Since our goal is not to achieve state-of-the-art
performance but to compare different image synthesis methods, we train
all the methods using up to 256 images randomly sampled from all classes
of MNIST, the ``horse'' class CIFAR-10, the ``male with straight
hair'' class of CelebA, and the ``daisy'' class of ImageNet, respectively.
We will conduct larger-scale experiments in Section \ref{subsec:Batch-Wise-GA-NTK}.
For more details about our experiment settings, please see Section
7 in Appendix.
\begin{table}
\noindent \begin{centering}
\caption{\label{tab:FID=000026SSIM}The FID and AM-SSIM scores of the images
generated by different methods.}
\par\end{centering}
\noindent \centering{}{\footnotesize{}}%
\begin{tabular}{ccc>{\centering}p{1cm}>{\centering}p{1cm}>{\centering}p{1cm}>{\centering}p{1cm}>{\centering}p{1cm}>{\centering}p{1cm}>{\centering}p{1cm}>{\centering}p{1cm}}
\toprule 
 & {\scriptsize{}$n$} & {\scriptsize{}Metric} & {\scriptsize{}DCGAN} & {\scriptsize{}LSGAN} & {\scriptsize{}WGAN} & {\scriptsize{}WGANGP} & {\scriptsize{}SNGAN} & {\scriptsize{}StyleGAN} & {\scriptsize{}GACNTK} & {\scriptsize{}GACNTKg}\tabularnewline
\midrule
\midrule 
\multirow{6}{*}{\begin{turn}{90}
{\scriptsize{}MNIST}
\end{turn}} & \multirow{2}{*}{{\scriptsize{}64}} & {\scriptsize{}FID} & {\scriptsize{}27.43} & {\scriptsize{}69.76} & {\scriptsize{}50.69} & {\scriptsize{}32.49} & {\scriptsize{}57.89} & {\scriptsize{}91.82} & {\scriptsize{}31.10} & {\scriptsize{}32.43}\tabularnewline
\cmidrule{3-11} \cmidrule{4-11} \cmidrule{5-11} \cmidrule{6-11} \cmidrule{7-11} \cmidrule{8-11} \cmidrule{9-11} \cmidrule{10-11} \cmidrule{11-11} 
 &  & {\scriptsize{}AMSSIM} & {\scriptsize{}0.84} & {\scriptsize{}0.79} & {\scriptsize{}0.77} & {\scriptsize{}0.83} & {\scriptsize{}0.67} & {\scriptsize{}0.69} & {\scriptsize{}0.49} & {\scriptsize{}0.71}\tabularnewline
\cmidrule{2-11} \cmidrule{3-11} \cmidrule{4-11} \cmidrule{5-11} \cmidrule{6-11} \cmidrule{7-11} \cmidrule{8-11} \cmidrule{9-11} \cmidrule{10-11} \cmidrule{11-11} 
 & \multirow{2}{*}{{\scriptsize{}128}} & {\scriptsize{}FID} & {\scriptsize{}31.89} & {\scriptsize{}38.52} & {\scriptsize{}49.28} & {\scriptsize{}30.20} & {\scriptsize{}38.33} & {\scriptsize{}88.31} & {\scriptsize{}21.14} & {\scriptsize{}36.50}\tabularnewline
\cmidrule{3-11} \cmidrule{4-11} \cmidrule{5-11} \cmidrule{6-11} \cmidrule{7-11} \cmidrule{8-11} \cmidrule{9-11} \cmidrule{10-11} \cmidrule{11-11} 
 &  & {\scriptsize{}AMSSIM} & {\scriptsize{}0.85} & {\scriptsize{}0.80} & {\scriptsize{}0.74} & {\scriptsize{}0.76} & {\scriptsize{}0.67} & {\scriptsize{}0.66} & {\scriptsize{}0.52} & {\scriptsize{}0.72}\tabularnewline
\cmidrule{2-11} \cmidrule{3-11} \cmidrule{4-11} \cmidrule{5-11} \cmidrule{6-11} \cmidrule{7-11} \cmidrule{8-11} \cmidrule{9-11} \cmidrule{10-11} \cmidrule{11-11} 
 & \multirow{2}{*}{{\scriptsize{}256}} & {\scriptsize{}FID} & {\scriptsize{}69.76} & {\scriptsize{}35.33} & {\scriptsize{}50.33} & {\scriptsize{}24.37} & {\scriptsize{}29.49} & {\scriptsize{}84.7} & {\scriptsize{}14.96} & {\scriptsize{}51.21}\tabularnewline
\cmidrule{3-11} \cmidrule{4-11} \cmidrule{5-11} \cmidrule{6-11} \cmidrule{7-11} \cmidrule{8-11} \cmidrule{9-11} \cmidrule{10-11} \cmidrule{11-11} 
 &  & {\scriptsize{}AMSSIM} & {\scriptsize{}0.69} & {\scriptsize{}0.78} & {\scriptsize{}0.72} & {\scriptsize{}0.73} & {\scriptsize{}0.70} & {\scriptsize{}0.65} & {\scriptsize{}0.54} & {\scriptsize{}0.73}\tabularnewline
\midrule 
\multirow{6}{*}{\begin{turn}{90}
{\scriptsize{}CIFAR-10}
\end{turn}} & \multirow{2}{*}{{\scriptsize{}64}} & {\scriptsize{}FID} & {\scriptsize{}312.21} & {\scriptsize{}258.41} & {\scriptsize{}117.85} & {\scriptsize{}49.29} & {\scriptsize{}118.16} & {\scriptsize{}406.02} & {\scriptsize{}55.54} & {\scriptsize{}106.44}\tabularnewline
\cmidrule{3-11} \cmidrule{4-11} \cmidrule{5-11} \cmidrule{6-11} \cmidrule{7-11} \cmidrule{8-11} \cmidrule{9-11} \cmidrule{10-11} \cmidrule{11-11} 
 &  & {\scriptsize{}AMSSIM} & {\scriptsize{}0.22} & {\scriptsize{}0.25} & {\scriptsize{}0.29} & {\scriptsize{}0.74} & {\scriptsize{}0.28} & {\scriptsize{}0.64} & {\scriptsize{}0.41} & {\scriptsize{}0.44}\tabularnewline
\cmidrule{2-11} \cmidrule{3-11} \cmidrule{4-11} \cmidrule{5-11} \cmidrule{6-11} \cmidrule{7-11} \cmidrule{8-11} \cmidrule{9-11} \cmidrule{10-11} \cmidrule{11-11} 
 & \multirow{2}{*}{{\scriptsize{}128}} & {\scriptsize{}FID} & {\scriptsize{}229.94} & {\scriptsize{}339.27} & {\scriptsize{}101.90} & {\scriptsize{}68.53} & {\scriptsize{}128.65} & {\scriptsize{}484.36} & {\scriptsize{}39.98} & {\scriptsize{}61.19}\tabularnewline
\cmidrule{3-11} \cmidrule{4-11} \cmidrule{5-11} \cmidrule{6-11} \cmidrule{7-11} \cmidrule{8-11} \cmidrule{9-11} \cmidrule{10-11} \cmidrule{11-11} 
 &  & {\scriptsize{}AMSSIM} & {\scriptsize{}0.36} & {\scriptsize{}0.10} & {\scriptsize{}0.26} & {\scriptsize{}0.60} & {\scriptsize{}0.21} & {\scriptsize{}0.39} & {\scriptsize{}0.41} & {\scriptsize{}0.44}\tabularnewline
\cmidrule{2-11} \cmidrule{3-11} \cmidrule{4-11} \cmidrule{5-11} \cmidrule{6-11} \cmidrule{7-11} \cmidrule{8-11} \cmidrule{9-11} \cmidrule{10-11} \cmidrule{11-11} 
 & \multirow{2}{*}{{\scriptsize{}256}} & {\scriptsize{}FID} & {\scriptsize{}181.15} & {\scriptsize{}255.19} & {\scriptsize{}111.92} & {\scriptsize{}85.34} & {\scriptsize{}107.29} & {\scriptsize{}426.58} & {\scriptsize{}28.40} & {\scriptsize{}55.46}\tabularnewline
\cmidrule{3-11} \cmidrule{4-11} \cmidrule{5-11} \cmidrule{6-11} \cmidrule{7-11} \cmidrule{8-11} \cmidrule{9-11} \cmidrule{10-11} \cmidrule{11-11} 
 &  & {\scriptsize{}AMSSIM} & {\scriptsize{}0.27} & {\scriptsize{}0.22} & {\scriptsize{}0.22} & {\scriptsize{}0.46} & {\scriptsize{}0.20} & {\scriptsize{}0.26} & {\scriptsize{}0.42} & {\scriptsize{}0.44}\tabularnewline
\midrule 
\multirow{6}{*}{\begin{turn}{90}
{\scriptsize{}CelebA}
\end{turn}} & \multirow{2}{*}{{\scriptsize{}64}} & {\scriptsize{}FID} & {\scriptsize{}489.82} & {\scriptsize{}83.71} & {\scriptsize{}122.36} & {\scriptsize{}83.71} & {\scriptsize{}169.04} & {\scriptsize{}323.37} & {\scriptsize{}30.83} & {\scriptsize{}95.91}\tabularnewline
\cmidrule{3-11} \cmidrule{4-11} \cmidrule{5-11} \cmidrule{6-11} \cmidrule{7-11} \cmidrule{8-11} \cmidrule{9-11} \cmidrule{10-11} \cmidrule{11-11} 
 &  & {\scriptsize{}AMSSIM} & {\scriptsize{}0.02} & {\scriptsize{}0.05} & {\scriptsize{}0.29} & {\scriptsize{}0.56} & {\scriptsize{}0.29} & {\scriptsize{}0.23} & {\scriptsize{}0.60} & {\scriptsize{}0.21}\tabularnewline
\cmidrule{2-11} \cmidrule{3-11} \cmidrule{4-11} \cmidrule{5-11} \cmidrule{6-11} \cmidrule{7-11} \cmidrule{8-11} \cmidrule{9-11} \cmidrule{10-11} \cmidrule{11-11} 
 & \multirow{2}{*}{{\scriptsize{}128}} & {\scriptsize{}FID} & {\scriptsize{}55.01} & {\scriptsize{}450.81} & {\scriptsize{}125.82} & {\scriptsize{}92.73} & {\scriptsize{}168.11} & {\scriptsize{}337.58} & {\scriptsize{}33.51} & {\scriptsize{}58.39}\tabularnewline
\cmidrule{3-11} \cmidrule{4-11} \cmidrule{5-11} \cmidrule{6-11} \cmidrule{7-11} \cmidrule{8-11} \cmidrule{9-11} \cmidrule{10-11} \cmidrule{11-11} 
 &  & {\scriptsize{}AMSSIM} & {\scriptsize{}0.03} & {\scriptsize{}0.11} & {\scriptsize{}0.28} & {\scriptsize{}0.54} & {\scriptsize{}0.28} & {\scriptsize{}0.21} & {\scriptsize{}0.51} & {\scriptsize{}0.38}\tabularnewline
\cmidrule{2-11} \cmidrule{3-11} \cmidrule{4-11} \cmidrule{5-11} \cmidrule{6-11} \cmidrule{7-11} \cmidrule{8-11} \cmidrule{9-11} \cmidrule{10-11} \cmidrule{11-11} 
 & \multirow{2}{*}{{\scriptsize{}256}} & {\scriptsize{}FID} & {\scriptsize{}461.95} & {\scriptsize{}403.79} & {\scriptsize{}108.07} & {\scriptsize{}79.36} & {\scriptsize{}161.20} & {\scriptsize{}333.16} & {\scriptsize{}63.15} & {\scriptsize{}78.46}\tabularnewline
\cmidrule{3-11} \cmidrule{4-11} \cmidrule{5-11} \cmidrule{6-11} \cmidrule{7-11} \cmidrule{8-11} \cmidrule{9-11} \cmidrule{10-11} \cmidrule{11-11} 
 &  & {\scriptsize{}AMSSIM} & {\scriptsize{}0.04} & {\scriptsize{}0.09} & {\scriptsize{}0.31} & {\scriptsize{}0.39} & {\scriptsize{}0.27} & {\scriptsize{}0.30} & {\scriptsize{}0.38} & {\scriptsize{}0.40}\tabularnewline
\bottomrule
\end{tabular}{\footnotesize\par}
\end{table}
\begin{figure*}
\noindent \begin{centering}
{\footnotesize{}}%
\begin{tabular*}{5.5in}{@{\extracolsep{\fill}}ll}
\begin{turn}{90}
{\footnotesize{}}%
\begin{tabular}{c}
{\footnotesize{}(a)}\tabularnewline
{\footnotesize{}WGAN}\tabularnewline
\end{tabular}
\end{turn} & {\footnotesize{}\includegraphics[width=4.2cm]{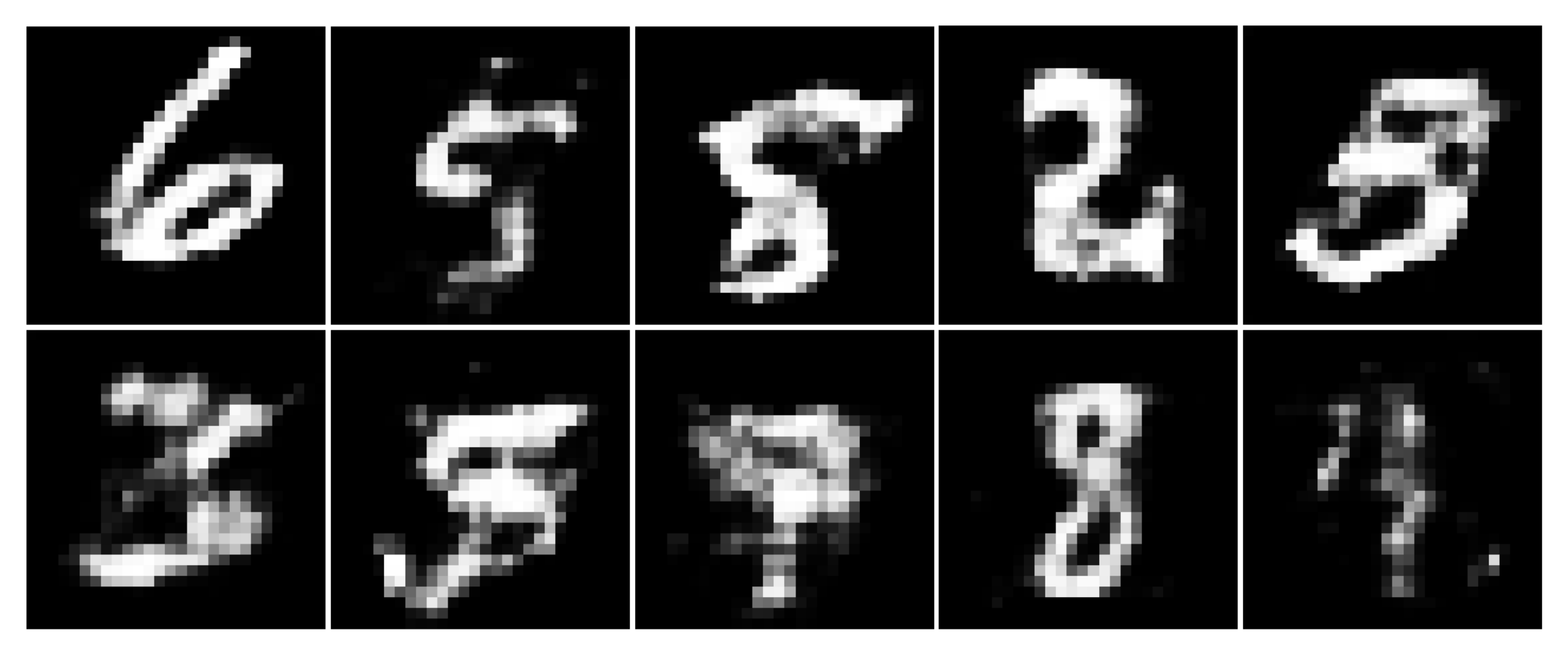}\includegraphics[width=4.2cm]{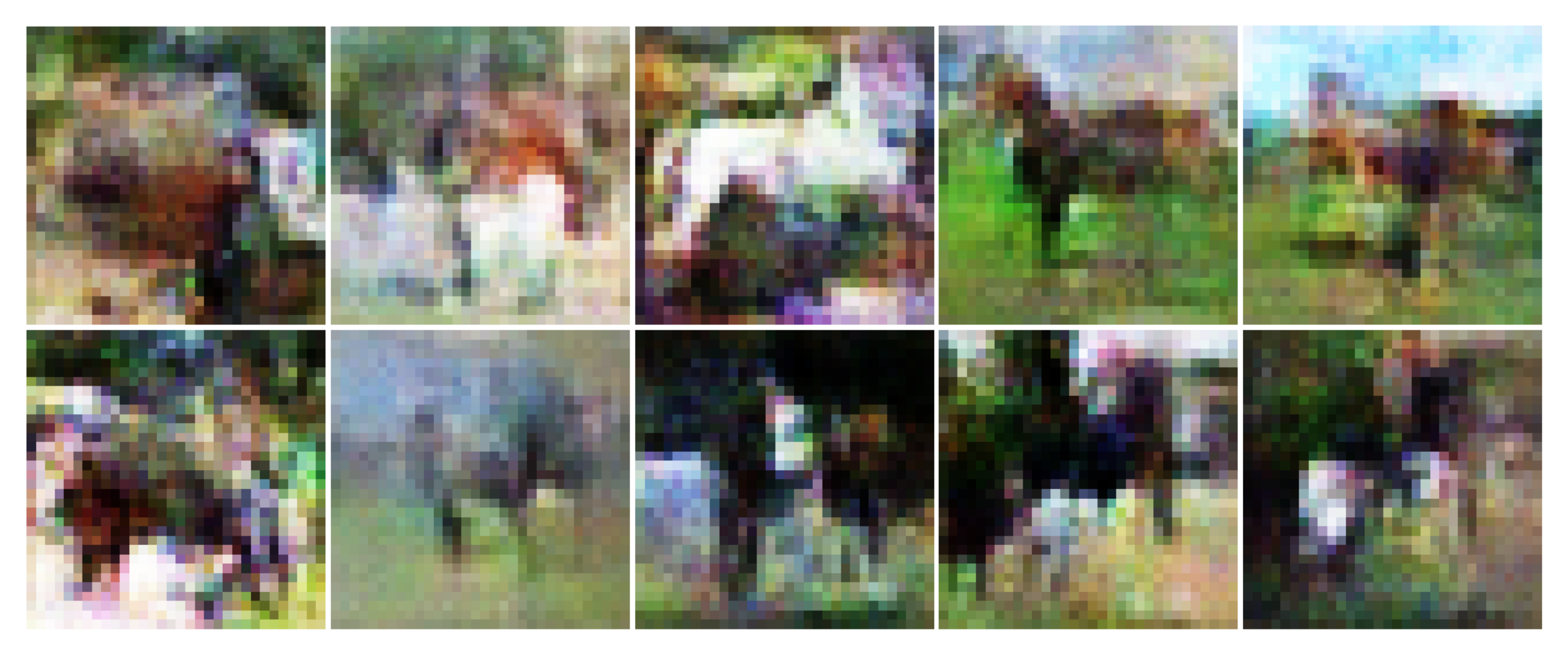}\includegraphics[width=4.2cm]{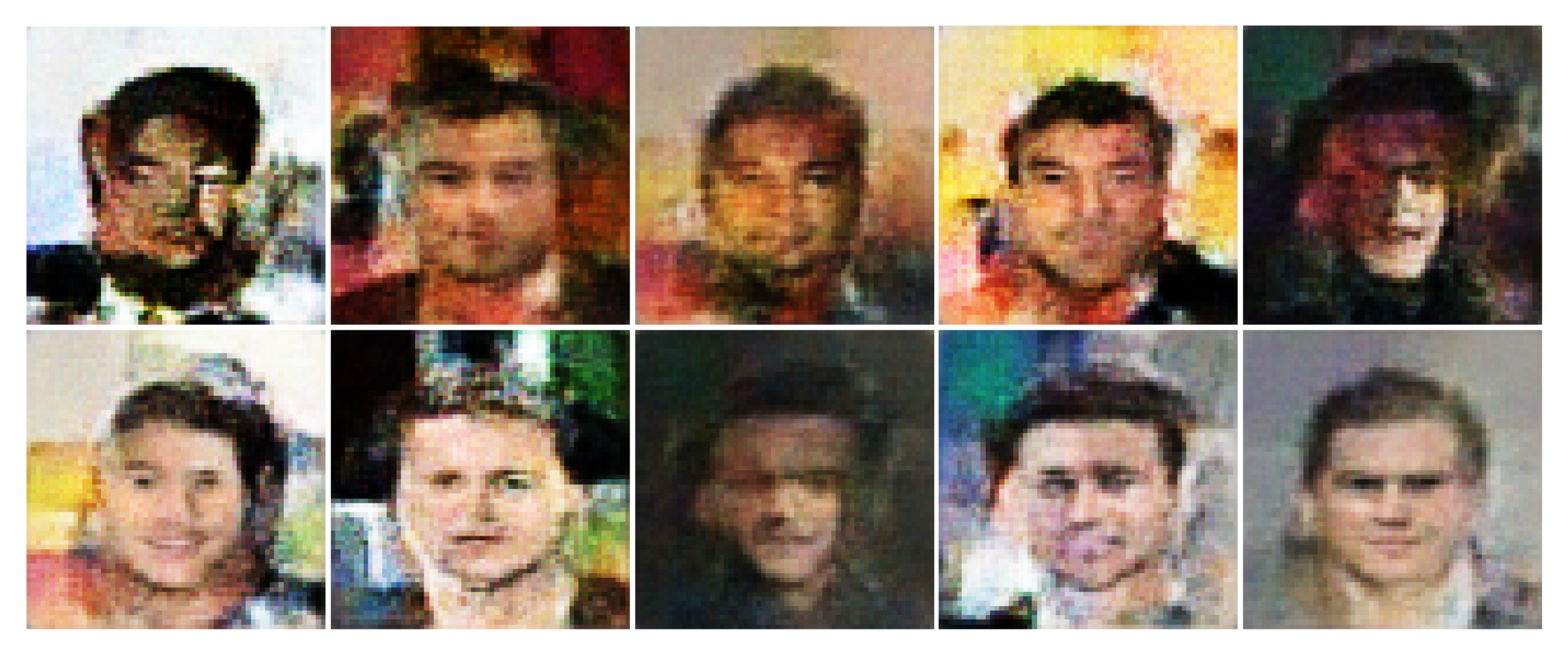}}\tabularnewline
\begin{turn}{90}
{\footnotesize{}}%
\begin{tabular}{c}
{\footnotesize{}(b)}\tabularnewline
{\footnotesize{}WGANGP}\tabularnewline
\end{tabular}
\end{turn} & {\footnotesize{}\includegraphics[width=4.2cm]{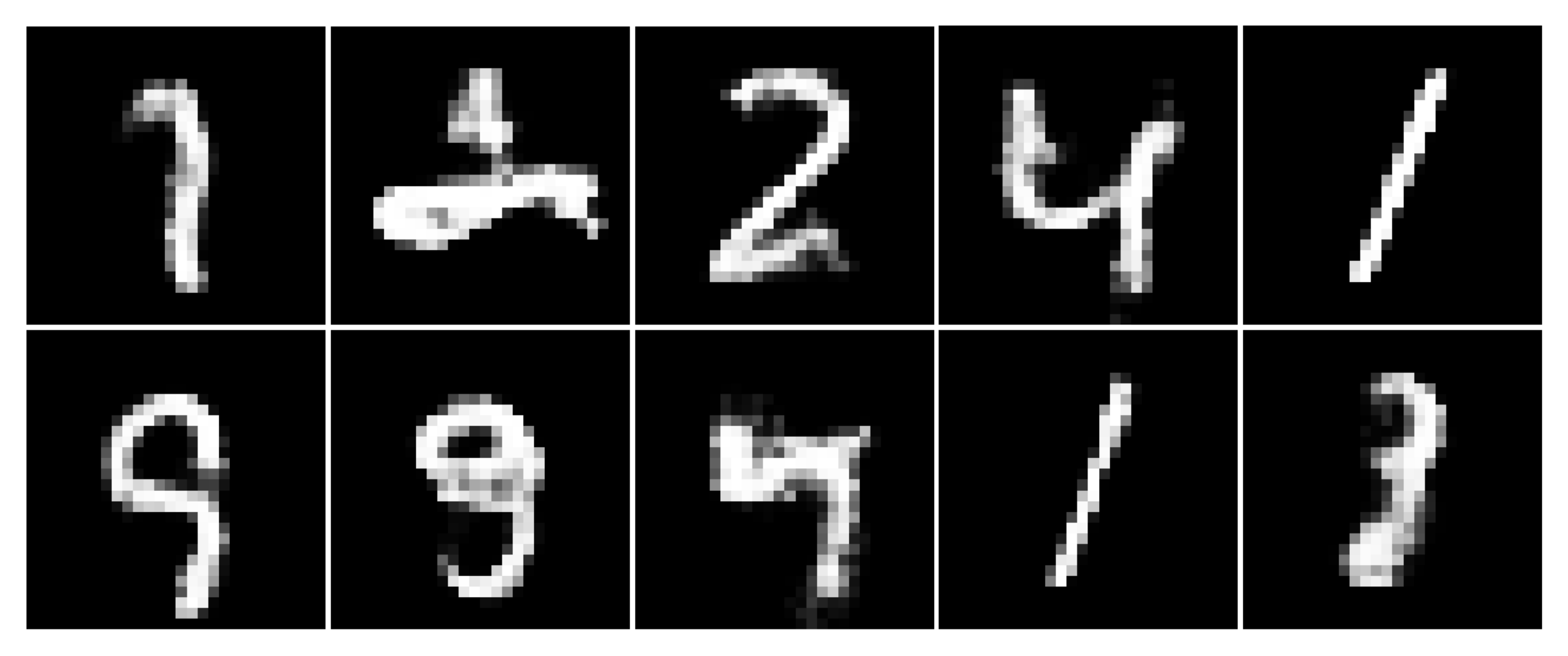}\includegraphics[width=4.2cm]{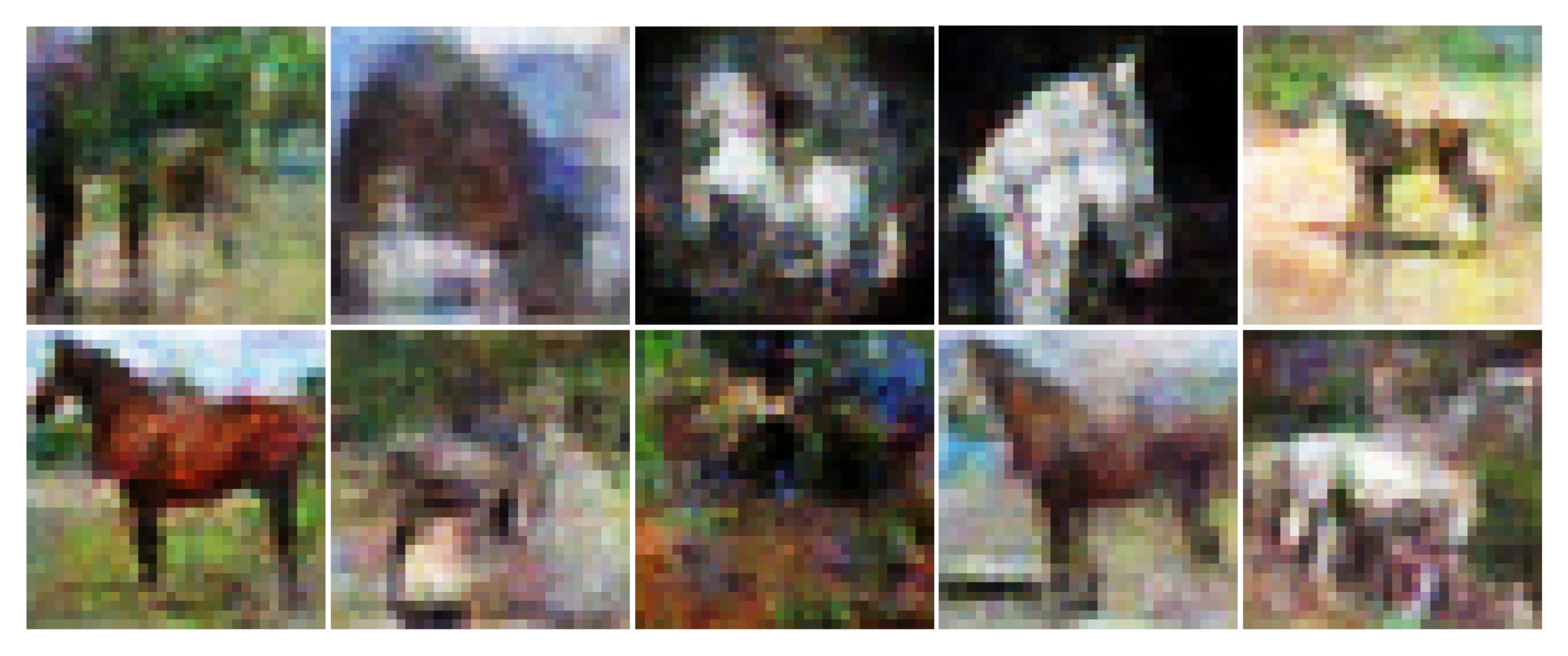}\includegraphics[width=4.2cm]{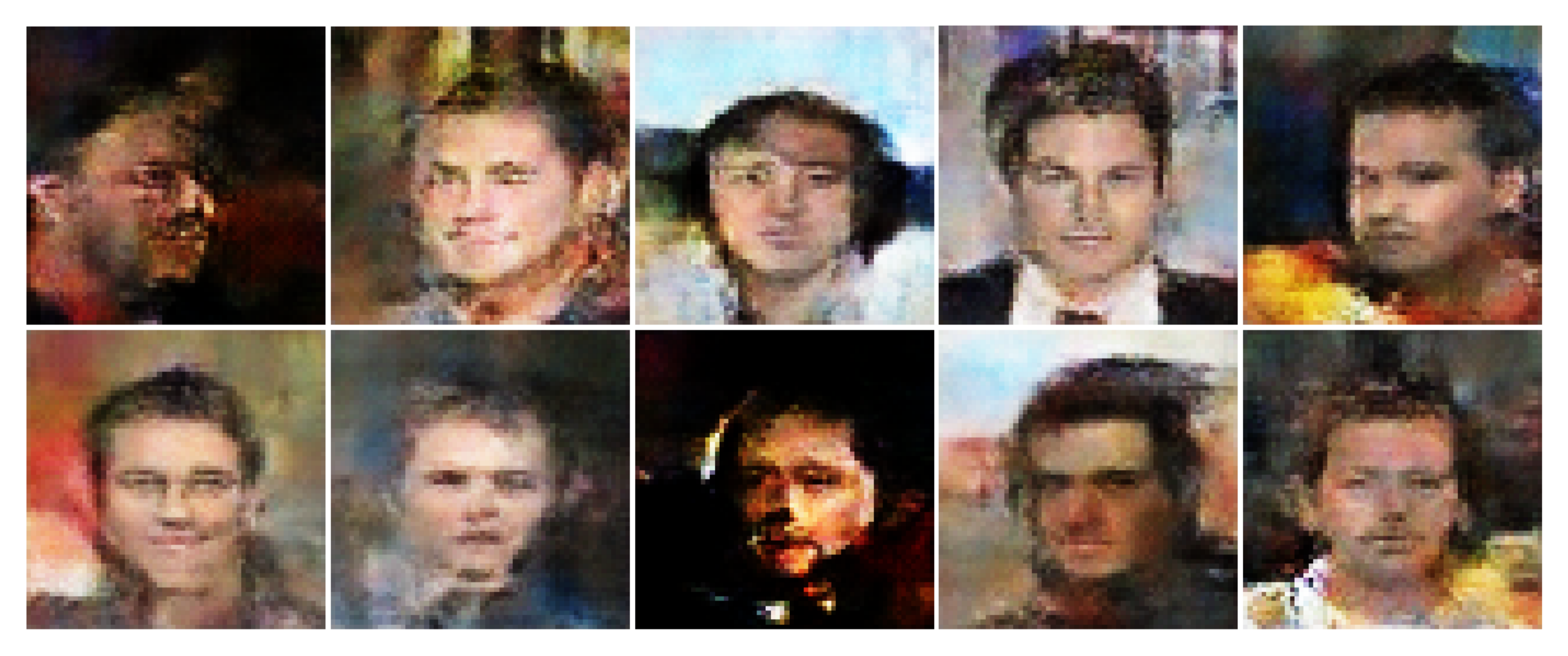}}\tabularnewline
\begin{turn}{90}
{\footnotesize{}}%
\begin{tabular}{c}
{\footnotesize{}(c)}\tabularnewline
{\footnotesize{}SNGAN}\tabularnewline
\end{tabular}
\end{turn} & {\footnotesize{}\includegraphics[width=4.2cm]{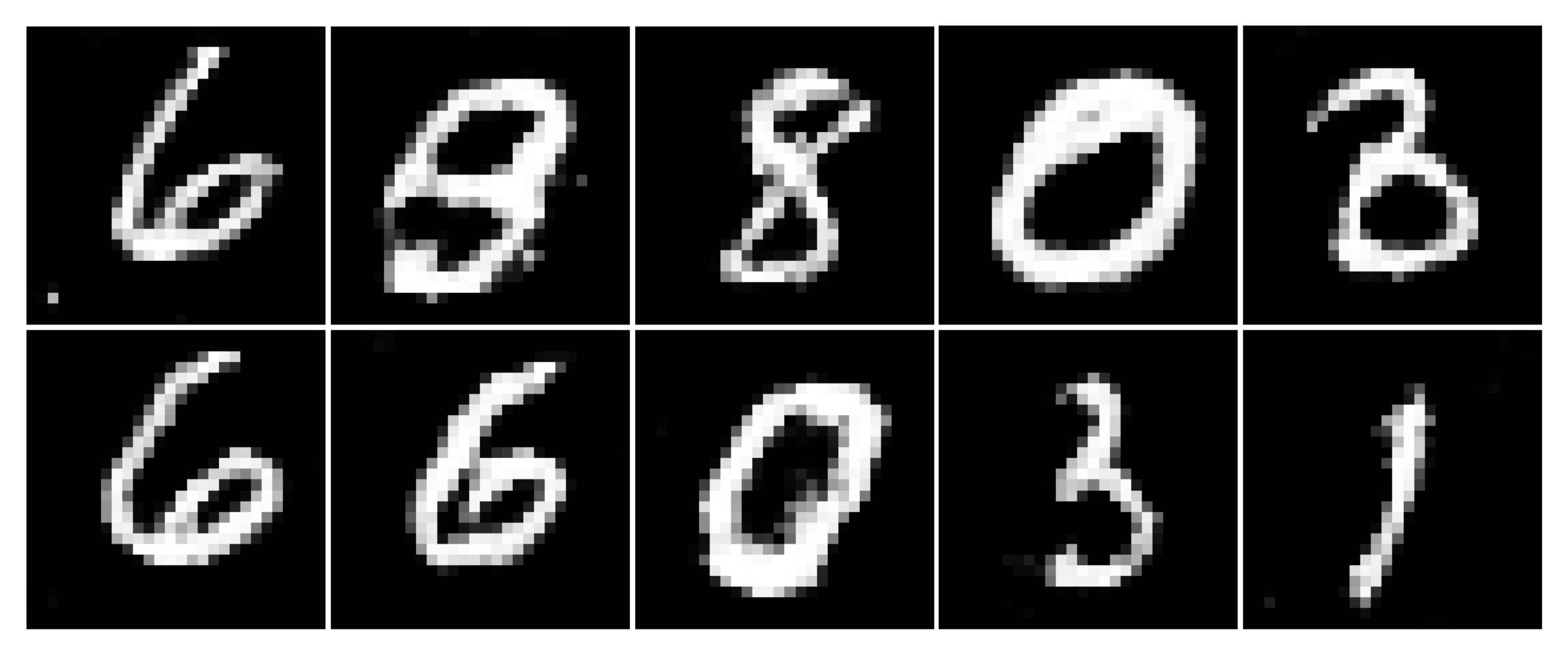}\includegraphics[width=4.2cm]{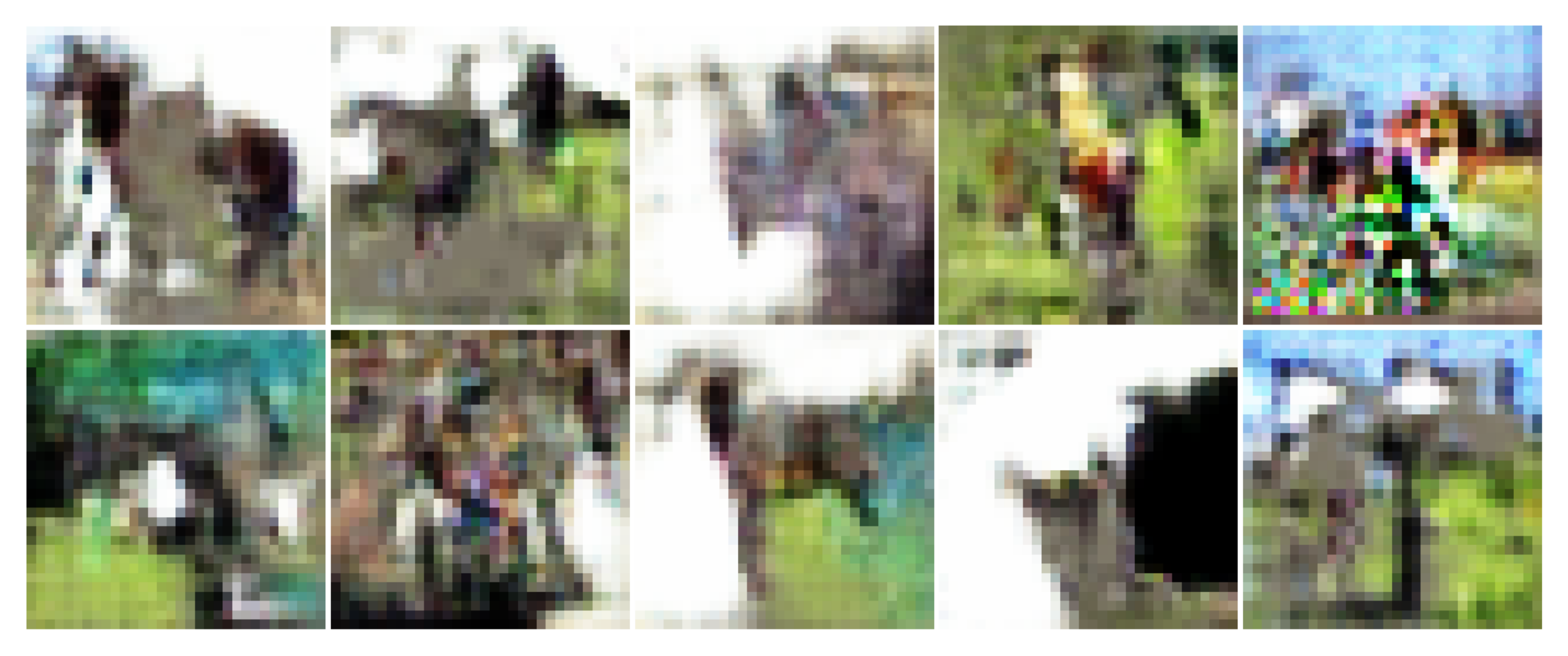}\includegraphics[width=4.2cm]{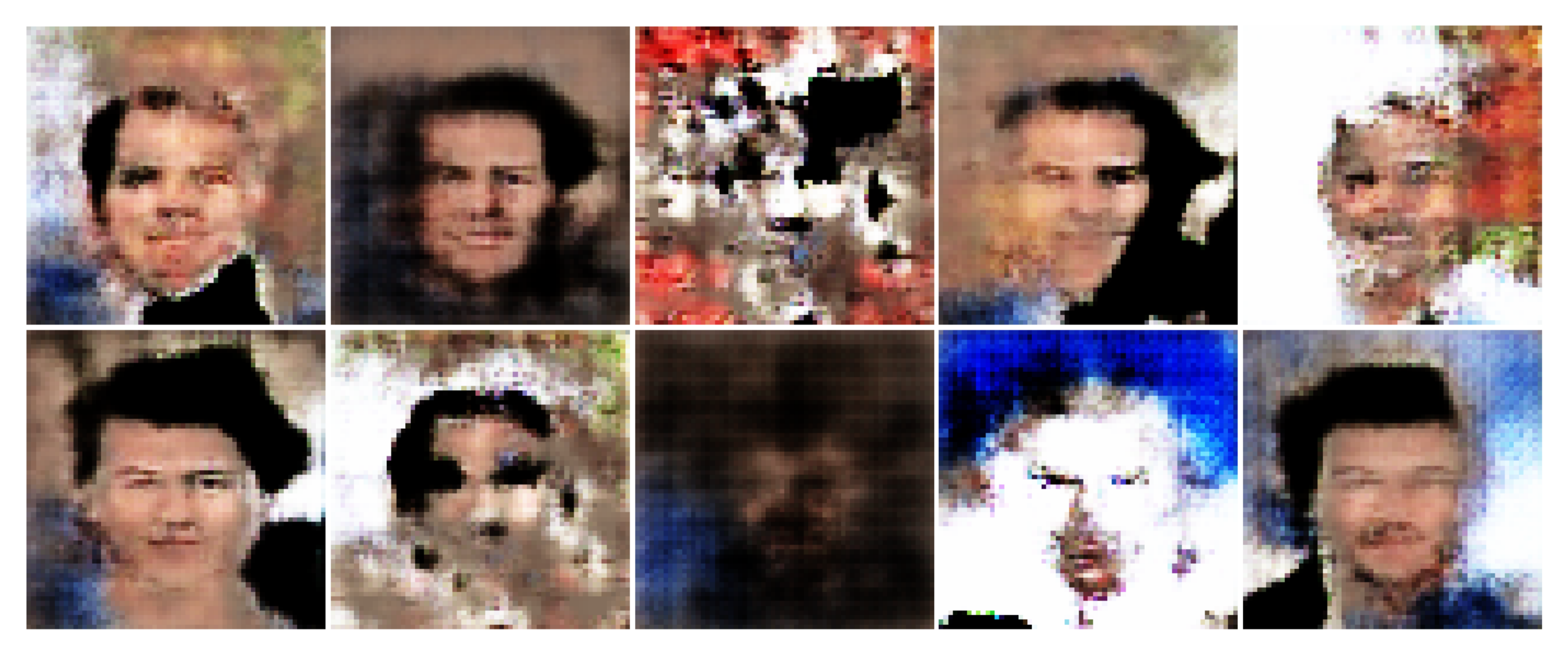}}\tabularnewline
\begin{turn}{90}
{\footnotesize{}}%
\begin{tabular}{c}
{\footnotesize{}(d)}\tabularnewline
{\footnotesize{}GACNTK}\tabularnewline
\end{tabular}
\end{turn} & {\footnotesize{}\includegraphics[width=4.2cm]{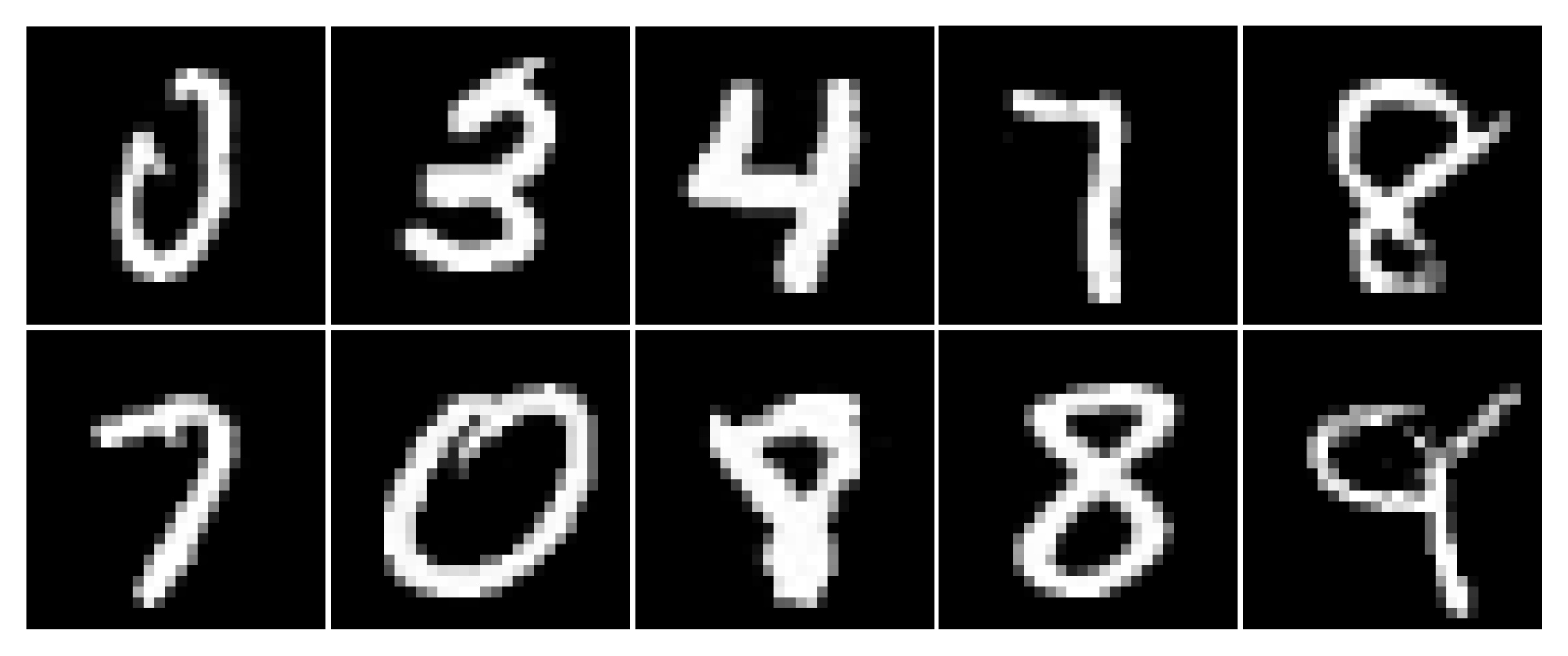}\includegraphics[width=4.2cm]{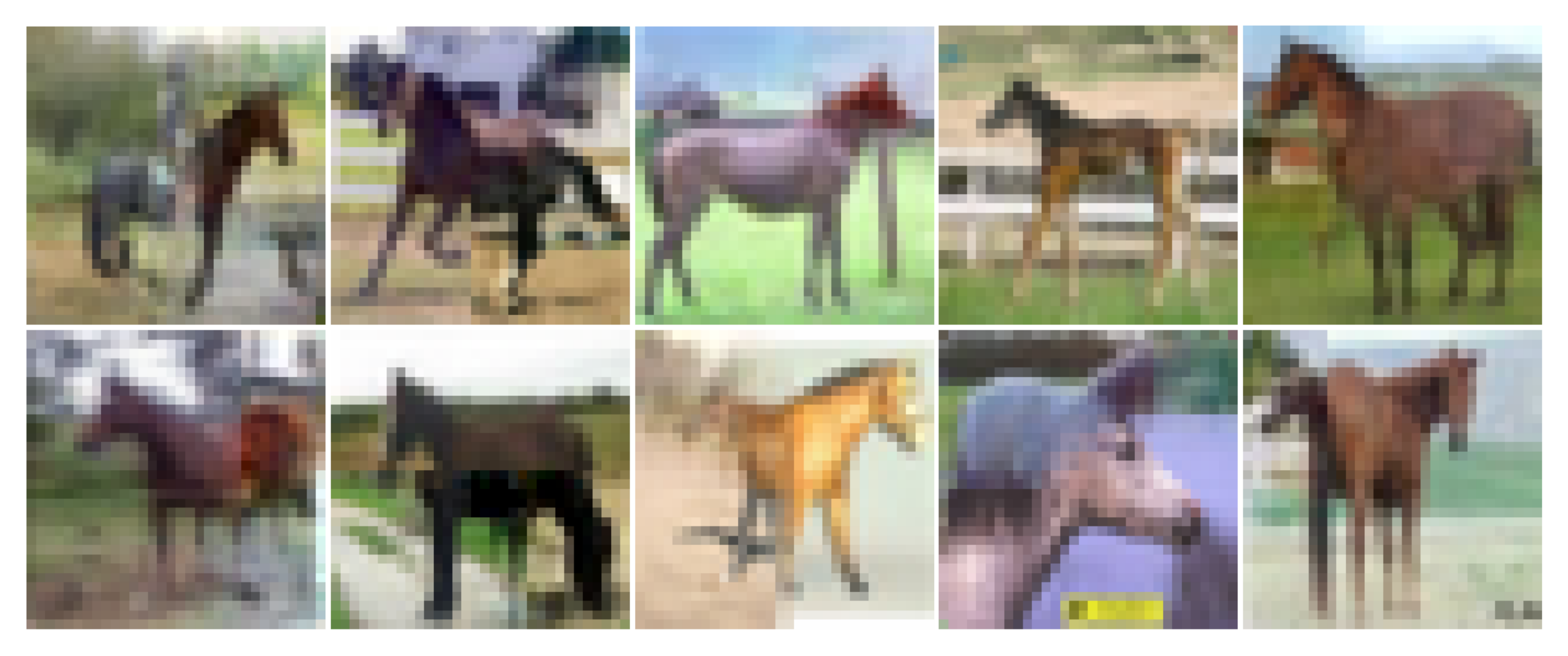}\includegraphics[width=4.2cm]{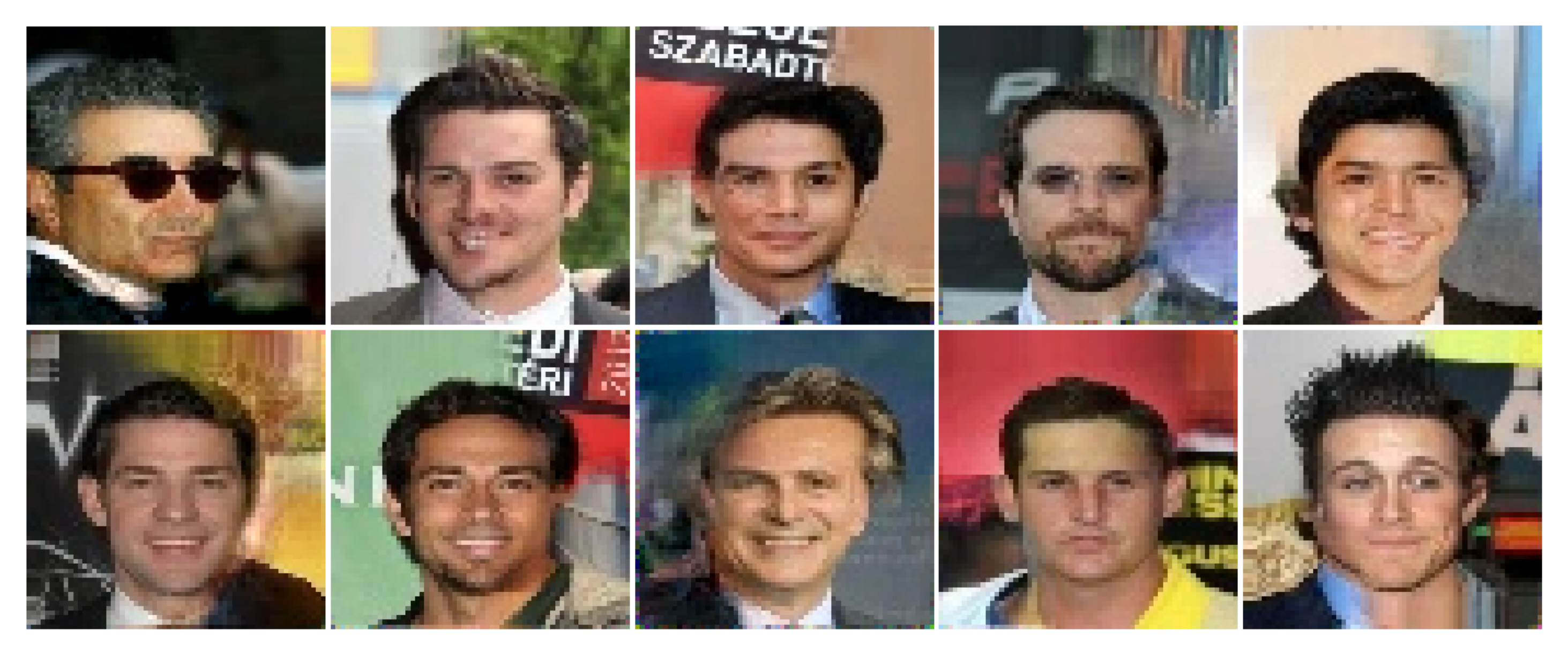}}\tabularnewline
\begin{turn}{90}
{\footnotesize{}}%
\begin{tabular}{c}
{\footnotesize{}(e)}\tabularnewline
{\footnotesize{}GACNTKg}\tabularnewline
\end{tabular}
\end{turn} & \includegraphics[width=4.2cm]{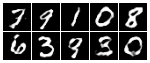}\includegraphics[width=4.2cm]{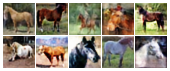}\includegraphics[width=4.2cm]{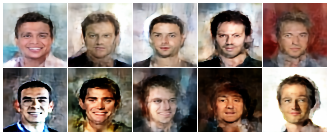}\tabularnewline
\end{tabular*}{\footnotesize\par}
\par\end{centering}
\caption{\label{fig:Demo}The images generated by different methods on MNIST,
CIFAR-10, and CelebA datasets given only 256 training images.}
\end{figure*}
\begin{figure}
\begin{centering}
{\footnotesize{}}%
\begin{tabular}{ccc}
{\footnotesize{}\includegraphics[viewport=0bp 0bp 272.6983bp 182bp,clip,width=4.2cm]{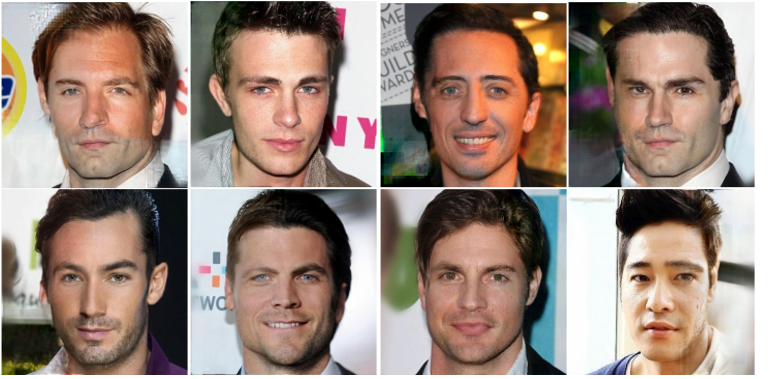}} & \hspace{2cm} & {\footnotesize{}\includegraphics[viewport=0bp 0bp 582bp 387.75bp,clip,width=4.2cm]{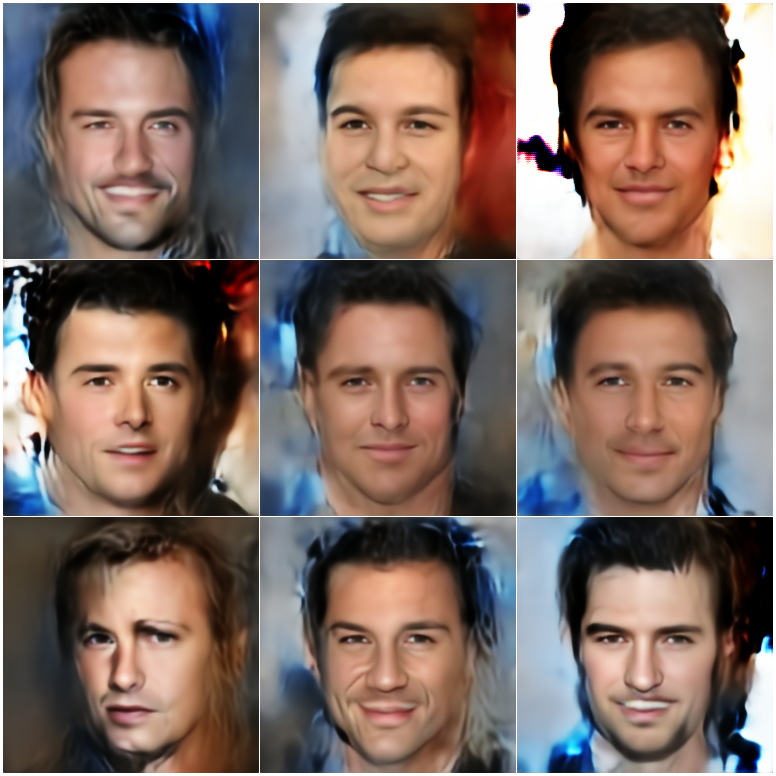}}\tabularnewline
{\footnotesize{}(a) GA-CNTK} & \hspace{2cm} & {\footnotesize{}(b) GA-CNTKg}\tabularnewline
\end{tabular}{\footnotesize\par}
\par\end{centering}
\caption{\label{fig:with-vs-without-generator}The images generated by GA-CNTK
(a) without and (b) with a generator given 256 CelebA-HQ training
images.}
\end{figure}

\subsection{Image Quality}

\label{subsec:Exp-image-quality}We first study the quality of the
images synthesized by different methods. Table \ref{tab:FID=000026SSIM}
summarizes the FID and AM-SSIM scores of the generated images. LSGAN
and DCGAN using $f$-divergence as the loss function give high FID
and fail to generate recognizable images on CIFAR-10 and CelebA datasets
due to the various training issues mentioned previously. StyleGAN,
although being able to generate impressive images with sufficient
training data, gives high FID here due to the high sample complexity
of the style-based generator. Other baselines, including WGAN, WGAN-GP,
and SN-GAN, can successfully generate recognizable images on all datasets,
as shown in Figure \ref{fig:Demo}. In particular, WGAN-GP performs
the best among the GAN variants. However, WGAN-GP limits the Lipschitz
continuity of the discriminator and gives higher FID scores than GA-CNTK.
Also, it gives higher AM-SSIM values as the size of the training set
decreases, implying there are many downgrade images that look identical
to some training images. This is because the Wasserstein distance,
which is also called the earth mover's distance, allows fewer ways
of moving when there are less available ``earth'' (i.e., the density
values of $\mathcal{P}_{\text{data}}$ and $\mathcal{P}_{\text{gen}}$)
due to a small $n$, and thus the $\mathcal{P}_{\text{gen}}$ needs
to be exactly the same as $\mathcal{P}_{\text{data}}$ to minimize
the distance.\footnote{The problem of Lipschitz continuity may be alleviated when $n$ becomes
larger.} The GA-NTK variants, including GA-CNTK and GA-CNTKg (``g'' means
``with generator''), perform relatively well due to their lower
sample complexity, which aligns with the previous observations \citep{Shankar2020kernelwithouttangent,Arora2020ntksmalldatset}
in different context.

Next, we compare the images generated by the multi-resolutional GA-CNTK
and GA-CNTKg (see Section \ref{subsec:GA-NTK-in-Practice}) on the
CelebA-HQ dataset. The multi-resolutional GA-CNTK employs 3 discriminators
working at 256$\times$256, 64$\times$64, and 16$\times$16 pixel
resolutions, respectively. Figure \ref{fig:with-vs-without-generator}
shows the results. We can see that the multi-resolutional GA-CNTK
(without a generator) gives better-looking images than GA-CNTKg (with
a generator) because learning a generator, which maps two spaces,
is essentially a harder problem than finding a set of plausible $\boldsymbol{z}$'s.
Although synthesizing data faster at inference time, a generator may
not be necessary to generate high-quality images under the adversarial
setting.
\begin{figure}
\begin{centering}
{\footnotesize{}}%
\begin{tabular}{c}
{\footnotesize{}\includegraphics[viewport=0bp 0bp 1160bp 267.5bp,clip,width=5.5in]{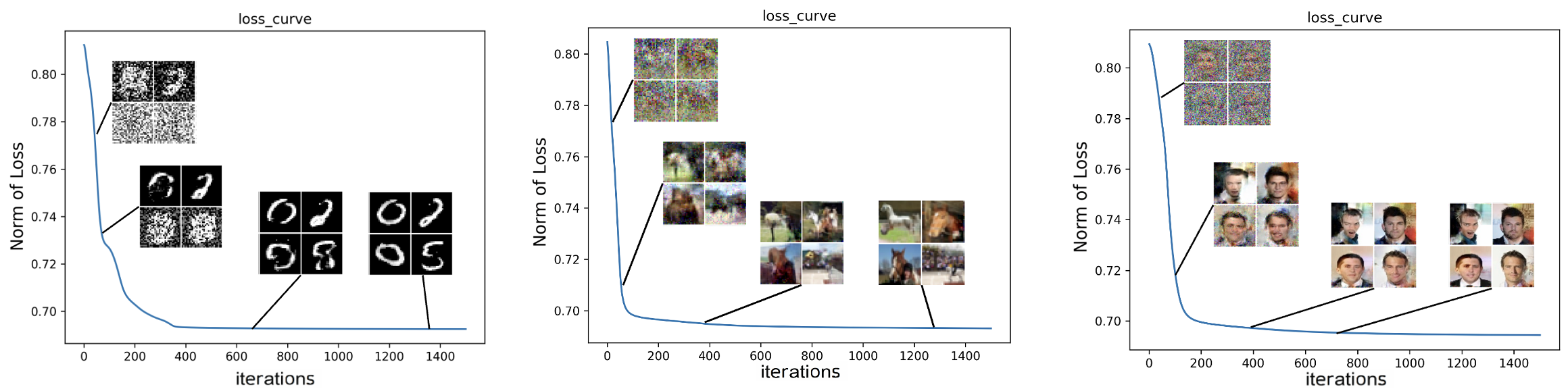}}\tabularnewline
{\footnotesize{}(a) MNIST\hspace{3.3cm}(b) CIFAR-10\hspace{3.3cm}(c)
CelebA}\tabularnewline
\end{tabular}{\footnotesize\par}
\par\end{centering}
\centering{}\caption{\label{fig:learning-curve}The learning curve and image quality at
different stages of a training process.}
\end{figure}
\begin{figure}
\noindent \centering{}{\footnotesize{}}%
\begin{tabular}{c}
{\footnotesize{}\includegraphics[width=1.99cm]{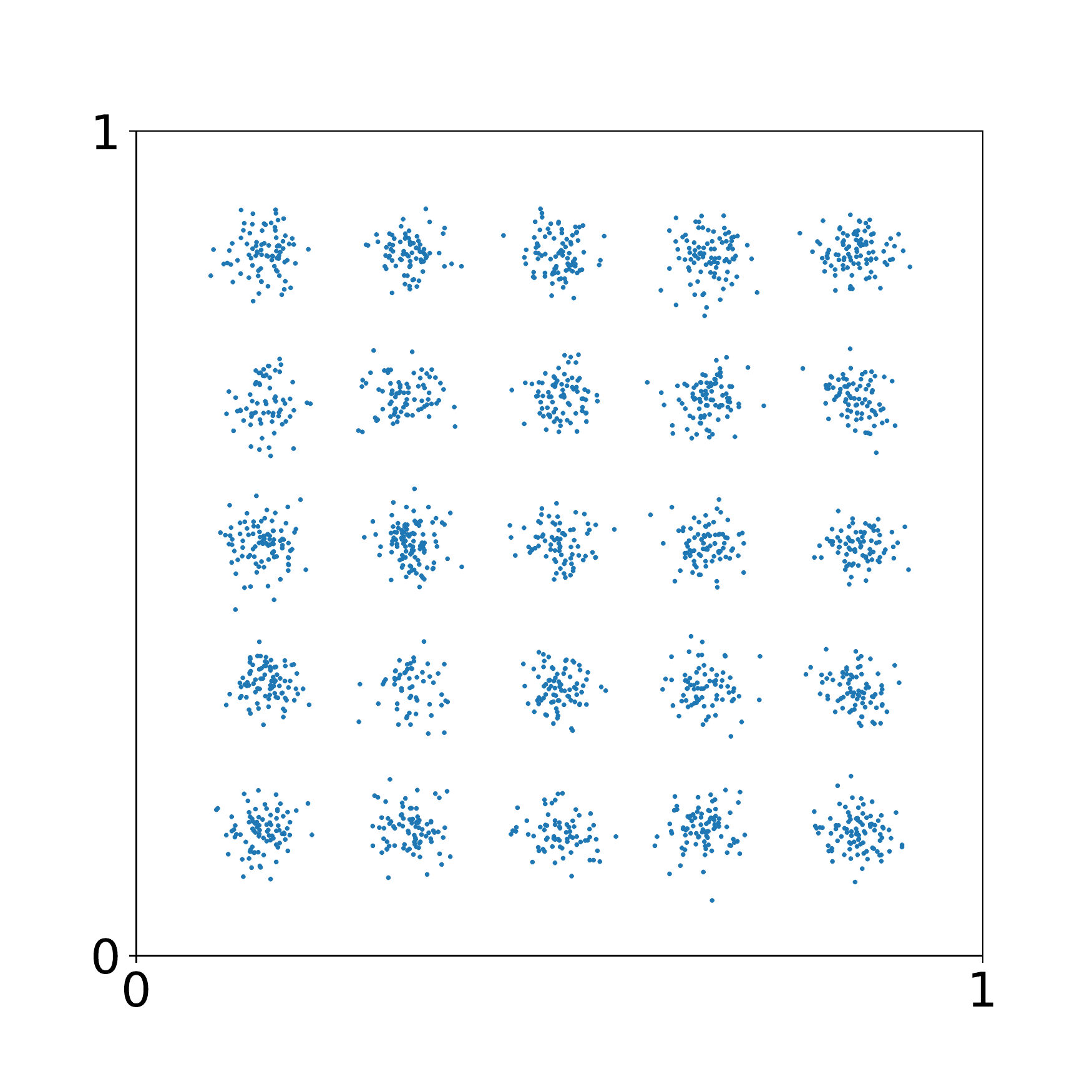}\includegraphics[width=1.99cm]{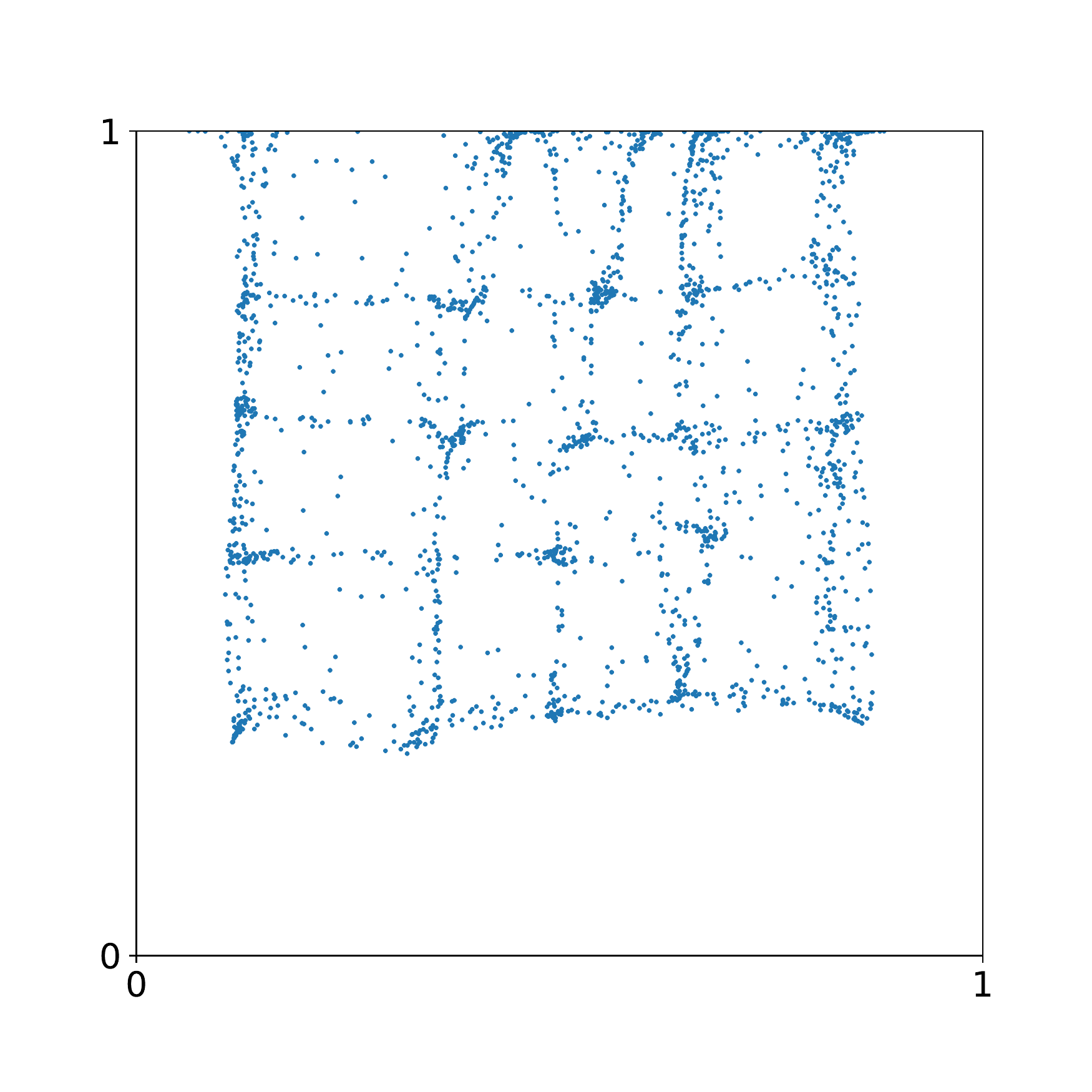}\includegraphics[width=1.99cm]{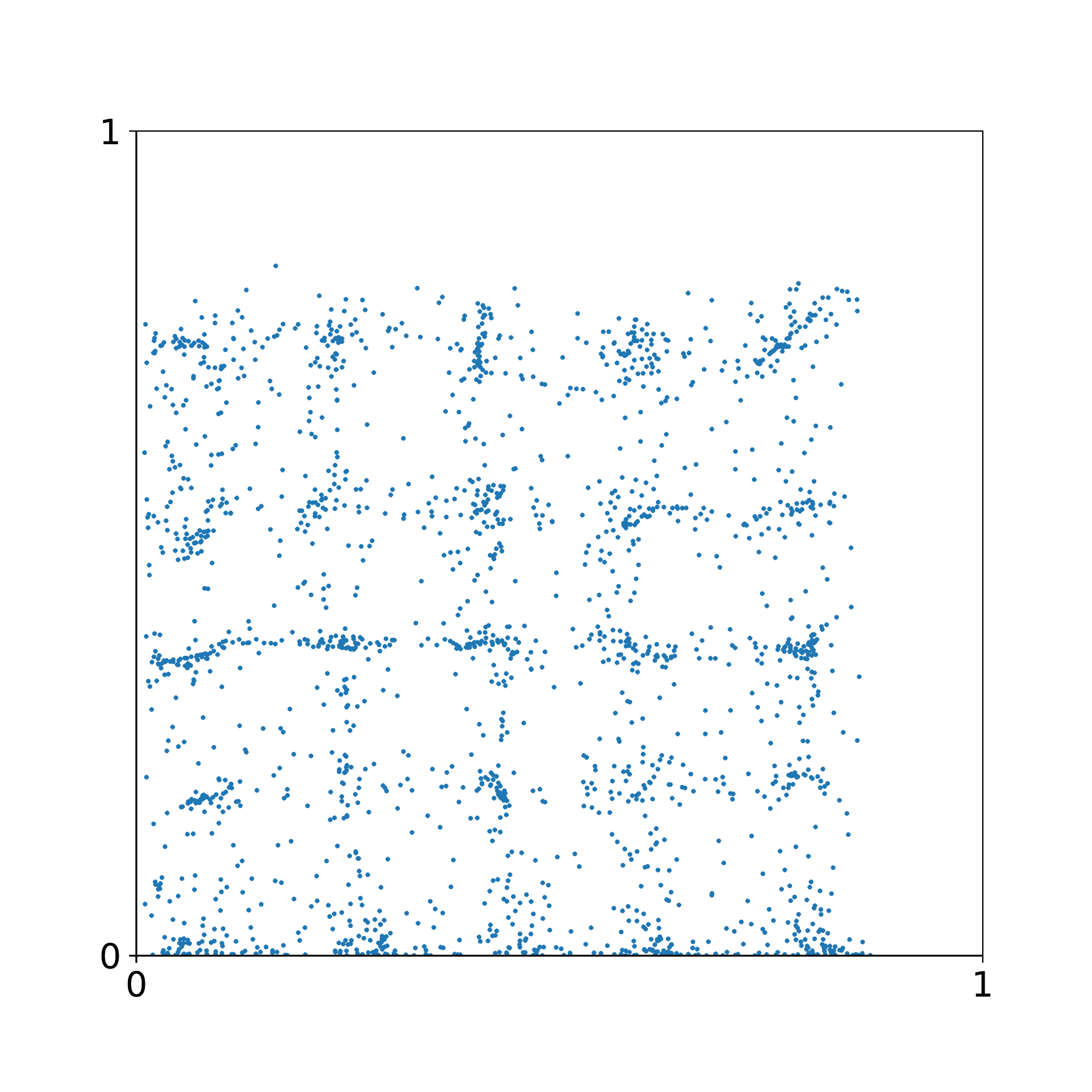}\includegraphics[width=1.99cm]{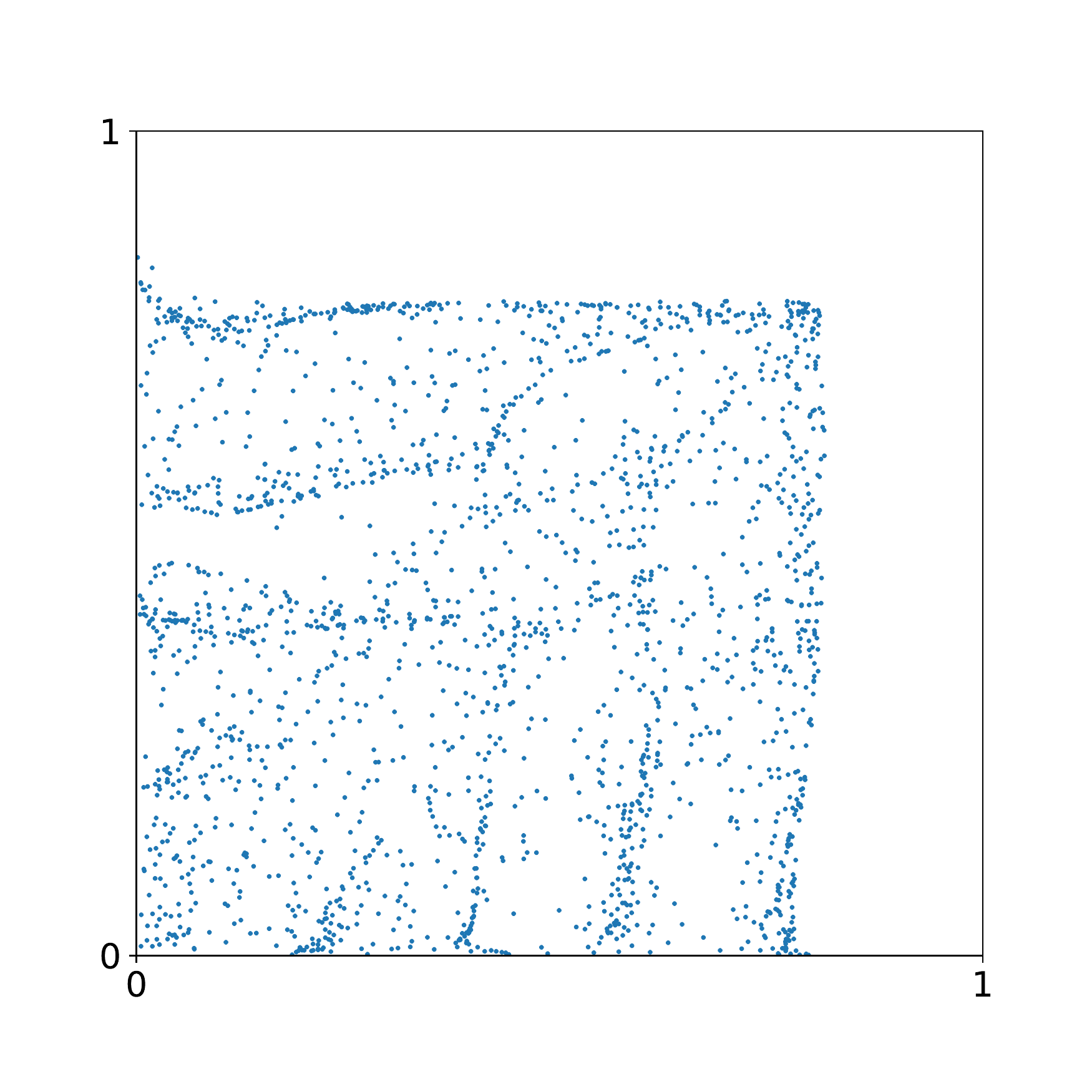}\includegraphics[width=1.99cm]{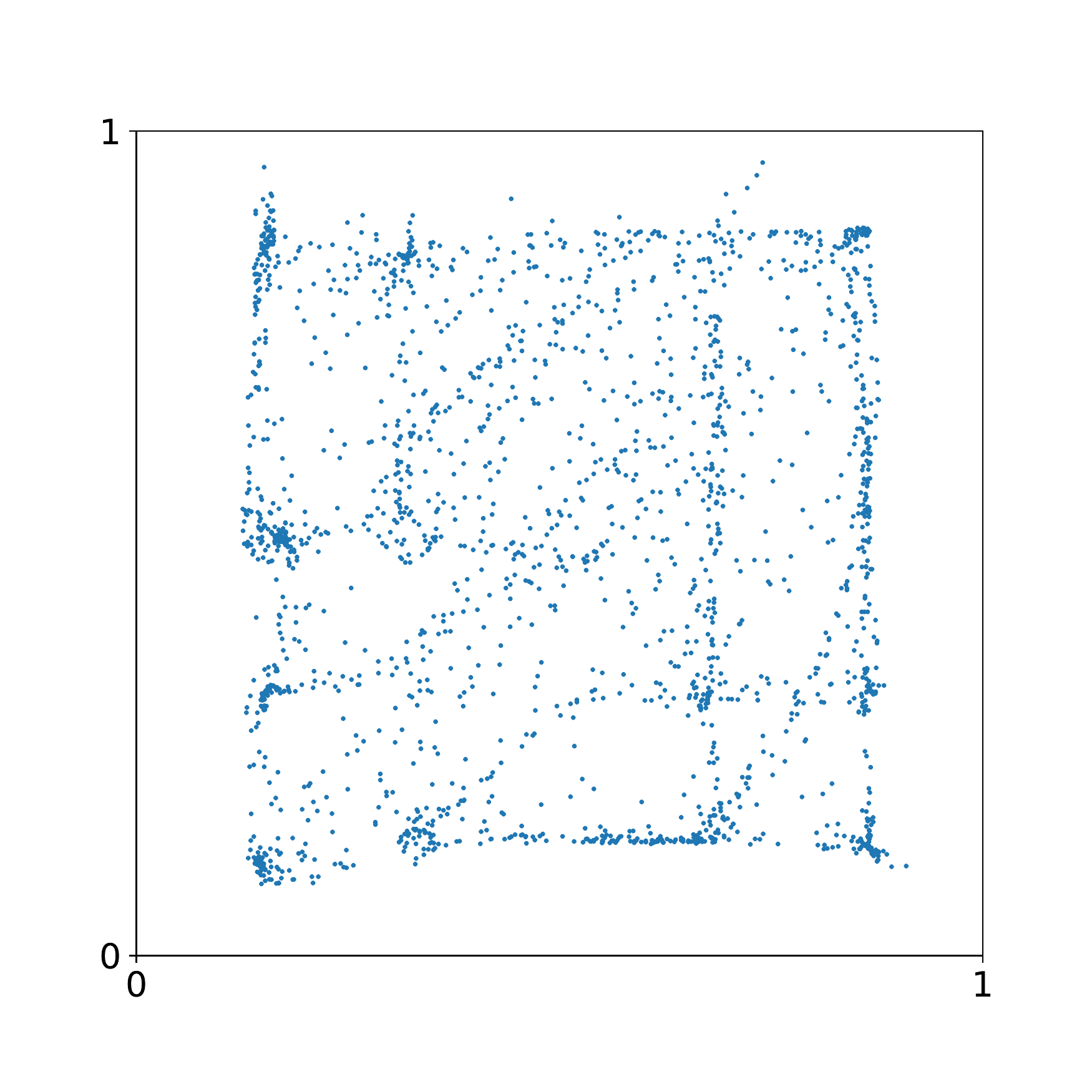}\includegraphics[width=1.99cm]{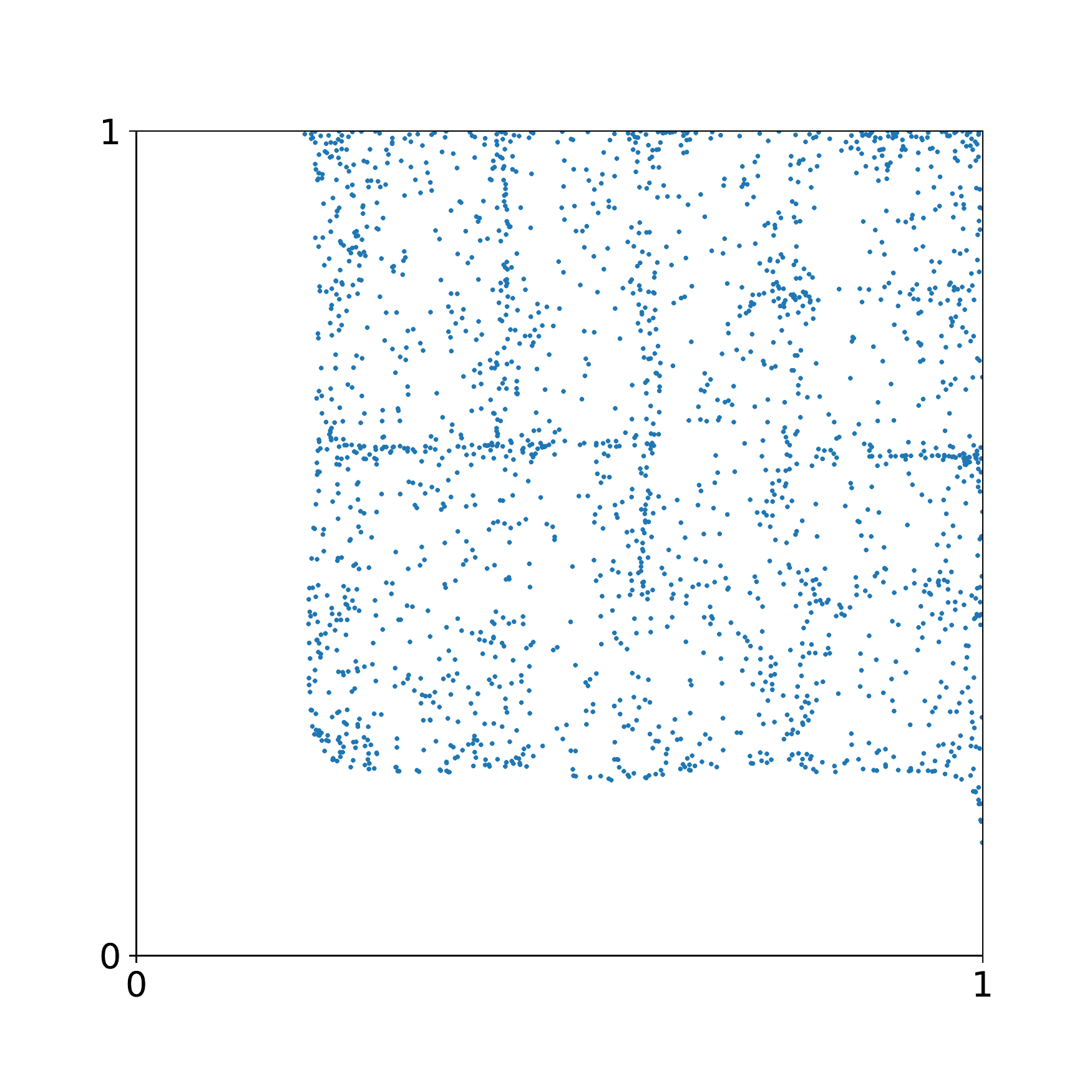}\includegraphics[width=1.99cm]{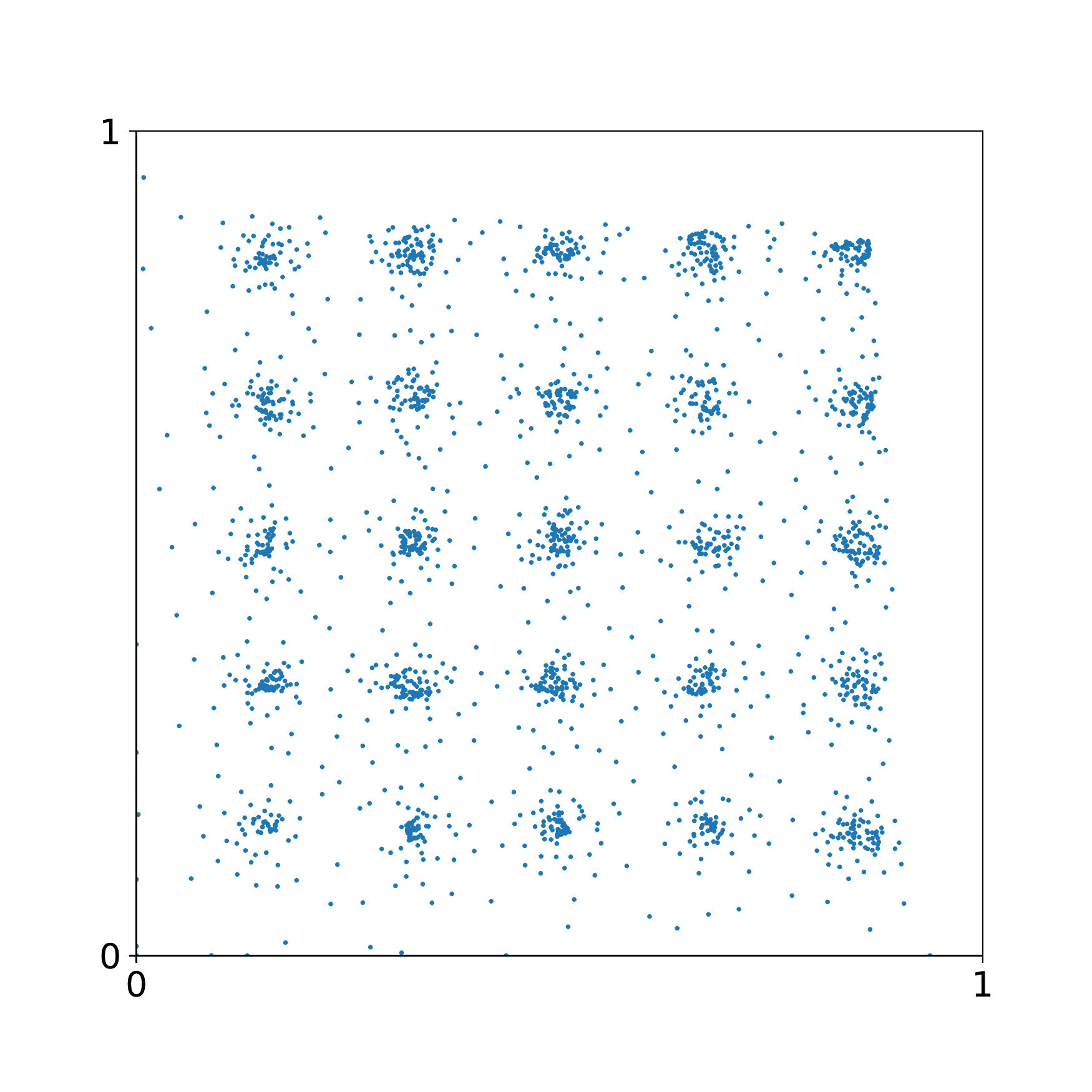}}\tabularnewline
{\footnotesize{}(a) Truth\hspace{0.4cm}(b) Vanilla GAN\hspace{0.38cm}(c)
LSGAN\hspace{0.3cm}(d) WGAN\hspace{0.4cm}(e) WGAN-GP\hspace{0.29cm}(f)
SN-GAN\hspace{0.29cm}(g) GA-FNTK}\tabularnewline
\end{tabular}\caption{\label{fig:2d-toy-dataset}Visualization of distribution alignment
and mode collapse on a 2D toy dataset.}
\end{figure}

\subsection{Training Stability}

\label{subsec:Exp-Training-Stability}\textbf{Convergence.} Figure
\ref{fig:learning-curve} shows the learning curve and the relationship
between the image quality and the number of gradient descent iterations
during a training process of GA-CNTK. We find that GA-CNTK easily
converges under various conditions, which is supported by Theorem
\ref{thm:convergence}. Furthermore, we can see a correlation between
the image quality and the loss value---as the loss becomes smaller,
the quality of the synthesized images improves. This correlation can
save human labor from monitoring the training processes, which is
common when training GANs. Note that the images generated in the latter
stage of training contain recognizable patterns that change over training
time. This is a major source of GA-CNTK creativity. Please see Section
\ref{subsec:evolution-of-images} for more discussions. \begin{wrapfigure}{o}{0.3\columnwidth}%
\begin{centering}
\includegraphics[width=4cm]{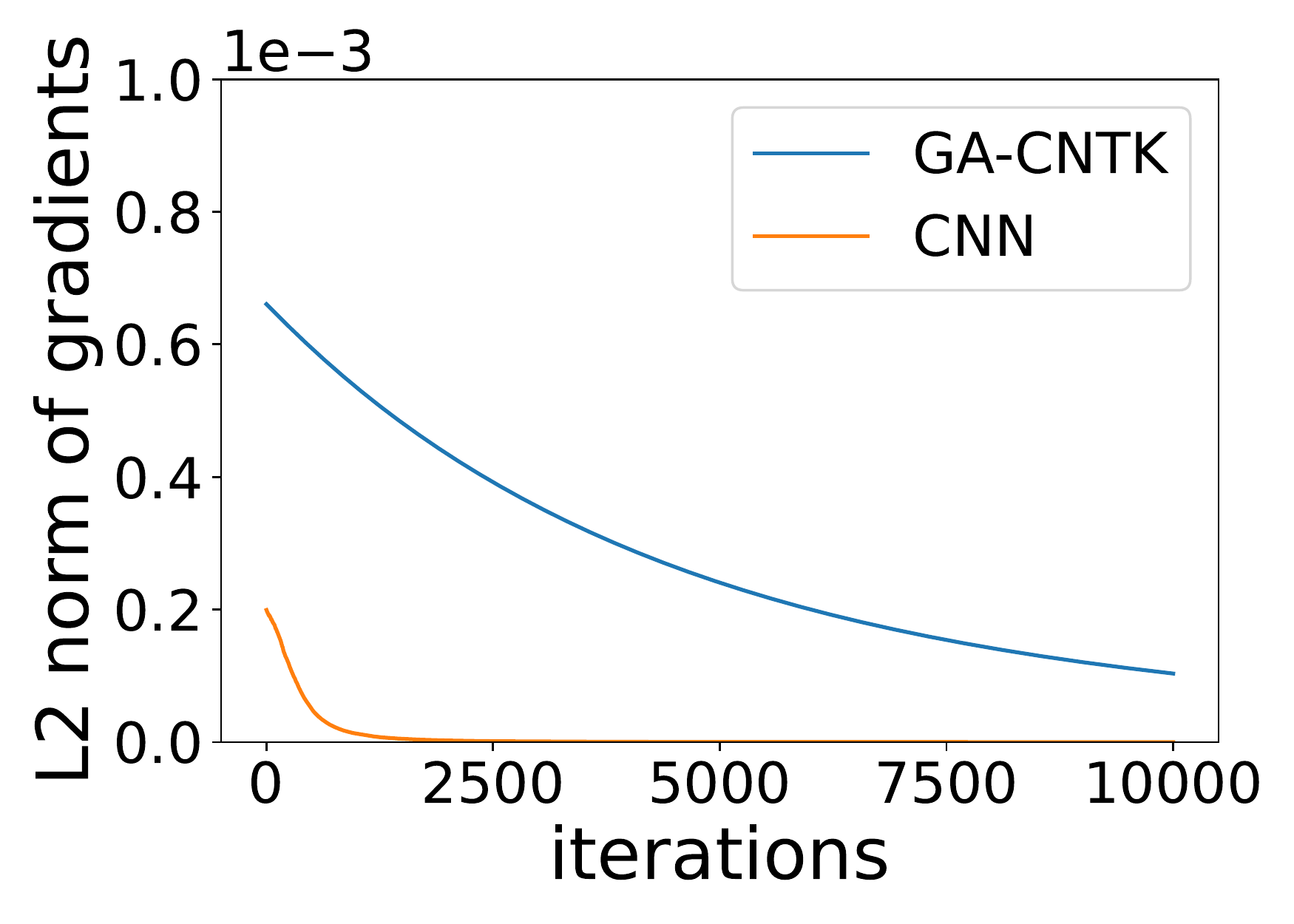}
\par\end{centering}
\caption{\label{fig:Comparison-of-gradients}Comparison between the gradients
of a $\boldsymbol{Z}_{i,:}$ in Eq. (\ref{eq:obj:ga-ntk}) obtained
from different types of $\mathcal{D}$.}
\end{wrapfigure}%
 \textbf{Mode collapse.} To study how different methods align $\mathcal{P}_{\text{gen}}$
with $\mathcal{P}_{\text{data}}$, we train them using a 2D toy training
set where $\mathcal{P}_{\text{data}}$ is a 25-modal Gaussian mixture.
We use two 3-layer fully-connected neural networks as the generator
and discriminator for each baseline and an ensemble of 3-layer, infinitely
wide counterpart as the discriminator in GA-FNTK. For GANs, we stop
the alternating SGD training when the generator receives 1000 updates,
and for GA-FNTK, we terminate the GD training after 1000 iterations.
Figure \ref{fig:2d-toy-dataset} shows the resultant $\mathcal{P}_{\text{gen}}$
of different methods.  GA-FNTK avoids mode collapse due to the use
of alternating SGD.\textbf{ Gradient vanishing.} To verify that GA-NTK
gives no vanishing gradients with a finite $\lambda$, we conduct
an experiment using another toy dataset consisting of 256 MNIST images
and 256 random noises. We replace the discriminator of GA-CNTK with
a single parametric network of the same architecture but finite width.
We train the finite-width network on the toy dataset by minimizing
the MSE loss using gradient descent. We set the training iteration
to a large value (65536) to simulate the situation where the network
value becomes saturated on both sides of the decision boundary. Figure
\ref{fig:Comparison-of-gradients} compares the gradients of a generated
image $\boldsymbol{Z}_{i,:}^{n}$ in Eq. (\ref{eq:obj:ga-ntk}) obtained
from 1) the finite-width network and 2) the corresponding GA-CNTK
with a large $t$. As $\boldsymbol{Z}_{i,:}^{n}$ evolves through
gradient descent iterations, the norm of its gradients obtained from
the finite-width discriminator quickly shrinks to zero. On the other
hand, the gradient norm obtained from the discriminator of GA-CNTK
is always positive thanks to the infinite ensembling.

\subsection{Scalability}

\begin{wraptable}{o}{0.4\columnwidth}%
\caption{\label{tab:scaled2000}The FID and AM-SSIM scores of the images output
by WGAN-GP and GA-CNTKg trained on 2048 CelebA images with batch size
256.}
\smallskip{}

\noindent \centering{}{\footnotesize{}}%
\begin{tabular}{cccc}
\hline 
{\scriptsize{}$n$=2048} & {\scriptsize{}Metric} & {\scriptsize{}WGANGP} & {\scriptsize{}GACNTKg}\tabularnewline
\hline 
\hline 
\multirow{2}{*}{{\scriptsize{}MNIST}} & {\scriptsize{}FID} & {\scriptsize{}23.47} & {\scriptsize{}56.73}\tabularnewline
\cline{2-4} \cline{3-4} \cline{4-4} 
 & {\scriptsize{}ASSIM} & {\scriptsize{}0.786} & {\scriptsize{}0.787}\tabularnewline
\hline 
\multirow{2}{*}{{\scriptsize{}CIFAR-10}} & {\scriptsize{}FID} & {\scriptsize{}110.70} & {\scriptsize{}78.85}\tabularnewline
\cline{2-4} \cline{3-4} \cline{4-4} 
 & {\scriptsize{}ASSIM} & {\scriptsize{}0.404} & {\scriptsize{}0.432}\tabularnewline
\hline 
\multirow{2}{*}{{\scriptsize{}CelebA}} & {\scriptsize{}FID} & {\scriptsize{}67.29} & {\scriptsize{}59.91}\tabularnewline
\cline{2-4} \cline{3-4} \cline{4-4} 
 & {\scriptsize{}ASSIM} & {\scriptsize{}0.337} & {\scriptsize{}0.411}\tabularnewline
\hline 
\end{tabular}{\footnotesize\par}\end{wraptable}%
Unlike GA-CNTK, the GA-CNTKg is batch-wise and thus can be trained
by more examples. Here, we scale up WGAN-GP and GA-CNTKg by training
them on CelebA dataset consisting of 2048 images. The batch size is
256. Table \ref{tab:scaled2000} summarizes the FID and AM-SSIM scores
of the generated images. On MNIST, WGAN-GP slightly outperforms GA-CNTKg.
The training of WGAN-GP on MNIST is easy, so GA-CNTKg does not offer
much advantage. However, in a more complex task like CIFAR-10 or CelebA,
GA-CNTKg outperforms WGAN-GP, suggesting that our single-level modeling
is indeed beneficial.

We have conducted more experiments. Please see Appendix for their
results. 

\section{Conclusion}

We proposed GA-NTK and showed that adversarial data synthesis can
be done via single-level modeling. It can be solved by ordinary gradient
descent, avoiding the difficulties of bi-level training of GANs. We
analyzed the convergence behavior of GA-NTK and gave sufficient conditions
for convergence. Extensive experiments were conducted to study the
advantages and limitations of GA-NTK. We proposed the batch-wise and
multi-resolutional variants to improve memory efficiency and image
quality, and showed that GA-NTK works either with or without a generator
network. GA-NTK works well with small data, making it suitable for
applications where data are hard to collect. GA-NTK also opens up
opportunities for one to adapt various GAN enhancements into the kernel
regime. These are matters of our future inquiry.

\bibliographystyle{iclr2023_conference}
\bibliography{gantk}

\begin{thebibliography}{73}
\providecommand{\natexlab}[1]{#1}
\providecommand{\url}[1]{\texttt{#1}}
\expandafter\ifx\csname urlstyle\endcsname\relax
  \providecommand{\doi}[1]{doi: #1}\else
  \providecommand{\doi}{doi: \begingroup \urlstyle{rm}\Url}\fi

\bibitem[Alemohammad et~al.(2021)Alemohammad, Wang, Balestriero, and
  Baraniuk]{alemohammad2021rntk}
Sina Alemohammad, Zichao Wang, Randall Balestriero, and Richard~G. Baraniuk.
\newblock The recurrent neural tangent kernel.
\newblock In \emph{Proc. of ICLR}, 2021.

\bibitem[Anil et~al.(2019)Anil, Lucas, and Grosse]{anil2019sorting}
Cem Anil, James Lucas, and Roger Grosse.
\newblock Sorting out lipschitz function approximation.
\newblock In \emph{Proc. of ICML}, 2019.

\bibitem[Arjovsky \& Bottou(2017)Arjovsky and Bottou]{arjovsky2017towards}
Martin Arjovsky and L{\'e}on Bottou.
\newblock Towards principled methods for training generative adversarial
  networks.
\newblock In \emph{Proc. of ICLR}, 2017.

\bibitem[Arjovsky et~al.(2017)Arjovsky, Chintala, and
  Bottou]{arjovsky2017wasserstein-gan}
Martin Arjovsky, Soumith Chintala, and L{\'e}on Bottou.
\newblock Wasserstein generative adversarial networks.
\newblock In \emph{Proc. of ICML}, 2017.

\bibitem[Arora et~al.(2019)Arora, Du, Hu, Li, Salakhutdinov, and
  Wang]{arora2019cntk}
Sanjeev Arora, Simon~S. Du, Wei Hu, Zhiyuan Li, Ruslan Salakhutdinov, and
  Ruosong Wang.
\newblock On exact computation with an infinitely wide neural net.
\newblock In \emph{Proc. of NeurIPS}, 2019.

\bibitem[Arora et~al.(2020)Arora, Du, Li, Salakhutdinov, Wang, and
  Yu]{Arora2020ntksmalldatset}
Sanjeev Arora, Simon~S. Du, Zhiyuan Li, Ruslan Salakhutdinov, Ruosong Wang, and
  Dingli Yu.
\newblock Harnessing the power of infinitely wide deep nets on small-data
  tasks.
\newblock In \emph{Proc. of ICLR}, 2020.

\bibitem[Bang \& Shim(2021)Bang and Shim]{bang2021mggan}
Duhyeon Bang and Hyunjung Shim.
\newblock Mggan: Solving mode collapse using manifold-guided training.
\newblock In \emph{Proc. of CVPR}, 2021.

\bibitem[Bergmann et~al.(2019)Bergmann, L{\"o}we, Fauser, Sattlegger, and
  Steger]{Bergmann2019ImprovingUD}
Paul Bergmann, Sindy L{\"o}we, Michael Fauser, David Sattlegger, and Carsten
  Steger.
\newblock Improving unsupervised defect segmentation by applying structural
  similarity to autoencoders.
\newblock In \emph{VISIGRAPP}, 2019.

\bibitem[Bietti \& Mairal(2019)Bietti and Mairal]{bietti19inductive-bais-ntk}
Alberto Bietti and Julien Mairal.
\newblock On the inductive bias of neural tangent kernels.
\newblock In \emph{Proc. of NeurIPS}, 2019.

\bibitem[Brock et~al.(2019)Brock, Donahue, and Simonyan]{brock2018large}
Andrew Brock, Jeff Donahue, and Karen Simonyan.
\newblock Large scale gan training for high fidelity natural image synthesis.
\newblock In \emph{Proc. of ICLR}, 2019.

\bibitem[Che et~al.(2017)Che, Li, Jacob, Bengio, and Li]{che2017modereg}
Tong Che, Yanran Li, Athul~Paul Jacob, Yoshua Bengio, and Wenjie Li.
\newblock Mode regularized generative adversarial networks.
\newblock In \emph{Proc. of ICLR}, 2017.

\bibitem[Chen et~al.(2016)Chen, Duan, Houthooft, Schulman, Sutskever, and
  Abbeel]{Chen16InfoGAN}
Xi~Chen, Yan Duan, Rein Houthooft, John Schulman, Ilya Sutskever, and Pieter
  Abbeel.
\newblock Infogan: Interpretable representation learning by information
  maximizing generative adversarial nets.
\newblock In \emph{Proc. of NeurIPS}, 2016.

\bibitem[Chizat et~al.(2019)Chizat, Oyallon, and Bach]{chizat2018ntk}
Lenaic Chizat, Edouard Oyallon, and Francis Bach.
\newblock On lazy training in differentiable programming.
\newblock In \emph{Proc. of NeurIPS}, 2019.

\bibitem[Daskalakis et~al.(2018)Daskalakis, Ilyas, Syrgkanis, and
  Zeng]{daskalakisISZ2018gans-optimism}
Constantinos Daskalakis, Andrew Ilyas, Vasilis Syrgkanis, and Haoyang Zeng.
\newblock Training gans with optimism.
\newblock In \emph{Proc. of ICLR}, 2018.

\bibitem[Deng et~al.(2009)Deng, Dong, Socher, Li, Li, and
  Fei-Fei]{deng2009imagenet}
Jia Deng, Wei Dong, Richard Socher, Li-Jia Li, Kai Li, and Li~Fei-Fei.
\newblock Imagenet: A large-scale hierarchical image database.
\newblock In \emph{Proc. of CVPR}, 2009.

\bibitem[Durugkar et~al.(2017)Durugkar, Gemp, and
  Mahadevan]{durugkar2017multigan}
Ishan~P. Durugkar, Ian Gemp, and Sridhar Mahadevan.
\newblock Generative multi-adversarial networks.
\newblock In \emph{Proc. of ICLR}, 2017.

\bibitem[Farnia \& Ozdaglar(2020)Farnia and
  Ozdaglar]{farnia2020nash-equilibria}
Farzan Farnia and Asuman~E. Ozdaglar.
\newblock Do gans always have nash equilibria?
\newblock In \emph{Proc. of ICML}, 2020.

\bibitem[Franceschi et~al.(2021)Franceschi, de~B{\'{e}}zenac, Ayed, Chen,
  Lamprier, and Gallinari]{franceschi2021ntk-perspetive-of-gans}
Jean{-}Yves Franceschi, Emmanuel de~B{\'{e}}zenac, Ibrahim Ayed, Micka{\"{e}}l
  Chen, Sylvain Lamprier, and Patrick Gallinari.
\newblock A neural tangent kernel perspective of gans.
\newblock \emph{CoRR}, abs/2106.05566, 2021.

\bibitem[Garriga{-}Alonso et~al.(2019)Garriga{-}Alonso, Rasmussen, and
  Aitchison]{Garriga-Alonso2019cnngp}
Adri{\`{a}} Garriga{-}Alonso, Carl~Edward Rasmussen, and Laurence Aitchison.
\newblock Deep convolutional networks as shallow gaussian processes.
\newblock In \emph{Proc. of ICLR}, 2019.

\bibitem[Geifman et~al.(2020)Geifman, Yadav, Kasten, Galun, Jacobs, and
  Ronen]{geifman2020similarity}
Amnon Geifman, Abhay Yadav, Yoni Kasten, Meirav Galun, David Jacobs, and Basri
  Ronen.
\newblock On the similarity between the laplace and neural tangent kernels.
\newblock In \emph{Proc. of NeurIPS}, 2020.

\bibitem[Ghosh et~al.(2018)Ghosh, Kulharia, Namboodiri, Torr, and
  Dokania]{ghosh18multiagentgan}
Arnab Ghosh, Viveka Kulharia, Vinay~P. Namboodiri, Philip H.~S. Torr, and
  Puneet~Kumar Dokania.
\newblock Multi-agent diverse generative adversarial networks.
\newblock In \emph{Proc. of CVPR}, 2018.

\bibitem[Goodfellow(2016)]{goodfellow2016nips}
Ian Goodfellow.
\newblock Nips 2016 tutorial: Generative adversarial networks.
\newblock \emph{arXiv preprint arXiv:1701.00160}, 2016.

\bibitem[Goodfellow et~al.(2014)Goodfellow, Pouget-Abadie, Mirza, Xu,
  Warde-Farley, Ozair, Courville, and Bengio]{goodfellow2014gan}
Ian Goodfellow, Jean Pouget-Abadie, Mehdi Mirza, Bing Xu, David Warde-Farley,
  Sherjil Ozair, Aaron Courville, and Yoshua Bengio.
\newblock Generative adversarial nets.
\newblock In \emph{Proc. of NeurIPS}, 2014.

\bibitem[Goodfellow et~al.(2015)Goodfellow, Shlens, and
  Szegedy]{goodfellow2014explaining}
Ian~J Goodfellow, Jonathon Shlens, and Christian Szegedy.
\newblock Explaining and harnessing adversarial examples.
\newblock In \emph{Proc. of ICLR}, 2015.

\bibitem[Gower(2022)]{gower2018convergence}
Robert~M Gower.
\newblock Convergence theorems for gradient descent, May 2022.
\newblock
  \url{https://gowerrobert.github.io/pdf/M2_statistique_optimisation/grad_conv.pdf}.

\bibitem[Gretton et~al.(2012)Gretton, Borgwardt, Rasch, Sch{\"{o}}lkopf, and
  Smola]{gretton2012mmd}
Arthur Gretton, Karsten~M. Borgwardt, Malte~J. Rasch, Bernhard Sch{\"{o}}lkopf,
  and Alexander~J. Smola.
\newblock A kernel two-sample test.
\newblock \emph{J. Mach. Learn. Res.}, 2012.

\bibitem[Gulrajani et~al.(2017)Gulrajani, Ahmed, Arjovsky, Dumoulin, and
  Courville]{ishaan2017improved-wgan}
Ishaan Gulrajani, Faruk Ahmed, Mart{\'{\i}}n Arjovsky, Vincent Dumoulin, and
  Aaron~C. Courville.
\newblock Improved training of wasserstein gans.
\newblock In \emph{Proc. of NeurIPS}, 2017.

\bibitem[Han et~al.(2021)Han, Avron, Shoham, Kim, and
  Shin]{han21random-feature-ntk}
Insu Han, Haim Avron, Neta Shoham, Chaewon Kim, and Jinwoo Shin.
\newblock Random features for the neural tangent kernel.
\newblock \emph{CoRR}, abs/2104.01351, 2021.

\bibitem[Heusel et~al.(2017)Heusel, Ramsauer, Unterthiner, Nessler, and
  Hochreiter]{Heusel17FID}
Martin Heusel, Hubert Ramsauer, Thomas Unterthiner, Bernhard Nessler, and Sepp
  Hochreiter.
\newblock Gans trained by a two time-scale update rule converge to a local nash
  equilibrium.
\newblock In \emph{Proc. of NeurIPS}, 2017.

\bibitem[Hron et~al.(2020)Hron, Bahri, Sohl{-}Dickstein, and
  Novak]{hron2020infinite_attention}
Jiri Hron, Yasaman Bahri, Jascha Sohl{-}Dickstein, and Roman Novak.
\newblock Infinite attention: {NNGP} and {NTK} for deep attention networks.
\newblock In \emph{Proc. of ICML}, 2020.

\bibitem[Ilyas et~al.(2019)Ilyas, Santurkar, Tsipras, Engstrom, Tran, and
  Madry]{ilyas2019adversarial}
Andrew Ilyas, Shibani Santurkar, Dimitris Tsipras, Logan Engstrom, Brandon
  Tran, and Aleksander Madry.
\newblock Adversarial examples are not bugs, they are features.
\newblock In \emph{Proc. of NeurIPS}, 2019.

\bibitem[Jacot et~al.(2018)Jacot, Gabriel, and Hongler]{jacot2018ntk}
Arthur Jacot, Franck Gabriel, and Clement Hongler.
\newblock Neural tangent kernel: Convergence and generalization in neural
  networks.
\newblock In \emph{Proc. of NeurIPS}, 2018.

\bibitem[Jeffreys(1946)]{jeffreys1946invariant}
Harold Jeffreys.
\newblock An invariant form for the prior probability in estimation problems.
\newblock \emph{Proc. of the Royal Society of London. Series A. Mathematical
  and Physical Sciences}, 186\penalty0 (1007):\penalty0 453--461, 1946.

\bibitem[Karras et~al.(2020)Karras, Laine, Aittala, Hellsten, Lehtinen, and
  Aila]{karras2020analyzing}
Tero Karras, Samuli Laine, Miika Aittala, Janne Hellsten, Jaakko Lehtinen, and
  Timo Aila.
\newblock Analyzing and improving the image quality of stylegan.
\newblock In \emph{Proc. of CVPR}, 2020.

\bibitem[Krizhevsky(2009)]{Krizhevsky09learningmultiple}
Alex Krizhevsky.
\newblock Learning multiple layers of features from tiny images.
\newblock Technical report, 2009.

\bibitem[LeCun et~al.(2010)LeCun, Cortes, and Burges]{lecun2010mnist}
Yann LeCun, Corinna Cortes, and CJ~Burges.
\newblock Mnist handwritten digit database.
\newblock \emph{ATT Labs [Online]. Available:
  http://yann.lecun.com/exdb/mnist}, 2, 2010.

\bibitem[Ledig et~al.(2017)Ledig, Theis, Huszar, Caballero, Cunningham, Acosta,
  Aitken, Tejani, Totz, Wang, and Shi]{ledig2017super-resolution}
Christian Ledig, Lucas Theis, Ferenc Huszar, Jose Caballero, Andrew Cunningham,
  Alejandro Acosta, Andrew~P. Aitken, Alykhan Tejani, Johannes Totz, Zehan
  Wang, and Wenzhe Shi.
\newblock Photo-realistic single image super-resolution using a generative
  adversarial network.
\newblock In \emph{Proc. of CVPR}, 2017.

\bibitem[Lee et~al.(2018)Lee, Bahri, Novak, Schoenholz, Pennington, and
  Sohl-Dickstein]{lee2017nngp}
Jaehoon Lee, Yasaman Bahri, Roman Novak, Samuel~S Schoenholz, Jeffrey
  Pennington, and Jascha Sohl-Dickstein.
\newblock Deep neural networks as gaussian processes.
\newblock In \emph{Proc. of ICLR}, 2018.

\bibitem[Lee et~al.(2019)Lee, Xiao, Schoenholz, Bahri, Novak, Sohl-Dickstein,
  and Pennington]{lee2019ntk}
Jaehoon Lee, Lechao Xiao, Samuel Schoenholz, Yasaman Bahri, Roman Novak, Jascha
  Sohl-Dickstein, and Jeffrey Pennington.
\newblock Wide neural networks of any depth evolve as linear models under
  gradient descent.
\newblock In \emph{Proc. of NeurIPS}, 2019.

\bibitem[Lee et~al.(2020)Lee, Schoenholz, Pennington, Adlam, Xiao, Novak, and
  Sohl{-}Dickstein]{lee2020fin-vs-inf}
Jaehoon Lee, Samuel~S. Schoenholz, Jeffrey Pennington, Ben Adlam, Lechao Xiao,
  Roman Novak, and Jascha Sohl{-}Dickstein.
\newblock Finite versus infinite neural networks: an empirical study.
\newblock In \emph{Proc. of NeurIPS}, 2020.

\bibitem[Li et~al.(2017)Li, Chang, Cheng, Yang, and
  P{\'{o}}czos]{Li2017mmd-gan}
Chun{-}Liang Li, Wei{-}Cheng Chang, Yu~Cheng, Yiming Yang, and Barnab{\'{a}}s
  P{\'{o}}czos.
\newblock {MMD} {GAN:} towards deeper understanding of moment matching network.
\newblock In \emph{Proc. of NeurIPS}, 2017.

\bibitem[Li et~al.(2021)Li, Fan, Wang, Ma, and Cui]{li2021tackling}
Wei Li, Li~Fan, Zhenyu Wang, Chao Ma, and Xiaohui Cui.
\newblock Tackling mode collapse in multi-generator gans with orthogonal
  vectors.
\newblock \emph{Pattern Recognition}, 2021.

\bibitem[Li et~al.(2015)Li, Swersky, and Zemel]{li2015gmmn}
Yujia Li, Kevin Swersky, and Richard~S. Zemel.
\newblock Generative moment matching networks.
\newblock In \emph{Proc. of ICML}, 2015.

\bibitem[Liu et~al.(2015)Liu, Luo, Wang, and Tang]{liu2015faceattributes}
Ziwei Liu, Ping Luo, Xiaogang Wang, and Xiaoou Tang.
\newblock Deep learning face attributes in the wild.
\newblock In \emph{Proc. of ICCV}, December 2015.

\bibitem[Lucic et~al.(2018)Lucic, Kurach, Michalski, Gelly, and
  Bousquet]{lucic2018are-gans-created-equal}
Mario Lucic, Karol Kurach, Marcin Michalski, Sylvain Gelly, and Olivier
  Bousquet.
\newblock Are gans created equal? {A} large-scale study.
\newblock In \emph{Proc. of NeurIPS}, 2018.

\bibitem[Mao et~al.(2019)Mao, Lee, Tseng, Ma, and Yang]{mao2019mode}
Qi~Mao, Hsin-Ying Lee, Hung-Yu Tseng, Siwei Ma, and Ming-Hsuan Yang.
\newblock Mode seeking generative adversarial networks for diverse image
  synthesis.
\newblock In \emph{Proc. of CVPR}, 2019.

\bibitem[Mao et~al.(2017)Mao, Li, Xie, Lau, Wang, and Smolley]{mao2017lsgan}
Xudong Mao, Qing Li, Haoran Xie, Raymond Y.~K. Lau, Zhen Wang, and Stephen~Paul
  Smolley.
\newblock Least squares generative adversarial networks.
\newblock In \emph{Proc. of ICCV}, 2017.

\bibitem[Matthews et~al.(2018)Matthews, Hron, Rowland, Turner, and
  Ghahramani]{matthews2018gaussian}
Alexander G de~G Matthews, Jiri Hron, Mark Rowland, Richard~E Turner, and
  Zoubin Ghahramani.
\newblock Gaussian process behaviour in wide deep neural networks.
\newblock In \emph{Proc. of ICLR}, 2018.

\bibitem[Mescheder et~al.(2018)Mescheder, Geiger, and
  Nowozin]{meschederGN18which-gan-do-actually-converge}
Lars~M. Mescheder, Andreas Geiger, and Sebastian Nowozin.
\newblock Which training methods for gans do actually converge?
\newblock In \emph{Proc. of ICML}, 2018.

\bibitem[Metz et~al.(2017)Metz, Poole, Pfau, and
  Sohl{-}Dickstein]{luke2017unroll-gan}
Luke Metz, Ben Poole, David Pfau, and Jascha Sohl{-}Dickstein.
\newblock Unrolled generative adversarial networks.
\newblock In \emph{Proc. of ICLR}, 2017.

\bibitem[Miyato et~al.(2018)Miyato, Kataoka, Koyama, and
  Yoshida]{miyato2018spec-norm}
Takeru Miyato, Toshiki Kataoka, Masanori Koyama, and Yuichi Yoshida.
\newblock Spectral normalization for generative adversarial networks.
\newblock In \emph{Proc. of ICLR}, 2018.

\bibitem[Mokhtari et~al.(2020)Mokhtari, Ozdaglar, and
  Pattathil]{mokhtari2020minimax-eg-ogda}
Aryan Mokhtari, Asuman~E. Ozdaglar, and Sarath Pattathil.
\newblock A unified analysis of extra-gradient and optimistic gradient methods
  for saddle point problems: Proximal point approach.
\newblock In \emph{Proc. of AISTATS}, 2020.

\bibitem[Nagarajan \& Kolter(2017)Nagarajan and
  Kolter]{nagarajan2017gan-gd-locally-stable}
Vaishnavh Nagarajan and J.~Zico Kolter.
\newblock Gradient descent {GAN} optimization is locally stable.
\newblock In \emph{Proc. of NeurIPS}, 2017.

\bibitem[Novak et~al.(2019{\natexlab{a}})Novak, Xiao, Hron, Lee, Alemi,
  Sohl-Dickstein, and Schoenholz]{novak2019neural-tangents}
Roman Novak, Lechao Xiao, Jiri Hron, Jaehoon Lee, Alexander~A Alemi, Jascha
  Sohl-Dickstein, and Samuel~S Schoenholz.
\newblock Neural tangents: Fast and easy infinite neural networks in python.
\newblock In \emph{Proc. of ICLR}, 2019{\natexlab{a}}.

\bibitem[Novak et~al.(2019{\natexlab{b}})Novak, Xiao, Lee, Bahri, Yang, Hron,
  Abolafia, Pennington, and Sohl-Dickstein]{novak2020bayesian}
Roman Novak, Lechao Xiao, Jaehoon Lee, Yasaman Bahri, Greg Yang, Jiri Hron,
  Daniel~A. Abolafia, Jeffrey Pennington, and Jascha Sohl-Dickstein.
\newblock Bayesian deep convolutional networks with many channels are gaussian
  processes.
\newblock In \emph{Proc. of ICLR}, 2019{\natexlab{b}}.

\bibitem[Poole et~al.(2016)Poole, Lahiri, Raghu, Sohl{-}Dickstein, and
  Ganguli]{Poole16transient-chaos}
Ben Poole, Subhaneil Lahiri, Maithra Raghu, Jascha Sohl{-}Dickstein, and Surya
  Ganguli.
\newblock Exponential expressivity in deep neural networks through transient
  chaos.
\newblock In \emph{Proc. of NeurIPS}, 2016.

\bibitem[Qi(2020)]{qi2020lipschitz-densities}
Guo{-}Jun Qi.
\newblock Loss-sensitive generative adversarial networks on lipschitz
  densities.
\newblock \emph{Int. J. Comput. Vis.}, 2020.

\bibitem[Radford et~al.(2016)Radford, Metz, and Chintala]{Radford2015dcgan}
Alec Radford, Luke Metz, and Soumith Chintala.
\newblock Unsupervised representation learning with deep convolutional
  generative adversarial networks.
\newblock In \emph{Proc. of ICLR}, 2016.

\bibitem[Raghu et~al.(2017)Raghu, Poole, Kleinberg, Ganguli, and
  Sohl{-}Dickstein]{raghus17exp-of-dnn}
Maithra Raghu, Ben Poole, Jon~M. Kleinberg, Surya Ganguli, and Jascha
  Sohl{-}Dickstein.
\newblock On the expressive power of deep neural networks.
\newblock In \emph{Proc. of ICML}, 2017.

\bibitem[R{\'e}nyi et~al.(1961)]{renyi1961measures}
Alfr{\'e}d R{\'e}nyi et~al.
\newblock On measures of entropy and information.
\newblock In \emph{Proc. of the 4th Berkeley symposium on mathematical
  statistics and probability}, volume~1, 1961.

\bibitem[Sajjadi et~al.(2018)Sajjadi, Parascandolo, Mehrjou, and
  Sch{\"{o}}lkopf]{sajjadi18tempered-gan}
Mehdi S.~M. Sajjadi, Giambattista Parascandolo, Arash Mehrjou, and Bernhard
  Sch{\"{o}}lkopf.
\newblock Tempered adversarial networks.
\newblock In \emph{Proc. of ICML}, 2018.

\bibitem[Salimans et~al.(2016)Salimans, Goodfellow, Zaremba, Cheung, Radford,
  and Chen]{tim2016training-gan}
Tim Salimans, Ian~J. Goodfellow, Wojciech Zaremba, Vicki Cheung, Alec Radford,
  and Xi~Chen.
\newblock Improved techniques for training gans.
\newblock In \emph{Proc. of NeurIPS}, 2016.

\bibitem[Schoenholz et~al.(2017)Schoenholz, Gilmer, Ganguli, and
  Sohl{-}Dickstein]{schoenholz17deep-information-prop}
Samuel~S. Schoenholz, Justin Gilmer, Surya Ganguli, and Jascha
  Sohl{-}Dickstein.
\newblock Deep information propagation.
\newblock In \emph{Proc. of ICLR}, 2017.

\bibitem[Shankar et~al.(2020)Shankar, Fang, Guo, Fridovich{-}Keil,
  Ragan{-}Kelley, Schmidt, and Recht]{Shankar2020kernelwithouttangent}
Vaishaal Shankar, Alex Fang, Wenshuo Guo, Sara Fridovich{-}Keil, Jonathan
  Ragan{-}Kelley, Ludwig Schmidt, and Benjamin Recht.
\newblock Neural kernels without tangents.
\newblock In \emph{Proc. of ICML}, 2020.

\bibitem[Srivastava et~al.(2017)Srivastava, Valkov, Russell, Gutmann, and
  Sutton]{srivastava2017veegan}
Akash Srivastava, Lazar Valkov, Chris Russell, Michael~U. Gutmann, and Charles
  Sutton.
\newblock {VEEGAN:} reducing mode collapse in gans using implicit variational
  learning.
\newblock In \emph{Proc. of NeurIPS}, 2017.

\bibitem[Thekumparampil et~al.(2019)Thekumparampil, Jain, Netrapalli, and
  Oh]{kumparampi2019smooth-minimax}
Kiran~Koshy Thekumparampil, Prateek Jain, Praneeth Netrapalli, and Sewoong Oh.
\newblock Efficient algorithms for smooth minimax optimization.
\newblock In \emph{Proc. of NeurIPS}, 2019.

\bibitem[Vondrick et~al.(2016)Vondrick, Pirsiavash, and
  Torralba]{vondrick16vgan}
Carl Vondrick, Hamed Pirsiavash, and Antonio Torralba.
\newblock Generating videos with scene dynamics.
\newblock In \emph{Proc. of NeurIPS}, 2016.

\bibitem[Wang et~al.(2018)Wang, Liu, Zhu, Tao, Kautz, and
  Catanzaro]{wang2018high}
Ting-Chun Wang, Ming-Yu Liu, Jun-Yan Zhu, Andrew Tao, Jan Kautz, and Bryan
  Catanzaro.
\newblock High-resolution image synthesis and semantic manipulation with
  conditional gans.
\newblock In \emph{Proceedings of the IEEE conference on computer vision and
  pattern recognition}, pp.\  8798--8807, 2018.

\bibitem[Wang et~al.(2004)Wang, Bovik, Sheikh, and Simoncelli]{wang2004ssim}
Zhou Wang, Alan~C Bovik, Hamid~R Sheikh, and Eero~P Simoncelli.
\newblock Image quality assessment: from error visibility to structural
  similarity.
\newblock \emph{IEEE transactions on image processing}, 2004.

\bibitem[Xu et~al.(2014)Xu, Ren, Liu, and Jia]{xu2014inpainting}
Li~Xu, Jimmy S.~J. Ren, Ce~Liu, and Jiaya Jia.
\newblock Deep convolutional neural network for image deconvolution.
\newblock In \emph{Proc. of NeurIPS}, 2014.

\bibitem[Yang(2019{\natexlab{a}})]{greg2019scalinglimit}
Greg Yang.
\newblock Scaling limits of wide neural networks with weight sharing: Gaussian
  process behavior, gradient independence, and neural tangent kernel
  derivation.
\newblock \emph{CoRR}, abs/1902.04760, 2019{\natexlab{a}}.

\bibitem[Yang(2019{\natexlab{b}})]{greg2019tensor_program_i}
Greg Yang.
\newblock Tensor programs {I:} wide feedforward or recurrent neural networks of
  any architecture are gaussian processes.
\newblock \emph{CoRR}, abs/1910.12478, 2019{\natexlab{b}}.

\bibitem[Zandieh et~al.(2021)Zandieh, Han, Avron, Shoham, Kim, and
  Shin]{zandieh2021scaling}
Amir Zandieh, Insu Han, Haim Avron, Neta Shoham, Chaewon Kim, and Jinwoo Shin.
\newblock Scaling neural tangent kernels via sketching and random features.
\newblock In \emph{Proc. of NeurIPS}, 2021.

\end{thebibliography}

\section{Statistical Interpretation of GA-NTK}

\label{sec:stats}Statistically, minimizing Eq. (\ref{eq:obj:ga-ntk})
or (\ref{eq:obj:bga-ntk-gen}) amounts to minimizing the Pearson $\chi^{2}$-divergence
\citep{jeffreys1946invariant}, a case of $f$-divergence \citep{renyi1961measures},
between $\mathcal{P}_{\text{data}}+\mathcal{P}_{\text{gen}}$ and
$2\mathcal{P}_{\text{gen}}$, where $\mathcal{P}_{\text{data}}$ is
the distribution of real data and $\mathcal{P}_{\text{gen}}$ is the
distribution of generated points. To see this, we first rewrite the
loss of our discrimonator $\mathcal{D}$, denoted by $\mathcal{L}(\mathcal{D})$,
in expectation: 
\begin{equation}
\arg\min_{\mathcal{D}}\mathcal{L}(\mathcal{D})=\arg\min_{\mathcal{D}}\mathbb{E}_{\boldsymbol{x}\sim\mathcal{P}_{\text{data}}}\bigl[(\mathcal{D}(\boldsymbol{x})-1)^{2}\bigr]+\mathbb{E}_{\boldsymbol{x}\sim\mathcal{P}_{\text{gen}}}\bigl[(\mathcal{D}(\boldsymbol{x})-0)^{2}\bigr].\label{eq:obj-stats-d}
\end{equation}
Here, $\mathcal{P}_{\text{gen}}$ can represent either $\boldsymbol{Z}$
in Eq. (\ref{eq:obj:ga-ntk}) or the output of the generator $\mathcal{G}$
in Eq. (\ref{eq:obj:bga-ntk-gen}). Similarly, the loss function for
our $\mathcal{P}_{\text{gen}}$, denoted by $\mathcal{L}(\mathcal{P}_{\text{gen}};\mathcal{D})$,
can be written as follows:
\begin{equation}
\arg\min_{\mathcal{P}_{\text{gen}}}\mathcal{L}(\mathcal{P}_{\text{gen}};\mathcal{D})=\arg\min_{\mathcal{P}_{\text{gen}}}\mathbb{E}_{\boldsymbol{x}\sim\mathcal{P}_{\text{data}}}\bigl[(\mathcal{D}(\boldsymbol{x})-1)^{2}\bigr]+\mathbb{E}_{\boldsymbol{x}\sim\mathcal{P}_{\text{gen}}}\bigl[(\mathcal{D}(\boldsymbol{x})-1)^{2}\bigr].\label{eq:obj-stats-g}
\end{equation}
GA-NTK, in the form of Eqs. (\ref{eq:obj-stats-d}) and (\ref{eq:obj-stats-g}),
is a special case of LSGAN \citep{mao2017lsgan}. Let $\mathcal{D}^{*}$
be the minimizer of Eq. (\ref{eq:obj-stats-d}). We can see that Eqs.
(\ref{eq:obj:ga-ntk}) and (\ref{eq:obj:bga-ntk-gen}) effectively
solve the problem:
\begin{equation}
\arg\min_{\mathcal{P}_{\text{gen}}}\mathcal{L}(\mathcal{P}_{\text{gen}};\mathcal{D}^{*})=\arg\min_{\mathcal{P}_{\text{gen}}}\mathbb{E}_{\boldsymbol{x}\sim\mathcal{P}_{\text{data}}}\bigl[(\mathcal{D}^{*}(\boldsymbol{x})-1)^{2}\bigr]+\mathbb{E}_{\boldsymbol{x}\sim\mathcal{P}_{\text{gen}}}\bigl[(\mathcal{D}^{*}(\boldsymbol{x})-1)^{2}\bigr].\label{eq:obj-stats-gd}
\end{equation}
\citet{mao2017lsgan} show that, under mild relaxation, minimizing
Eq. (\ref{eq:obj-stats-gd}) yields minimizing the Pearson $\chi^{2}$-divergence
between $\mathcal{P}_{\text{data}}+\mathcal{P}_{\text{gen}}$ and
$2\mathcal{P}_{\text{gen}}$:
\begin{align*}
\arg\min_{\mathcal{P}_{\text{gen}}}\mathcal{L}(\mathcal{P}_{\text{gen}};\mathcal{D}^{*}) & =\arg\min_{\mathcal{P}_{\text{gen}}}\chi_{\text{Pearson}}^{2}(\mathcal{P}_{\text{data}}+\mathcal{P}_{\text{gen}}\Vert2\mathcal{P}_{\text{gen}})\\
 & =\arg\min_{\mathcal{P}_{\text{gen}}}\int(\mathcal{P}_{\text{data}}(\boldsymbol{x})+\mathcal{P}_{\text{gen}}(\boldsymbol{x}))\left(\frac{2\mathcal{P}_{\text{gen}}(\boldsymbol{x})}{\mathcal{P}_{\text{data}}(\boldsymbol{x})+\mathcal{P}_{\text{gen}}(\boldsymbol{x})}-1\right)^{2}\textrm{d}\boldsymbol{x}.
\end{align*}
The loss becomes zero when $\mathcal{P}_{\text{data}}(\boldsymbol{x})=\mathcal{P}_{\text{gen}}(\boldsymbol{x})$
for all $\boldsymbol{x}$. Therefore, minimizing Eq. (\ref{eq:obj:ga-ntk})
or (\ref{eq:obj:bga-ntk-gen}) brings $\mathcal{P}_{\text{gen}}$
closer to $\mathcal{P}_{\text{data}}$.

\section{Proof of Theorem 3.1}

In this section, we prove the convergence of a GA-NTK whose discriminator
$\mathcal{D}$ approximates an infinite ensemble of infinitely-wide,
fully-connected, feedforward neural networks. The proof can be easily
extended to other network architectures such as convolutional neural
networks.

\subsection{Background and Notation}

Consider a fully-connected, feedforward neural network $f:\mathbb{R}^{d}\rightarrow\mathbb{R}$,
\begin{equation}
f(\boldsymbol{x};\boldsymbol{\theta})=\frac{\sigma_{w}}{\sqrt{d^{L-1}}}\boldsymbol{w}^{L}\phi\left(\frac{\sigma_{w}}{\sqrt{d^{L-2}}}\boldsymbol{W}^{L-1}\phi\left(\cdots\phi\left(\frac{\sigma_{w}}{\sqrt{d}}\boldsymbol{W}^{1}\boldsymbol{x}+\sigma_{b}\boldsymbol{b}^{1}\right)\cdots\right)+\sigma_{b}\boldsymbol{b}^{L-1}\right)+\sigma_{b}\boldsymbol{b}^{L},\label{eq:f}
\end{equation}
where $\phi(\cdot)$ is the activation function (applied element-wisely),
$L$ is the number of hidden layers, $\{d^{1},\cdots,d^{L-1}\}$ are
the dimensions (widths) of hidden layers, $\boldsymbol{\theta}=\cup_{l=1}^{L}\boldsymbol{\theta}^{l}=\cup_{l=1}^{L}(\boldsymbol{W}^{l}\in\mathbb{R}^{d^{l}\times d^{l-1}},\boldsymbol{b}^{l}\in\mathbb{R}^{d^{l}})$
are trainable weights and biases whose initial values are i.i.d. Gaussian
random variables $\mathcal{N}(0,1)$, and $\sigma_{w}^{2}$ and $\sigma_{b}^{2}$
are scaling factors that control the variances of weights and biases,
respectively.  Suppose $f$ is trained on a labeled dataset $\mathbb{D}^{2n}=(\boldsymbol{X}^{n}\oplus\boldsymbol{Z}^{n}\in\mathbb{R}^{2n\times d},\boldsymbol{1}^{n}\oplus\boldsymbol{0}^{n}\in\mathbb{R}^{2n})$
by minimizing the MSE loss using $t$ gradient-descent iterations
with the learning rate $\eta$. Let $\boldsymbol{\theta}^{(0)}$ and
$\boldsymbol{\theta}^{(t)}$ be the initial and trained parameters,
respectively. As $d^{1},\cdots,d^{L}\rightarrow\infty$, we can approximate
the distribution of $f(\boldsymbol{x};\boldsymbol{\theta}^{(t)})$
as a Gaussian process (NTK-GP) \citep{jacot2018ntk,lee2019ntk,chizat2018ntk}
whose behavior is controlled by a kernel matrix 
\begin{equation}
\boldsymbol{K}^{2n,2n}=\nabla_{\boldsymbol{\theta}}f(\boldsymbol{X}^{n}\oplus\boldsymbol{Z}^{n};\boldsymbol{\theta}^{(0)})^{\top}\nabla_{\boldsymbol{\theta}}f(\boldsymbol{X}^{n}\oplus\boldsymbol{Z}^{n};\boldsymbol{\theta}^{(0)})\in\mathbb{R}^{2n\times2n},\label{eq:kernel-L}
\end{equation}
where $f(\boldsymbol{X}^{n}\oplus\boldsymbol{Z}^{n};\boldsymbol{\theta}^{(0)})\in\mathbb{R}^{2n}$
is the vector of in-sample predictions made by the initial $f$. The
value of each element $K_{i,j}^{2n,2n}=k^{L}((\boldsymbol{X}^{n}\oplus\boldsymbol{Z}^{n})_{i,:},(\boldsymbol{X}^{n}\oplus\boldsymbol{Z}^{n})_{j,:})$
presents the similarity score of two rows (points) of $\boldsymbol{X}^{n}\oplus\boldsymbol{Z}^{n}$
in a kernel space, and it can be expressed by a kernel function $k^{L}:\mathbb{R}^{d}\times\mathbb{R}^{d}\rightarrow\mathbb{R}$,
called the neural tangent kernel (NTK). The NTK is deterministic as
it depends only on $\phi(\cdot)$, $\sigma_{w}$, $\sigma_{b}$, and
$L$ rather than the specific values in $\boldsymbol{\theta}^{(0)}$.
Furthermore, it can be evaluated layer-wisely. Let $h_{j}^{l}(\boldsymbol{x})\in\mathbb{R}^{d^{l}}$
be the pre-activation of the $j$-th neuron at the $l$-th layer of
$f(\boldsymbol{x};\boldsymbol{\theta}^{(t)})$. The distribution of
$h_{j}^{l}(\boldsymbol{x})$ is still an NTK-GP, and its associated
NTK is defined as $k^{l}:\mathbb{R}^{d}\times\mathbb{R}^{d}\rightarrow\mathbb{R}$,
\[
k^{l}(\boldsymbol{x},\boldsymbol{x}')=\nabla_{\boldsymbol{\theta}^{\leq l}}h_{j}^{l}(\boldsymbol{x})^{\top}\nabla_{\boldsymbol{\theta}^{\leq l}}h_{j}^{l}(\boldsymbol{x}'),
\]
where $\boldsymbol{\theta}^{\leq l}=\cup_{i=1}^{l}\boldsymbol{\theta}^{i}$.
Note that all $h_{j}^{l}(\boldsymbol{x})$'s, $\forall j$, are i.i.d.
and thus share the same kernel. It can be shown that
\begin{equation}
\begin{array}{lcl}
k^{l}(\boldsymbol{x},\boldsymbol{x}') & = & \nabla_{\boldsymbol{\theta}^{l}}h_{j}^{l}(\boldsymbol{x})^{\top}\nabla_{\boldsymbol{\theta}^{l}}h_{j}^{l}(\boldsymbol{x}')+\nabla_{\boldsymbol{\theta}^{\leq l-1}}h_{j}^{l}(\boldsymbol{x})^{\top}\nabla_{\boldsymbol{\theta}^{\leq l-1}}h_{j}^{l}(\boldsymbol{x}')\\
 & = & \tilde{k}^{l}(\boldsymbol{x},\boldsymbol{x}')+\sigma_{w}^{2}k^{l-1}(\boldsymbol{x},\boldsymbol{x}')\mathbb{E}_{(h_{j}^{(l-1)}(\boldsymbol{x}),\,h_{j}^{(l-1)}(\boldsymbol{x}'))\sim\mathcal{N}(\boldsymbol{0}^{2},\,\tilde{\boldsymbol{K}}^{l-1})}\left[\phi'(h_{j}^{(l-1)}(\boldsymbol{x}))\phi'(h_{j}^{(l-1)}(\boldsymbol{x}'))\right]
\end{array}\label{eq:kernel-recursion}
\end{equation}
and
\begin{equation}
k^{1}(\boldsymbol{x},\boldsymbol{x}')=\frac{\sigma_{w}^{2}}{d}\boldsymbol{x}^{\top}\boldsymbol{x}'+\sigma_{b}^{2}\label{eq:kernel-1}
\end{equation}
where $\tilde{k}^{l}:\mathbb{R}^{d}\times\mathbb{R}^{d}\rightarrow\mathbb{R}$
is the NNGP kernel \citep{lee2017nngp,matthews2018gaussian} that
controls the behavior of another Gaussian process, called NNGP, approximating
the distribution of $f(\boldsymbol{x};\boldsymbol{\theta}^{(0)})$,
and
\[
\tilde{\boldsymbol{K}}^{l-1}=\left[\begin{array}{cc}
\tilde{k}^{l-1}(\boldsymbol{x},\boldsymbol{x}) & \tilde{k}^{l-1}(\boldsymbol{x},\boldsymbol{x}')\\
\tilde{k}^{l-1}(\boldsymbol{x},\boldsymbol{x}') & \tilde{k}^{l-1}(\boldsymbol{x}',\boldsymbol{x}')
\end{array}\right]\in\mathbb{R}^{2\times2}.
\]

\subsection{Convergence }

The GA-NTK employs the above NTK-GP as the discriminator $\mathcal{D}$.
So, the in-sample mean predictions of $\mathcal{D}$ can be written
as a closed-form formula:
\begin{equation}
\mathcal{D}(\boldsymbol{X}^{n},\boldsymbol{Z}^{n})=(\boldsymbol{I}^{2n}-e^{-\eta t\boldsymbol{K}^{2n,2n}})\boldsymbol{y}^{2n}\in\mathbb{R}^{2n},\label{eq:prediction}
\end{equation}
where $\boldsymbol{I}^{2n}$ is an identity matrix and $\boldsymbol{y}^{2n}=\boldsymbol{1}^{n}\oplus\boldsymbol{0}^{n}\in\mathbb{R}^{2n}$
is the ``correct'' label vector for training $\mathcal{D}$. We
formulate the objective of GA-NTK as:
\begin{equation}
\arg\min_{\boldsymbol{Z}^{n}}\mathcal{L}(\boldsymbol{Z}^{n})=\arg\min_{\boldsymbol{Z}^{n}}\frac{1}{2}\Vert\boldsymbol{1}^{2n}-\mathcal{D}(\boldsymbol{X}^{n},\boldsymbol{Z}^{n})\Vert^{2},\label{eq:objective}
\end{equation}
where $\boldsymbol{1}^{2n}\in\mathbb{R}^{2n}$ in the loss $\mathcal{L}(\cdot)$
is the ``wrong'' label vector that guides us to find the points
($\boldsymbol{Z}^{n}$) that best deceive the discriminator. We show
that
\begin{thm}
\label{thm:convergence-1}Let $s$ be the number of the gradient descent
iterations solving Eq. (\ref{eq:objective}), and let $\boldsymbol{Z}^{n,(s)}$
be the solution at the $s$-th iteration. Suppose the following values
are bounded: (a) $\boldsymbol{X}_{i,j}^{n}$ and $\boldsymbol{Z}_{i,j}^{n,(0)}$,
$\forall i,j$, (b) $t$ and $\eta$, and (c) $\sigma$ and $L$.
Also, assume that (d) $\boldsymbol{X}^{n}$ contains finite, non-identical,
normalized rows. Then, for a sufficiently large $t$, we have 
\[
\min_{j\leq s}\Vert\nabla_{\boldsymbol{Z}^{n}}\mathcal{L}(\boldsymbol{Z}^{n,(j)})\Vert^{2}\leq O(\frac{1}{s-1}).
\]
\end{thm}

\subsection{Proof}

\noindent To prove Theorem \ref{thm:convergence-1}, we first introduce
the notion of $\beta$ smoothness:
\begin{definitn}
A continuously differentiable function $g:\mathbb{R}^{d}\to\mathbb{R}$
is $\beta$-smooth if there exits $\beta\in\mathbb{R}$ such that
\[
\Vert\nabla_{\boldsymbol{a}}g(\boldsymbol{a})-\nabla_{\boldsymbol{b}}g(\boldsymbol{b})\Vert\leq\beta\Vert\boldsymbol{a}-\boldsymbol{b}\Vert
\]
 for any $\boldsymbol{a},\boldsymbol{b}\in\mathbb{R}^{d}$.
\end{definitn}

\noindent It can be shown that gradient descent finds a stationary
point of a $\beta$-smooth function efficiently \citep{gower2018convergence}.
\begin{lemma}
\label{lemma:beta-smooth}Let $\boldsymbol{a}^{(s)}$ be the input
of a function $g:\mathbb{R}^{d}\to\mathbb{R}$ after applying $s$
gradient descent iterations to an initial input $\boldsymbol{a}^{(0)}$.
If $g$ is $\beta$-smooth, then $g(\boldsymbol{a}^{(s)})$ converges
to a stationary point at rate 
\[
\min_{j\leq s}\Vert\nabla_{\boldsymbol{a}}g(\boldsymbol{a}^{(j)})\Vert^{2}\leq O(\frac{1}{s-1}).
\]
\end{lemma}

\noindent So, our goal is to show that the loss $\mathcal{L}(\boldsymbol{Z}^{n})$
in Eq. (\ref{eq:objective}) is $\beta$-smooth w.r.t. any generated
point $\boldsymbol{z}\in\mathbb{R}^{d}$. 
\begin{cor}
\label{corol:bounded-gradient}If all the conditions (a)-(d) in Theorem
\ref{thm:convergence-1} hold, there exits a constant $c_{1}\in\mathbb{R}^{+}$
such that $\Vert\nabla_{\boldsymbol{z}}\mathcal{L}(\boldsymbol{Z}^{n})\Vert\leq c_{1}$
for each row $\boldsymbol{z}\in\mathbb{R}^{d}$ of $\boldsymbol{Z}^{n}$.
This makes $\mathcal{L}(\boldsymbol{Z}^{n})$ $\beta$-smooth.
\end{cor}

\noindent To prove Corollary \ref{corol:bounded-gradient}, consider
$\mathcal{D}_{i}(\boldsymbol{X}^{n},\boldsymbol{Z}^{n})$ and $\nabla_{z_{j}}\mathcal{L}(\boldsymbol{Z}^{n})$,
the $i$-th and $j$-th elements of $\mathcal{D}(\boldsymbol{X}^{n},\boldsymbol{Z}^{n})\in\mathbb{R}^{2n}$
and $\nabla_{\boldsymbol{z}}\mathcal{L}(\boldsymbol{Z}^{n})\in\mathbb{R}^{d}$,
respectively. We have
\begin{equation}
\begin{array}{ll}
\nabla_{z_{j}}\mathcal{L}(\boldsymbol{Z}^{n}) & =\nabla_{z_{j}}\frac{1}{2}\Vert\boldsymbol{1}^{2n}-\mathcal{D}(\boldsymbol{X}^{n},\boldsymbol{Z}^{n})\Vert^{2}\\
 & =\sum_{i=1}^{2n}\left(\mathcal{D}_{i}(\boldsymbol{X}^{n},\boldsymbol{Z}^{n})-1\right)\cdot\nabla_{z_{j}}\mathcal{D}_{i}(\boldsymbol{X}^{n},\boldsymbol{Z}^{n})
\end{array}\label{eq:gradient}
\end{equation}
Given a sufficiently large $t$, the $\mathcal{D}_{i}(\boldsymbol{X}^{n},\boldsymbol{Z}^{n})$
can be arbitrarily close to $y_{i}\in\{0,1\}$ because $\boldsymbol{K}^{2n,2n}$
is positive definite \citep{jacot2018ntk} and therefore $(\boldsymbol{I}^{2n}-e^{-\eta t\boldsymbol{K}^{2n,2n}})\rightarrow\boldsymbol{I}^{2n}$
as $t\rightarrow\infty$ in Eq. (\ref{eq:prediction}). There exists
$\epsilon\in\mathbb{R}^{+}$ such that
\[
\begin{array}{lcl}
\vert\nabla_{z_{j}}\mathcal{L}(\boldsymbol{Z}^{n})\vert & \leq & \epsilon\sum_{i=1}^{n}\vert\nabla_{z_{j}}\mathcal{D}_{i}(\boldsymbol{X}^{n},\boldsymbol{Z}^{n})+(1+\epsilon)\sum_{i=n+1}^{2n}\vert\nabla_{z_{j}}\mathcal{D}_{i}(\boldsymbol{X}^{n},\boldsymbol{Z}^{n})\vert\\
 & \leq & (1+\epsilon)\sum_{i=1}^{2n}\vert\nabla_{z_{j}}\mathcal{D}_{i}(\boldsymbol{X}^{n},\boldsymbol{Z}^{n})\vert\\
 & = & (1+\epsilon)\sum_{i=1}^{2n}\vert\nabla_{z_{j}}\sum_{p=1}^{2n}(I_{i,p}^{2n}-e_{i,p}^{-\eta t\boldsymbol{K}^{2n,2n}})y_{p}^{2n}\vert\\
 & = & (1+\epsilon)\eta t\sum_{i,p,q=1}^{2n}e_{i,q}^{-\eta t\boldsymbol{K}^{2n,2n}}\vert\nabla_{z_{j}}k^{L}((\boldsymbol{X}^{n}\oplus\boldsymbol{Z}^{n})_{q,:},(\boldsymbol{X}^{n}\oplus\boldsymbol{Z}^{n})_{p,:})y_{p}^{2n}\vert.
\end{array}
\]
Note that $e_{i,q}^{-\eta t\boldsymbol{K}^{2n,2n}}\in\mathbb{R}^{+}$
can be arbitrarily close to 0 with a sufficiently large $t$. Hence,
Corollary \ref{corol:bounded-gradient} holds as long as $\nabla_{z_{j}}k^{L}((\boldsymbol{X}^{n}\oplus\boldsymbol{Z}^{n})_{q,:},(\boldsymbol{X}^{n}\oplus\boldsymbol{Z}^{n})_{p,:})$
is bounded.
\begin{cor}
\label{corol:bounded-k}If the conditions (a)-(d) in Theorem \ref{thm:convergence-1}
hold, there exits a constant $c_{2}\in\mathbb{R}^{+}$ such that $\nabla_{z_{j}}k^{L}(\boldsymbol{a},\boldsymbol{b})\leq c_{2}$
for any two rows $\boldsymbol{a}$ and $\boldsymbol{b}$ of $\boldsymbol{X}^{n}\oplus\boldsymbol{Z}^{n}$. 
\end{cor}

\noindent It is clear that $\nabla_{z_{j}}k^{L}(\boldsymbol{a},\boldsymbol{b})=0$
if $\boldsymbol{a},\boldsymbol{b}\neq\boldsymbol{z}$. So, without
loss of generality, we consider $\nabla_{z_{j}}k^{L}(\boldsymbol{a},\boldsymbol{z})$
only. From Eq. (\ref{eq:kernel-recursion}), we have 
\[
\frac{\partial k^{L}(\boldsymbol{a},\boldsymbol{z})}{\partial z_{j}}=\frac{\partial k^{L}(\boldsymbol{a},\boldsymbol{z})}{\partial k^{L-1}(\boldsymbol{a},\boldsymbol{z})}\frac{\partial k^{L-1}(\boldsymbol{a},\boldsymbol{z})}{\partial k^{L-2}(\boldsymbol{a},\boldsymbol{z})}\cdots\frac{\partial k^{1}(\boldsymbol{a},\boldsymbol{z})}{\partial z_{j}}.
\]
For each $l=2,\cdots,L$, we can bound $\partial k^{l}(\boldsymbol{a},\boldsymbol{z})/\partial k^{l-1}(\boldsymbol{a},\boldsymbol{z})$
by
\[
\begin{array}{lll}
\frac{\partial k^{l}(\boldsymbol{a},\boldsymbol{z})}{\partial k^{l-1}(\boldsymbol{a},\boldsymbol{z})} & = & \sigma_{w}^{2}\mathbb{E}_{(h_{j}^{(l-1)}(\boldsymbol{x}),\,h_{j}^{(l-1)}(\boldsymbol{x}'))\sim\mathcal{N}(\boldsymbol{0}^{2},\,\tilde{\boldsymbol{K}}^{l-1})}\left[\phi'(h_{j}^{(l-1)}(\boldsymbol{x}))\phi'(h_{j}^{(l-1)}(\boldsymbol{x}'))\right]\\
 & \leq & (\sigma_{w}\max_{h}\phi'(h))^{2}
\end{array}
\]
provided that the maximum slope of $\phi$ is limited, which is true
for many popular activation functions including ReLU and erf. Also,
by Eq. (\ref{eq:kernel-1}), the value
\[
\frac{\partial k^{1}(\boldsymbol{a},\boldsymbol{z})}{\partial z_{j}}=\frac{\sigma_{w}^{2}}{d}a_{j}
\]
is bounded. Therefore, Corollary \ref{corol:bounded-k} holds, which
in turn makes $\mathcal{L}(\boldsymbol{Z}^{n})$ $\beta$-smooth via
Corollary \ref{corol:bounded-gradient}. By Lemma \ref{lemma:beta-smooth},
we obtain the proof of Theorem \ref{thm:convergence-1}.

\section{Experiment Settings}

This section provides more details about the settings of our experiments.

\subsection{Model Settings}

The network architectures of the baseline GANs used in our experiments
are based on InfoGAN \citep{Chen16InfoGAN}. We set the latent dimensions,
training iterations, and batch size according to the study \citep{lucic2018are-gans-created-equal}.
The latent dimensions for the generator are all 64. The batch size
for all baselines is set to 64. The training iterations are 80K, 100K,
and 400K for MNIST, CelebA, and CIFAR-10 datasets, respectively. For
the optimizers, we follow the setting from the respective original
papers. Below we list the network architecture of the baselines for
each dataset as well as the optimizer settings. 

\begin{table}[H]
\caption{The architectures of the discriminator and generator in the baseline
GANs for the MNIST dataset.}
\medskip{}

\centering{}%
\begin{tabular}{ll}
\toprule 
Discriminator & Generator\tabularnewline
\midrule
\midrule 
Input 28$\times$28$\times$1 Gray image & Input$\in\mathbb{R}^{64}\sim\mathcal{N}(\boldsymbol{0},\boldsymbol{I})$\tabularnewline
\midrule 
4$\times$4 conv; 64 leaky ReLU; stride 2 & Fully Connected 1024 ReLU; batchnorm\tabularnewline
\midrule 
4$\times$4 conv; 128 leaky ReLU; stride 2. batchnorm & Fully Connected $7\times7\times128$ ReLU; batchnorm\tabularnewline
\midrule 
Fully Connected 1024 leaky ReLU; batchnorm & 4$\times$4 deconv; 64 ReLU. stride 2; batchnorm\tabularnewline
\midrule 
Fully Connected 1 output & 4$\times$4 deconv; 1 sigmoid\tabularnewline
\bottomrule
\end{tabular}
\end{table}

\begin{table}[H]
\caption{The architectures of the discriminator and generator in the baseline
GANs for the CIFAR-10 dataset.}
\medskip{}

\centering{}%
\begin{tabular}{ll}
\toprule 
discriminator & generator\tabularnewline
\midrule
\midrule 
Input 32$\times$32$\times$3 Image & Input$\in\mathbb{R}^{64}\sim\mathcal{N}(\boldsymbol{0},\boldsymbol{I})$\tabularnewline
\midrule 
4$\times$4 conv; 64 leaky ReLU; stride 2 & Fully Connected $2\times2\times448$ ReLU; batchnorm\tabularnewline
\midrule 
4$\times$4 conv; 128 leaky ReLU; stride 2; batchnorm & 4$\times$4 deconv; 256 ReLU; stride 2; batchnorm\tabularnewline
\midrule 
4$\times$4 conv; 256 leaky ReLU; stride 2; batchnorm & 4$\times$4 deconv; 128 ReLU; stride 2\tabularnewline
\midrule 
Fully Connected 1 output & 4$\times$4 deconv; 64 ReLU; stride 2\tabularnewline
\midrule 
 & 4$\times$4 deconv; 3 Tanh; stride 2.\tabularnewline
\bottomrule
\end{tabular}
\end{table}

\begin{table}[H]
\caption{The architectures of the discriminator and generator in the baseline
GANs for the CelebA dataset.}
\medskip{}

\centering{}%
\begin{tabular}{ll}
\toprule 
discriminator & generator\tabularnewline
\midrule
\midrule 
Input 64$\times$64$\times$3 Image & Input$\in\mathbb{R}^{64}\sim\mathcal{N}(\boldsymbol{0},\boldsymbol{I})$\tabularnewline
\midrule 
4$\times$4 conv; 64 leaky ReLU; stride 2 & Fully Connected $2\times2\times448$ ReLU; batchnorm\tabularnewline
\midrule 
4$\times$4 conv; 128 leaky ReLU; stride 2; batchnorm & 4$\times$4 deconv; 256 ReLU; stride 2; batchnorm\tabularnewline
\midrule 
4$\times$4 conv; 256 leaky ReLU; stride 2; batchnorm & 4$\times$4 deconv; 128 ReLU; stride 2\tabularnewline
\midrule 
4$\times$4 conv; 256 leaky ReLU; stride 2; batchnorm & 4$\times$4 deconv; 64 ReLU; stride 2\tabularnewline
\midrule 
Fully Connected 1 output & 4$\times$4 deconv; 32 ReLU; stride 2\tabularnewline
\midrule 
 & 4$\times$4 deconv; 3 Tanh; stride 2.\tabularnewline
\bottomrule
\end{tabular}
\end{table}

\begin{table}[H]
\centering{}\caption{The optimizer settings for each GAN baseline. $n_{dis}$ denotes the
training steps for discriminators in the alternative training process.}
\medskip{}
\begin{tabular}{lllllll}
\toprule 
 &  & Optimizer type & Learning Rate & $\beta_{1}$ & $\beta_{2}$ & $n_{dis}$\tabularnewline
\midrule
\midrule 
DCGAN &  & Adam & 0.0002 & 0.5 & 0.999 & 1\tabularnewline
\midrule 
LSGAN &  & Adam & 0.0002 & 0.5 & 0.999 & 1\tabularnewline
\midrule 
WGAN &  & RMSProp & 0.00005 & None & None & 5\tabularnewline
\midrule 
WGAN-GP &  & Adam & 0.0001 & 0.5 & 0.9 & 5\tabularnewline
\midrule 
SN-GAN &  & Adam & 0.0001 & 0.9 & 0.999 & 5\tabularnewline
\bottomrule
\end{tabular}
\end{table}

Note that we remove all the batchnorm layers for the discriminators
in WGAN-GP. We architect the element network of the discriminator
in our GA-NTK following InfoGAN \citep{Chen16InfoGAN}, except that
the width (or the number of filters) of the network is infinite at
each layer and has no batchnorm layers.

The generator of GA-NCTKg consumes memory. To reduce memory consumption,
we let $\mathcal{D}$ discriminates true and fake images in the code
space of a pre-trained autoencoder $\mathcal{A}$ \citep{Bergmann2019ImprovingUD}.
After training, a code output by $\mathcal{G}$ is fed into the decoder
of $\mathcal{A}$ to obtain an image. The architectures of the pre-trained
$\mathcal{A}$ for different datasets are summarized as follows: \newpage{}
\begin{table}[H]
\centering{}\caption{The architectures of $\mathcal{A}$ for different datasets. }
\medskip{}
\begin{tabular}{ll}
\toprule 
MNIST & CIFAR-10\tabularnewline
\midrule
\midrule 
Input 28$\times$28$\times$1 Image & Input 32$\times$32$\times$3 Image\tabularnewline
\midrule 
3$\times$3 conv; 16 SeLU; stride 2 & 3$\times$3 conv; 32 SeLU; stride 2\tabularnewline
\midrule 
3$\times$3 conv; 32 SeLU; stride 2 & 3$\times$3 conv; 64 SeLU; stride 2\tabularnewline
\midrule 
3$\times$3 conv; 64 SeLU; stride 2 & 3$\times$3 conv; 128 SeLU; stride 2\tabularnewline
\midrule 
Fully Connected; 128 tanh & Fully Connected; 1024 tanh\tabularnewline
\midrule 
3$\times$3 transposeconv; 64 SeLU; stride 2 & 3$\times$3 transposeconv; 128 SeLU; stride 2\tabularnewline
\midrule 
3$\times$3 transposeconv; 32 SeLU; stride 2 & 3$\times$3 transposeconv; 64 SeLU; stride 2\tabularnewline
\midrule 
3$\times$3 transposeconv; 16 SeLU; stride 2 & 3$\times$3 transposeconv; 32 SeLU; stride 2\tabularnewline
\midrule 
output & output\tabularnewline
\midrule
 & \tabularnewline
\midrule 
CelebA & CelebA-HQ\tabularnewline
\midrule
\midrule 
Input 64$\times$64$\times$3 Image & Input 256$\times$256$\times$3 Image\tabularnewline
\midrule 
3$\times$(3$\times$3 conv; 32 SeLU; stride 1) & 3$\times$(3$\times$3 conv; 64 SeLU; stride 1)\tabularnewline
\midrule 
3$\times$3 conv; 32 SeLU; stride 2 & 3$\times$3 conv; 64 SeLU; stride 2\tabularnewline
\midrule 
3$\times$(3$\times$3 conv; 64 SeLU; stride 1) & 3$\times$(3$\times$3 conv; 128 SeLU; stride 1)\tabularnewline
\midrule 
3$\times$3 conv; 64 SeLU; stride 2 & 3$\times$3 conv; 128 SeLU; stride 2\tabularnewline
\midrule 
3$\times$(3$\times$3 conv; 128 SeLU; stride 1) & 3$\times$(3$\times$3 conv; 256 SeLU; stride 1)\tabularnewline
\midrule 
3$\times$3 conv; 128 SeLU; stride 2 & 3$\times$3 conv; 256 SeLU; stride 2\tabularnewline
\midrule 
Fully Connected; 2048 tanh & 3$\times$(3$\times$3 conv; 512 SeLU; stride 1)\tabularnewline
\midrule 
3$\times$(3$\times$3 transposeconv; 128 SeLU; stride 1) & 3$\times$3 conv; 512 SeLU; stride 2\tabularnewline
\midrule 
3$\times$3 transposeconv; 128 SeLU; stride 2 & Fully Connected; 2048 tanh\tabularnewline
\midrule 
3$\times$(3$\times$3 transposeconv; 64 SeLU; stride 1) & 3$\times$(3$\times$3 transposeconv; 512 SeLU; stride 1)\tabularnewline
\midrule 
3$\times$3 transposeconv; 64 SeLU; stride 2 & 3$\times$3 transposeconv; 512 SeLU; stride 2\tabularnewline
\midrule 
3$\times$(3$\times$3 transposeconv; 32 SeLU; stride 1) & 3$\times$(3$\times$3 transposeconv; 256 SeLU; stride 1)\tabularnewline
\midrule 
3$\times$3 transposeconv; 32 SeLU; stride 2 & 3$\times$3 transposeconv; 256 SeLU; stride 2\tabularnewline
\midrule 
output & 3$\times$(3$\times$3 transposeconv; 128 SeLU; stride 1)\tabularnewline
\midrule 
 & 3$\times$3 transposeconv; 128 SeLU; stride 2\tabularnewline
\midrule 
 & 3$\times$3(3$\times$3 transposeconv; 64 SeLU; stride 1)\tabularnewline
\midrule 
 & 3$\times$3 transposeconv; 64 SeLU; stride 2\tabularnewline
\midrule 
 & output\tabularnewline
\bottomrule
\end{tabular}
\end{table}

\subsection{Metrics}

The FID scores are computed using the code from the original paper
\citep{Heusel17FID}. We sample 2048 images to compute the FID scores.
We calculate the AM-SSIM scores using the SSIM settings: filter size
4, filter sigma 1.5, $k_{1}$ 0.01, and $k_{2}$ 0.03 \citep{wang2004ssim}.

\subsection{Hyperparameter Tuning}

For each data synthesis method, we tune its hyperparameter using grid
search. \textbf{GA-NTK.} The computation of $\boldsymbol{K}^{2n,2n}$
requires one to determine the initialization and architecture of the
element networks in the ensemble discriminator. \citet{Poole16transient-chaos,schoenholz17deep-information-prop,raghus17exp-of-dnn}
have proposed a principled method to tune the hyperparameters for
the initialization. From our empirical results, we also find that
the quality of the images generated by GA-NTK is not significantly
impacted by the choice of the architecture---a fully connected network
with rectified linear unit (ReLU) activation suffices to generate
recognizable image patterns. Once $\boldsymbol{K}^{2n,2n}$ is decided,
there is only one hyperparameter $\lambda=\eta t$ to tune in Eq.
(\ref{eq:objective}). The $\lambda$ controls how well the discriminator
is trained on $\mathbb{D}$, so either a too small or large value
can lead to poor gradients for $\boldsymbol{Z}^{n}$ and final generated
points. But since there is no alternating updates as in GANs, we can
decide an appropriate value of $\lambda$ without worrying about canceling
the learning progress of $\boldsymbol{Z}^{n}$. We propose a simple,
unidirectional search algorithm for tuning $\lambda$, as shown in
Algorithm \ref{alg:search-lambda-1}. Basically, we search, from small
to large, for a value that makes the discriminator nearly separate
the real data from pure noises in an auxiliary learning task, and
then use this value to solve Eq. (\ref{eq:objective}). In practice,
a small positive $\epsilon$ ranging from $10^{-3}$ to $10^{-2}$
suffices to give an appropriate $\lambda$. \textbf{Multi-resolutional
GA-NTK.} We use 3 NTK-GP's as the discriminators, whose architectures
are listed in Table \ref{tab:multi-res ga-ntk}. 
\begin{algorithm}
\KwIn{Data $\boldsymbol{X}^{n}$, kernel $k$, and separation tolerance
$\epsilon$}

\KwOut{$\lambda$ for GA-NTK}

\BlankLine

Randomly initiate $\boldsymbol{Z}^{n}\in\mathbb{R}^{n\times d}$

$\lambda\leftarrow1$

\While{$\frac{1}{2n}\Vert\mathcal{D}(\boldsymbol{X}^{n},\boldsymbol{Z}^{n};k,\lambda)-(\boldsymbol{1}^{n}\oplus\boldsymbol{0}^{n})\Vert^{2}\leq\epsilon$}{

$\lambda\leftarrow\lambda\cdot2$

}

\Return{$\lambda$}\medskip{}

\caption{\label{alg:search-lambda-1}Unidirectional search for the hyperparameter
$\lambda$ of GA-NTK.}
\end{algorithm}
\begin{table}[H]
\caption{\label{tab:multi-res ga-ntk}The architectures of the discriminators
for multi-resolution GA-NTK.}
\medskip{}

\centering{}%
\begin{tabular}{l}
\toprule 
Discriminator small\tabularnewline
\midrule
\midrule 
Input 16$\times$16$\times$3 Image\tabularnewline
\midrule 
4$\times$4 conv; ReLU; stride 2\tabularnewline
\midrule 
4$\times$4 conv; ReLU; stride 2\tabularnewline
\midrule 
Fully Connected 1 output\tabularnewline
\bottomrule
\end{tabular}\quad{}%
\begin{tabular}{l}
\toprule 
Discriminator medium\tabularnewline
\midrule
\midrule 
Input 64$\times$64$\times$3 Image\tabularnewline
\midrule 
4$\times$4 conv; ReLU; stride 2\tabularnewline
\midrule 
4$\times$4 conv; ReLU; stride 2\tabularnewline
\midrule 
4$\times$4 conv; ReLU; stride 2\tabularnewline
\midrule 
4$\times$4 conv; ReLU; stride 2\tabularnewline
\midrule 
Fully Connected 1 output\tabularnewline
\bottomrule
\end{tabular}\quad{}%
\begin{tabular}{l}
\toprule 
Discriminator large\tabularnewline
\midrule
\midrule 
Input 256$\times$256$\times$3 Image\tabularnewline
\midrule 
4$\times$4 conv; ReLU; stride 2\tabularnewline
\midrule 
4$\times$4 conv; ReLU; stride 2\tabularnewline
\midrule 
4$\times$4 conv; ReLU; stride 2\tabularnewline
\midrule 
4$\times$4 conv; ReLU; stride 2\tabularnewline
\midrule 
4$\times$4 conv; ReLU; stride 2\tabularnewline
\midrule 
4$\times$4 conv; ReLU; stride 2\tabularnewline
\midrule 
Fully Connected 1 output\tabularnewline
\bottomrule
\end{tabular}
\end{table}

\section{More Experiments}

\subsection{GA-FNTK vs. GA-CNTK}

\begin{figure}
\begin{centering}
{\footnotesize{}}%
\begin{tabular}{cc}
{\footnotesize{}\includegraphics[width=5cm]{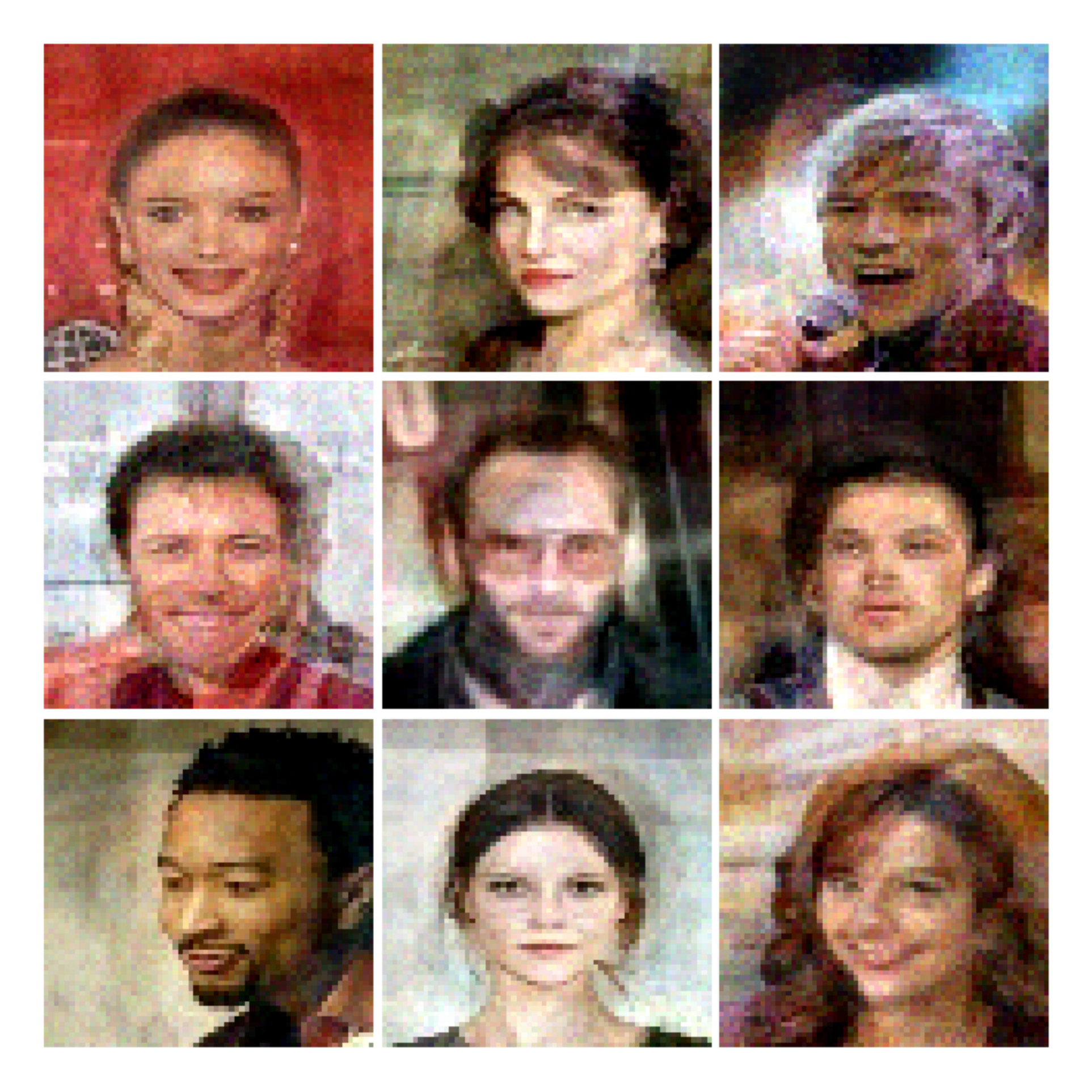}} & {\footnotesize{}\includegraphics[width=5cm]{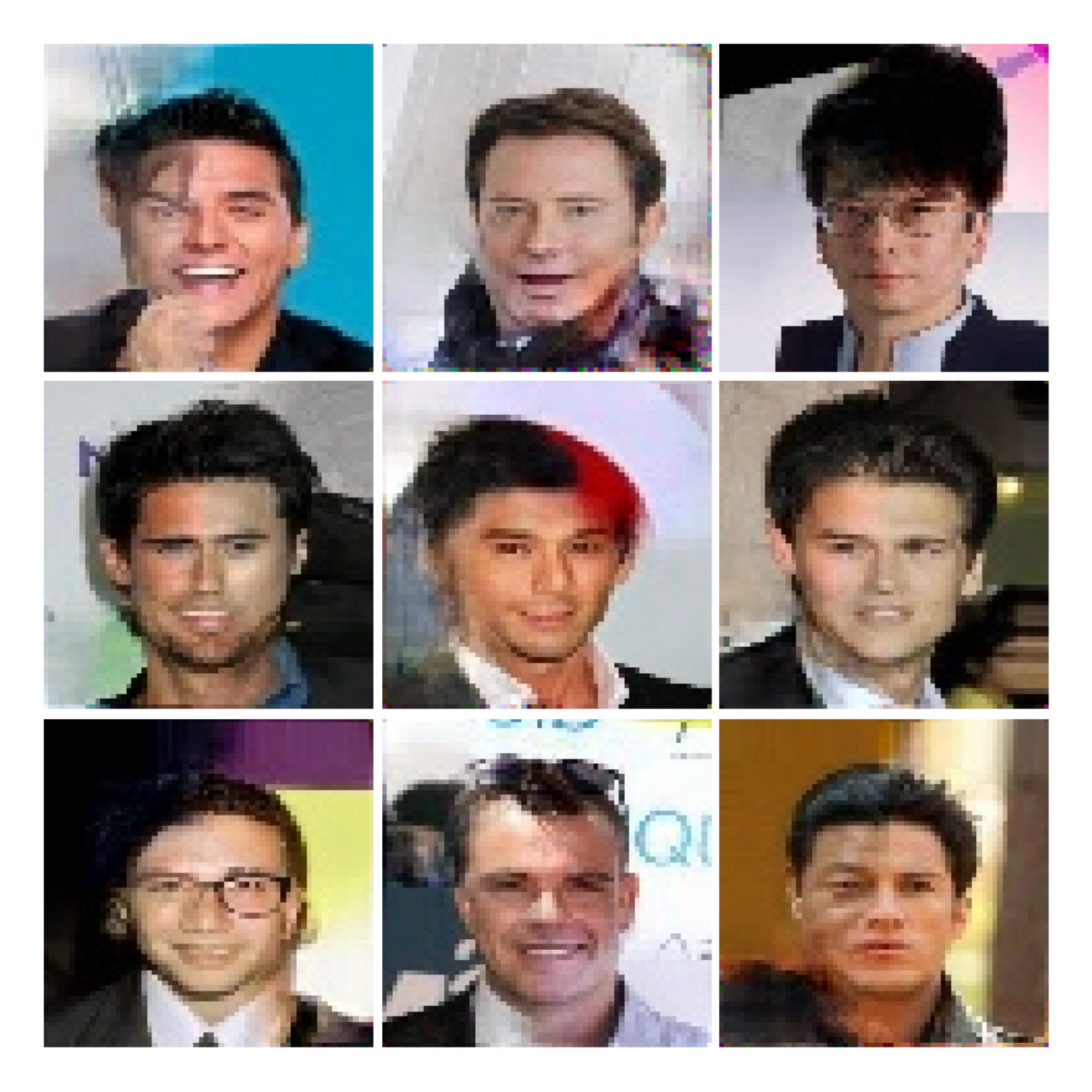}}\tabularnewline
{\footnotesize{}(a) GA-FNTK on CelebA} & {\footnotesize{}(b) GA-CNTK on CelebA}\tabularnewline
\end{tabular}{\footnotesize\par}
\par\end{centering}
\caption{\label{fig:FNTKvsCNTK}The images generated by (a) GA-FNTK and (b)
GA-CNTK given 256 CelebA training images.}
\end{figure}

Next, we compare the images generated by GA-FNTK, GA-CNTK, and the
multi-resolutional GA-CNTK described in Section \ref{subsec:GA-NTK-in-Practice}
on the CelebA and CelebA-HQ datasets. The multi-resolutional GA-CNTK
employs 3 discriminators working at $256\times256$, 64$\times$64,
and 16$\times$16 pixel resolutions, respectively. Figure \ref{fig:FNTKvsCNTK}
shows the results. To our surprise, GA-NTK (which models the discriminator
as an ensemble of fully connected networks) suffices to generate recognizable
faces. The images synthesized by GA-FNTK and GA-CNTK lack details
and global coherence, respectively, due to the characteristics of
FNNs and CNNs. On the other hand, the multi-resolutional GA-CNTK gives
both the details and global coherence thanks to the multiple discriminators
working at different pixel resolutions. The results also demonstrate
the potential of GA-NTK variants to generate high-quality data as
there are many other techniques for GANs that could be adapted into
GA-NTK.

\subsection{Batch-wise GA-NTK\label{subsec:Batch-Wise-GA-NTK}}

To work with a larger training set, we modify GA-CNTK by following
the instructions in Section \ref{subsec:GA-NTK-in-Practice} to obtain
the batch-wise GA-CNTK, which computes the gradients of $\boldsymbol{Z}^{n}$
in Eq. (\ref{eq:obj:ga-ntk}) from 256 randomly sampled training images
during each gradient descent iteration. We train the batch-wise GA-CNTK
on two larger datasets consisting of 2048 images from CelebA and 1300
images from ImageNet, respectively. Figure \ref{fig:Comparison-between-BGA-CNTK}
shows the results, and the batch-wise GA-CNTK can successfully generate
the ``daisy'' images on ImageNet.

\begin{figure}
\begin{centering}
\includegraphics[width=2cm]{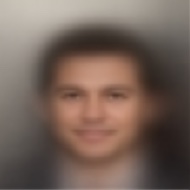}
\par\end{centering}
\caption{\label{fig:b1}When $b=1$, GA-NTK tends to generate a blurry mean
image.}
\end{figure}

\begin{figure}
\begin{centering}
\begin{tabular}{cc}
\includegraphics[width=5cm]{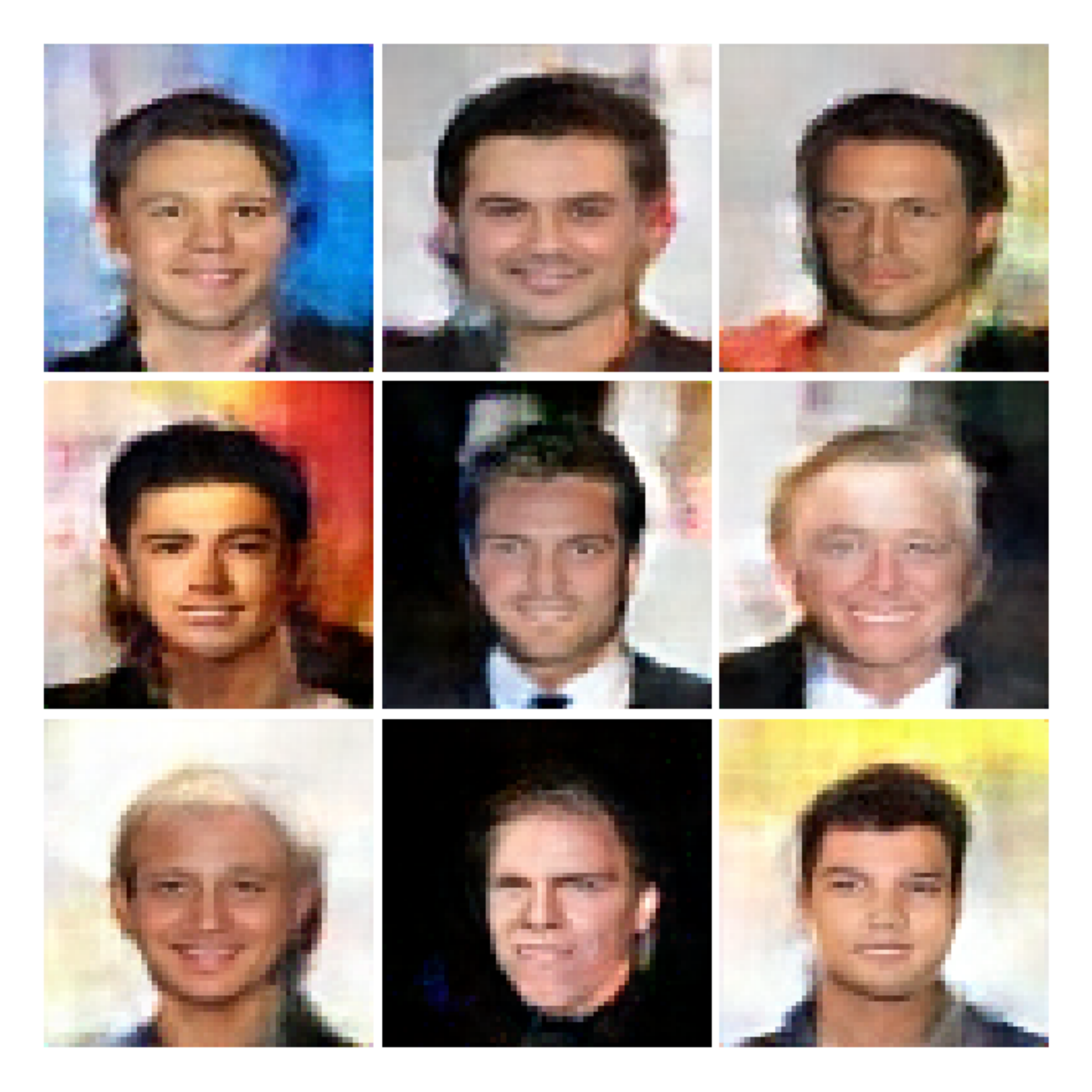} & \includegraphics[width=5cm]{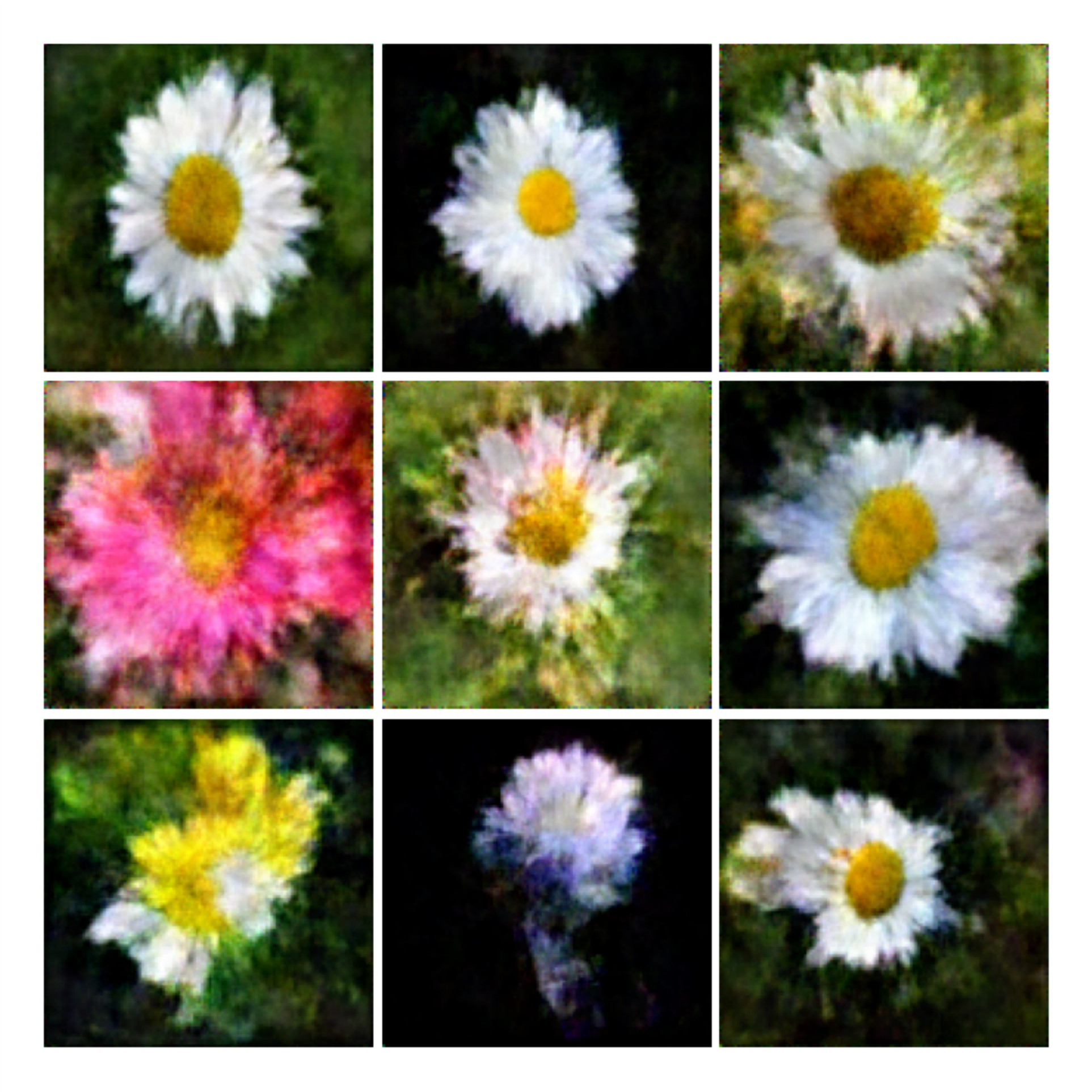}\tabularnewline
(a) & (b)\tabularnewline
\end{tabular}
\par\end{centering}
\caption{\label{fig:Comparison-between-BGA-CNTK}The images generated by batch-wise
GA-CNTK on (a) CelebA dataset of 2048 randomly sampled images and
(b) ImageNet dataset of 1300 randomly sampled images.}

\end{figure}

Note that the batch-wise GA-CNTK solves a different problem than the
original GA-CNTK---the former finds $\boldsymbol{Z}^{n}$ that deceives
\emph{multiple} discriminators, each trained on 256 examples, while
the latter searches for $\boldsymbol{Z}^{n}$ that fools a single
discriminator trained on 256 examples. We found that, when the batch
size is small ($b=1$), GA-NTK tends to generate a blurry mean image
regardless of model architectures and initializations of model weights
and $\boldsymbol{Z}^{n}$, as shown in Figure \ref{fig:b1}. This
is because the mean image is the best for simultaneously fooling many
NTK discriminators, each trained on a single example. However, in
practice this setting is less common as one usually aims to use the
largest \textbf{$b$} possible \citep{brock2018large}. Figure \ref{fig:Comparison-between-BGA-CNTK}
shows that a batch size of 256 suffices to give plausible results
on the CelebA and ImageNet datasets. Comparing the images in Figure
\ref{fig:Demo}(f) with those in Figure \ref{fig:Comparison-between-BGA-CNTK}(a),
we can see that the batch-wise GA-CNTK gives a little more blurry
images but the patterns in each synthesized image are more globally
coherent, both due to the effect of multiple discriminators.


\subsection{Sensitivity to Hyperparameters}

\begin{figure}
\begin{centering}
\begin{tabular}{cc}
\includegraphics[width=5cm]{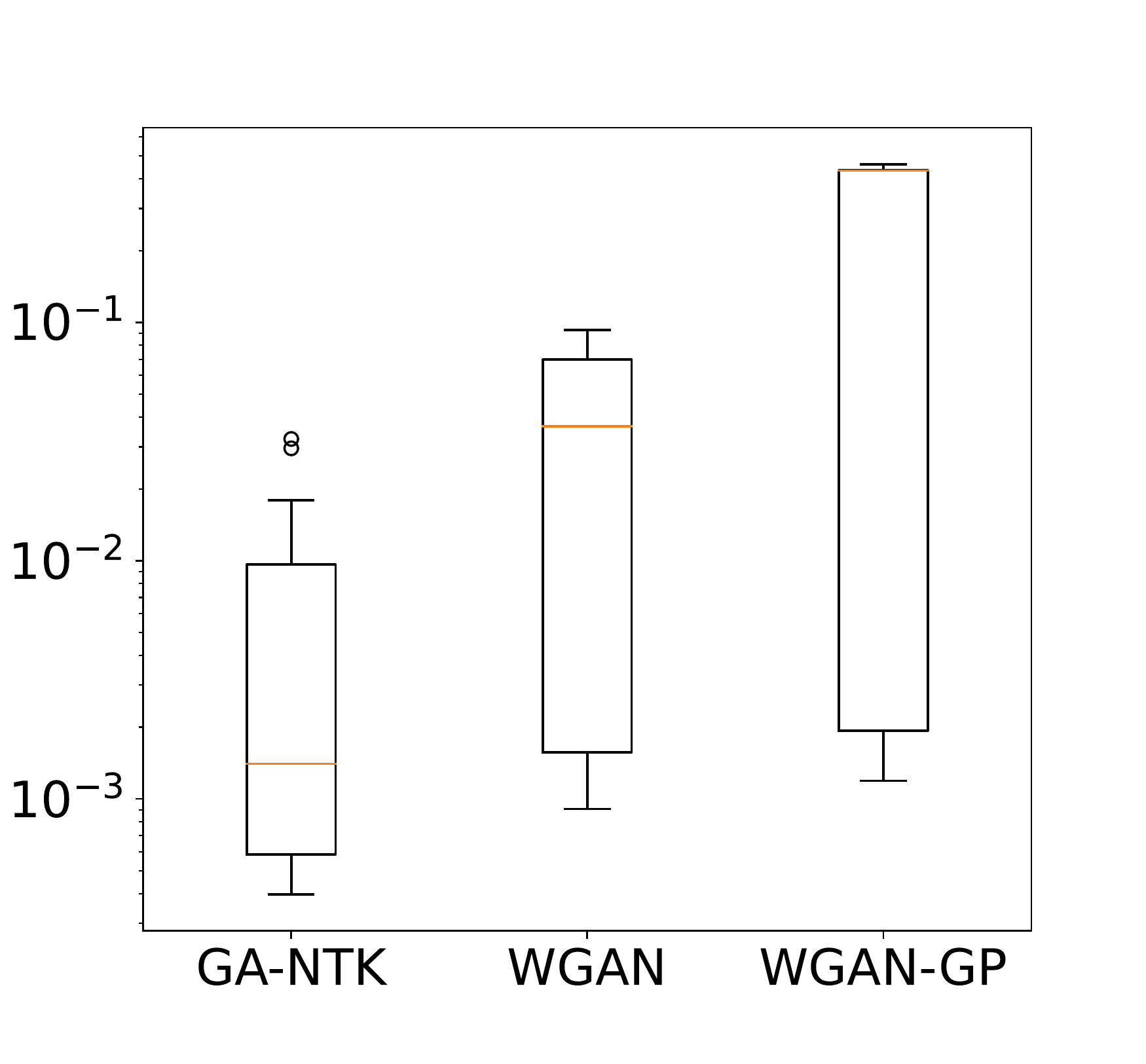} & \includegraphics[width=5cm]{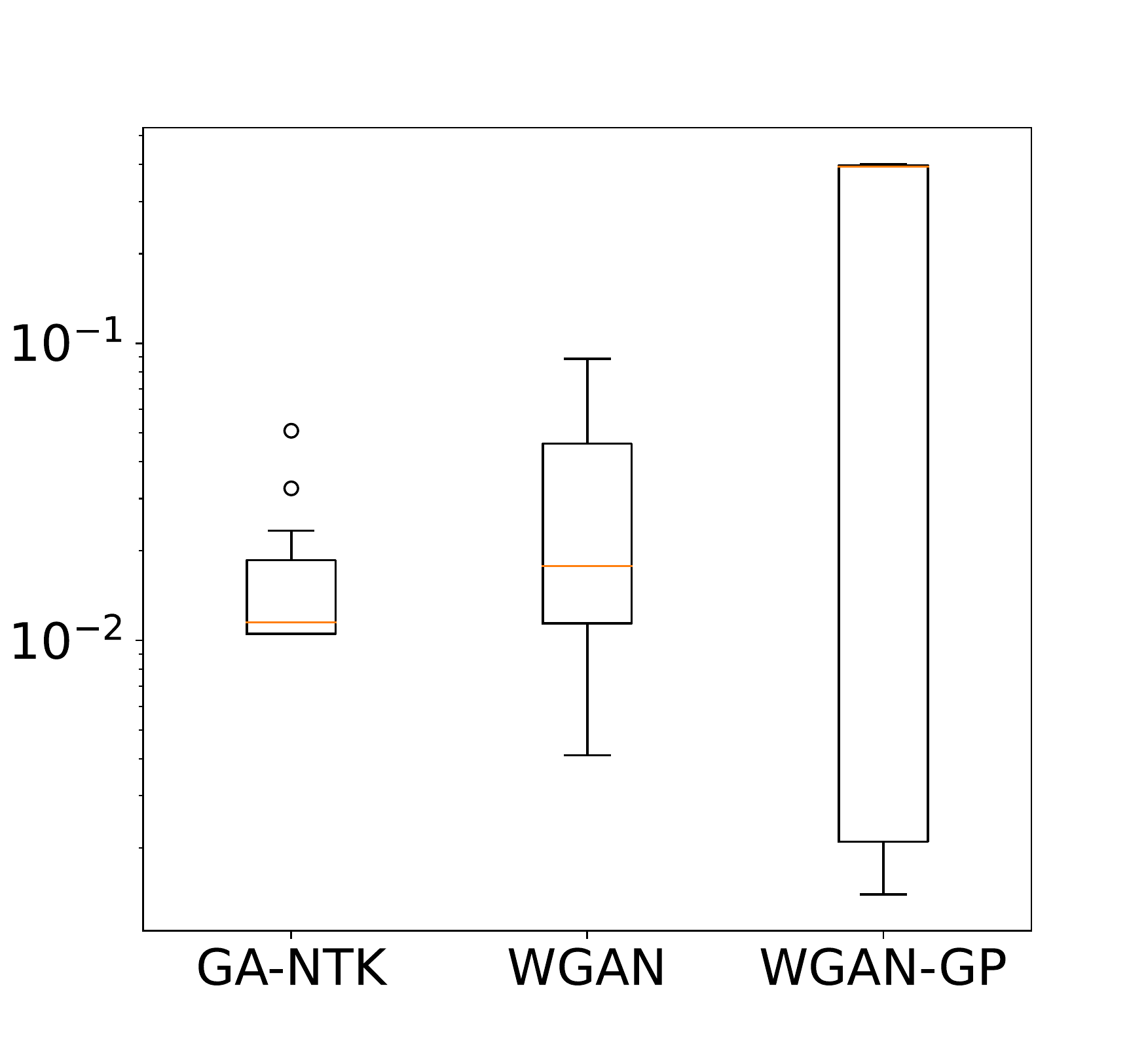}\tabularnewline
(a) & (b)\tabularnewline
\end{tabular}
\par\end{centering}
\caption{\label{fig:The-loss-distribution}The distribution of Wasserstein
distance between $\mathcal{P}_{\text{gen}}$ and $\mathcal{P}_{\text{data}}$
(used to measure the quality of the generated points) over the searched
hyper-parameters on training sets of (a) 8- and (b) 25-modal Gaussian
mixtures.}

\end{figure}

Here, we study how sensitive is the performance of WGAN, WGAN-GP,
and GA-FNTK to their hyperparameters. We adjust the hyperparameters
of different approaches using the grid search  under a time budget
of 3 hours, and then evaluate the quality of 2048 generated data points
by the Wasserstein distance between $\mathcal{P}_{\text{gen}}$ and
$\mathcal{P}_{\text{data}}$. We train different methods on two toy
datasets consisting of 8- and 25-modal Gaussian mixtures following
the settings described in Section \ref{subsec:Exp-Training-Stability}.
 Figure \ref{fig:The-loss-distribution} shows the results, and we
can see that GA-FNTK achieves the lowest average Wasserstein distance
in both cases. Moreover, its variances are smaller than the two other
baselines, too. This shows that the performance of GA-FNTK is less
sensitive to the hyperparameters and could be easier to tune in practice.

\begin{figure}[H]
\centering{}%
\begin{tabular}{cccc}
\includegraphics[width=3cm]{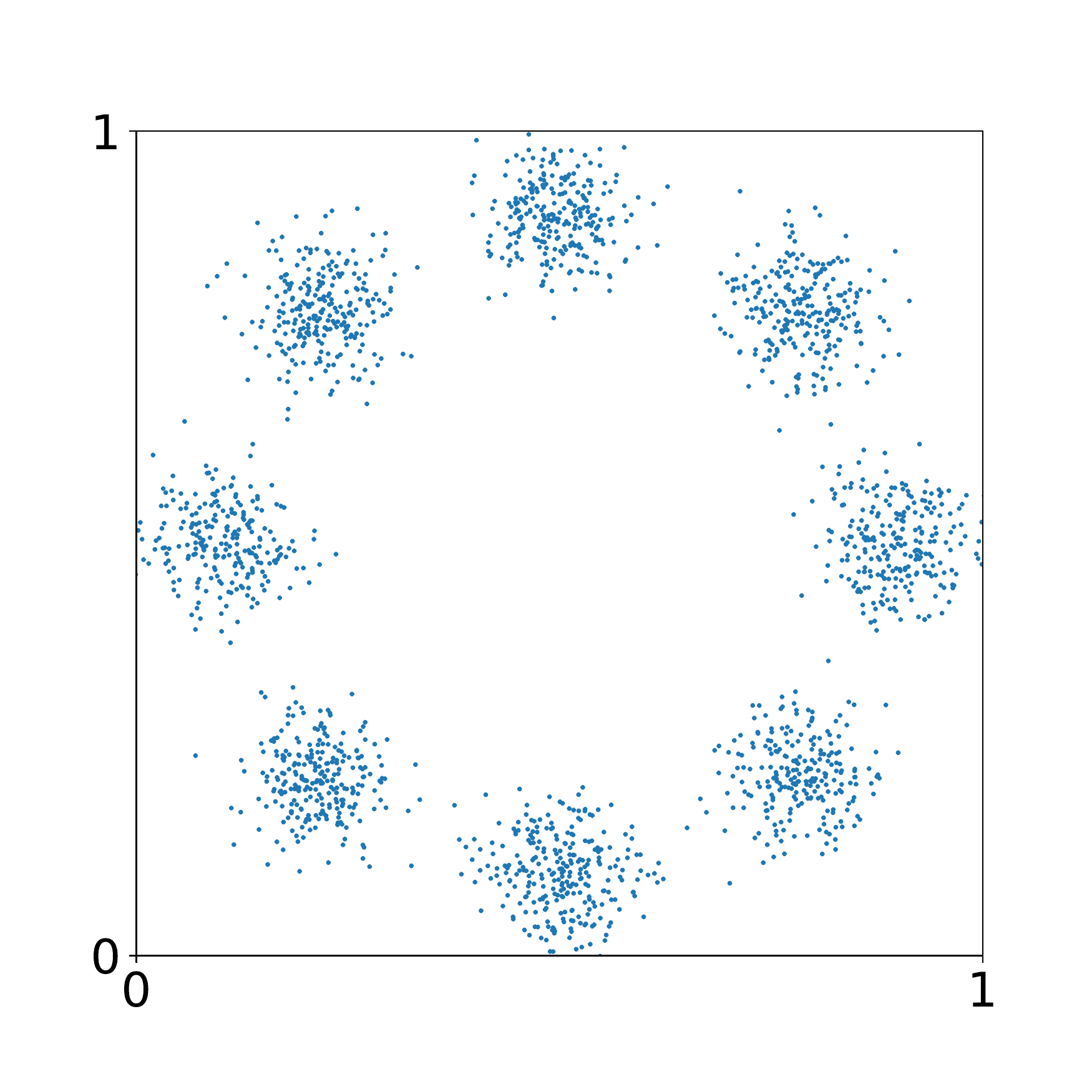} & \includegraphics[width=3cm]{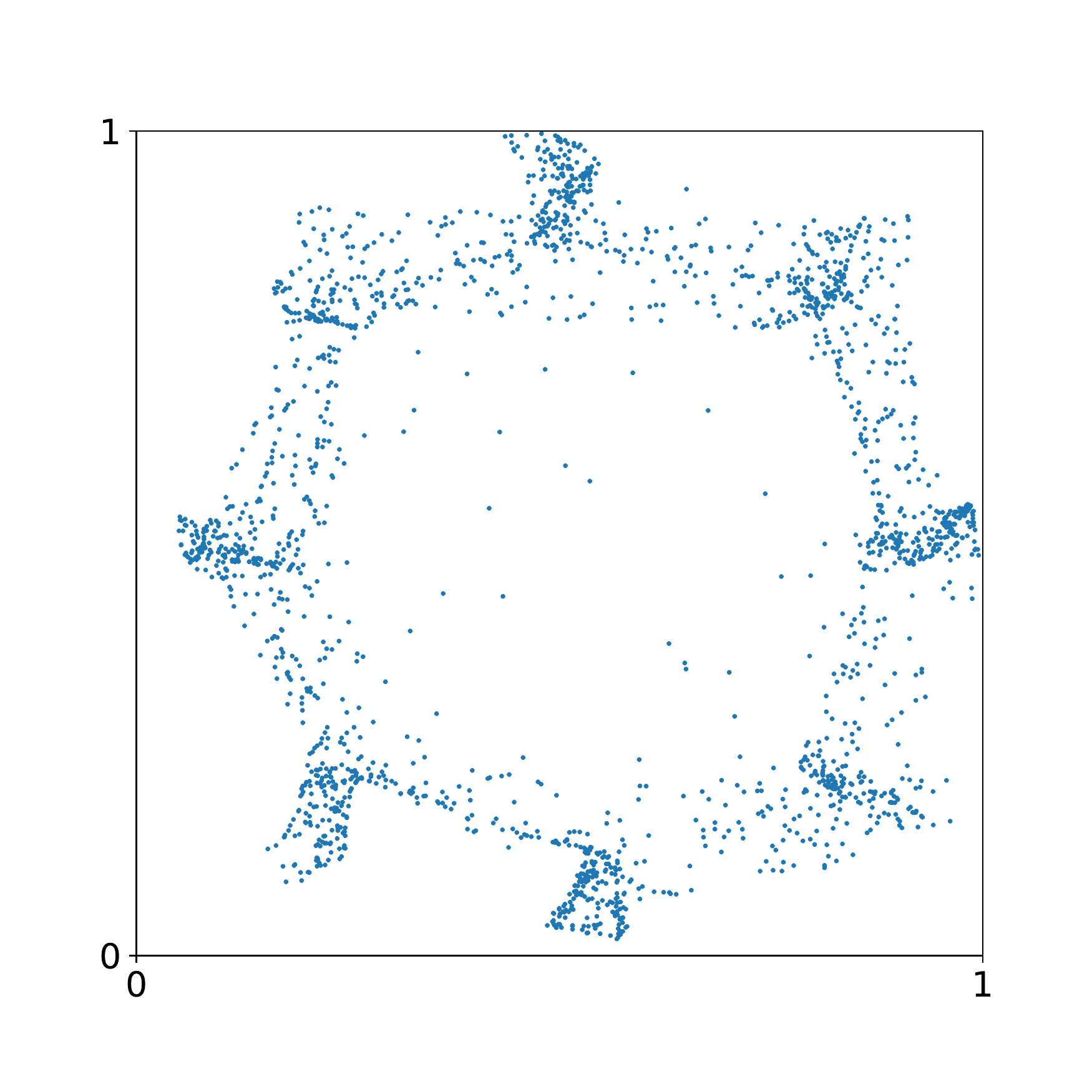} & \includegraphics[width=3cm]{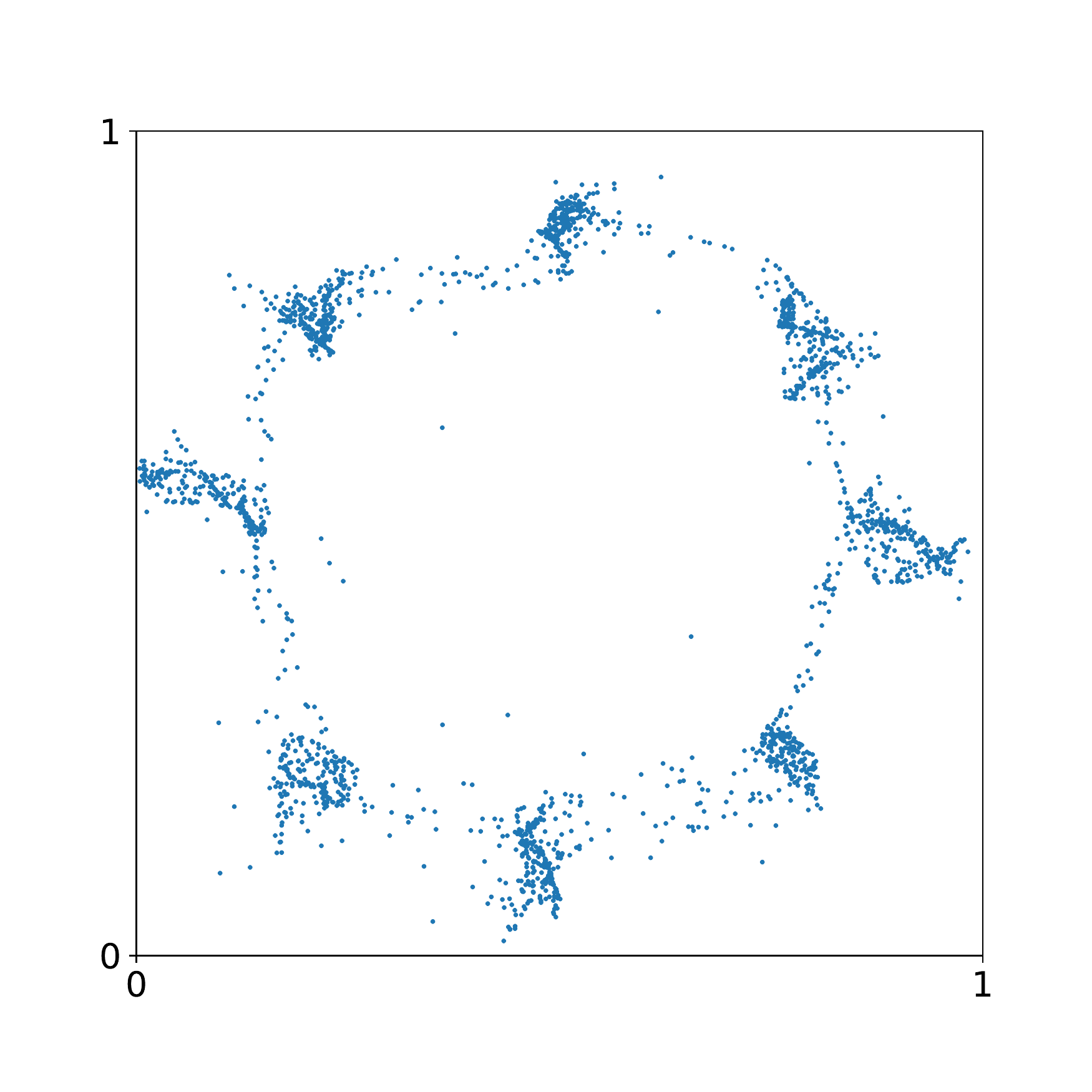} & \includegraphics[width=3cm]{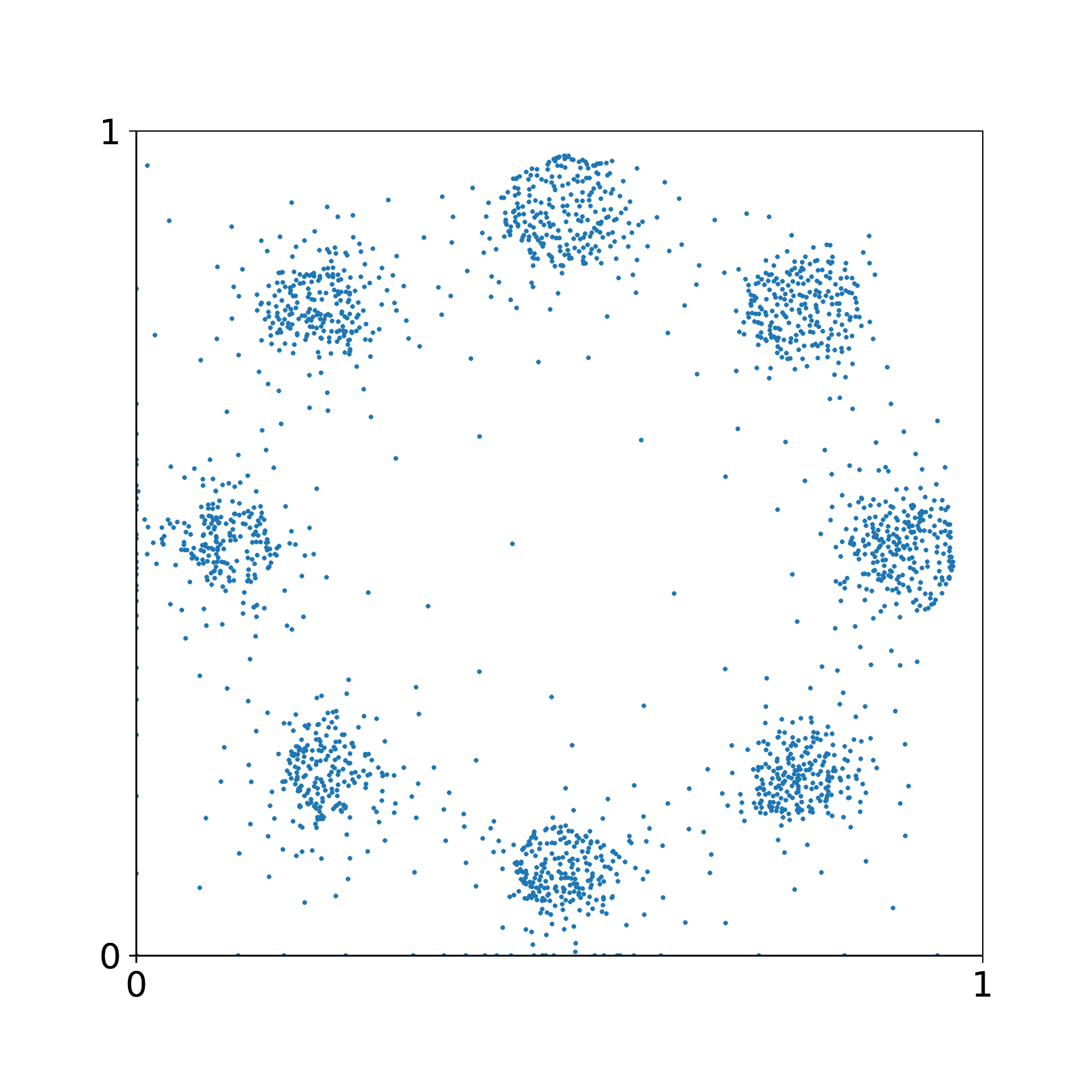}\tabularnewline
(a) Ground truth & (b) WGAN & (c) WGAN-GP & (d) GA-FNTK\tabularnewline
\end{tabular}\caption{\label{fig:2d-toy-dataset-1}Visualization of distribution alignment
and mode collapse on a 2D 8-modal Gaussian mixtures dataset.}
\end{figure}

Note that, with 3-hour time budget, the hyperparameters we obtained
through the grid search are good enough for reproducing the experiments
conducted by \citet{mao2017lsgan} on mode collapse. In the experiments,
the $\mathcal{P}_{\text{gen}}$ of different methods aim to align
a 2D 8-modal Gaussian mixtures in the ground truth. Our results are
shown in Figure \ref{fig:2d-toy-dataset-1}. 

\subsection{Evolution of Images during Training}

\label{subsec:evolution-of-images}
\begin{figure}
\begin{centering}
{\footnotesize{}}%
\begin{tabular}{ccc}
\includegraphics[viewport=0bp 0bp 295bp 193.036bp,clip,width=4.5cm]{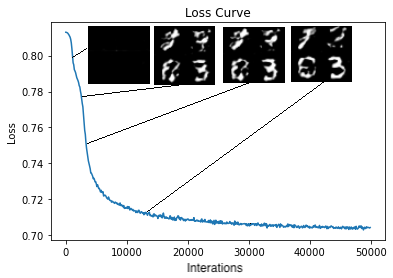} & \includegraphics[viewport=0bp 0bp 295bp 193.036bp,clip,width=4.5cm]{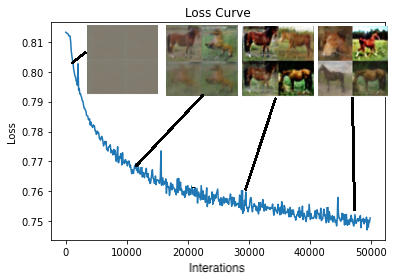} & \includegraphics[viewport=0bp 0bp 300bp 193.036bp,clip,width=4.5cm]{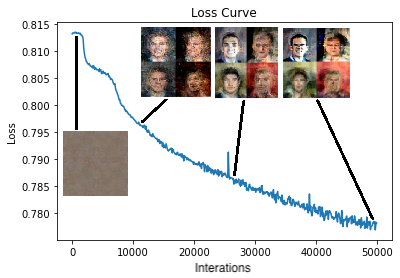}\tabularnewline
{\footnotesize{}(a) MNIST} & {\footnotesize{}(b) CIFAR-10} & {\footnotesize{}(c) CelebA}\tabularnewline
\end{tabular}{\footnotesize\par}
\par\end{centering}
\centering{}\caption{\label{fig:learning-curve-gen}The learning curve of $\mathcal{G}$
in GA-CNTKg and the generated images $\mathcal{G}(\boldsymbol{z})$
at different stages of training given the same input $\boldsymbol{z}$. }
\end{figure}

Figure \ref{fig:learning-curve-gen}
shows the learning curve of the generator in $\mathcal{G}$ in GA-CNTKg
and the relationship between the quality of images output by $\mathcal{G}$
and the number of gradient descent iterations. The results show that
the loss can be minimized even if it is an $f$-divergence, and a
lower loss score implies higher image quality. This is consistent
with the results of GA-CNTK (without a generator) shown in Figure
\ref{fig:learning-curve}. 

\textbf{Source of creativity.} The diversity of our generated data
not only comes from the randomness of an optimization algorithm (e.g.,
initialization of $\boldsymbol{Z}$ or splitting of $\boldsymbol{X}$
into batches, as discussed in Section \ref{subsec:GA-NTK-in-Practice})
but also from the objective in Eq. (\ref{eq:obj:ga-ntk}) itself.
To see this, observe in Figure \ref{fig:learning-curve} that the
images generated at the later stage of training contain recognizable
patterns that change constantly over training time, despite little
change in the loss score. The reason is that, in Eq. (\ref{eq:obj:ga-ntk}),
the $\boldsymbol{Z}^{n}$ is optimized for a \emph{moving} target---any
change of $\boldsymbol{Z}^{n}$ causes $\mathcal{D}$ to be ``retrained''
instantly. The training of the generator $\mathcal{G}$ in Eq. (\ref{eq:obj:bga-ntk-gen})
also shares this nice property. In Figure \ref{fig:learning-curve-gen},
the patterns of a generated image $\mathcal{G}(\boldsymbol{z})$ change
over training time even when the input $\boldsymbol{z}$ is fixed.
However, getting diverse artificial data through this property requires
prolonged training time. In practice, we can simply initialize $\boldsymbol{Z}$
differently to achieve diversity faster.

\section{More Images Generated by GA-CNTK and GA-CNTKg}

\begin{figure}[H]
\centering{}%
\begin{tabular}{c}
(a)\includegraphics[width=7cm]{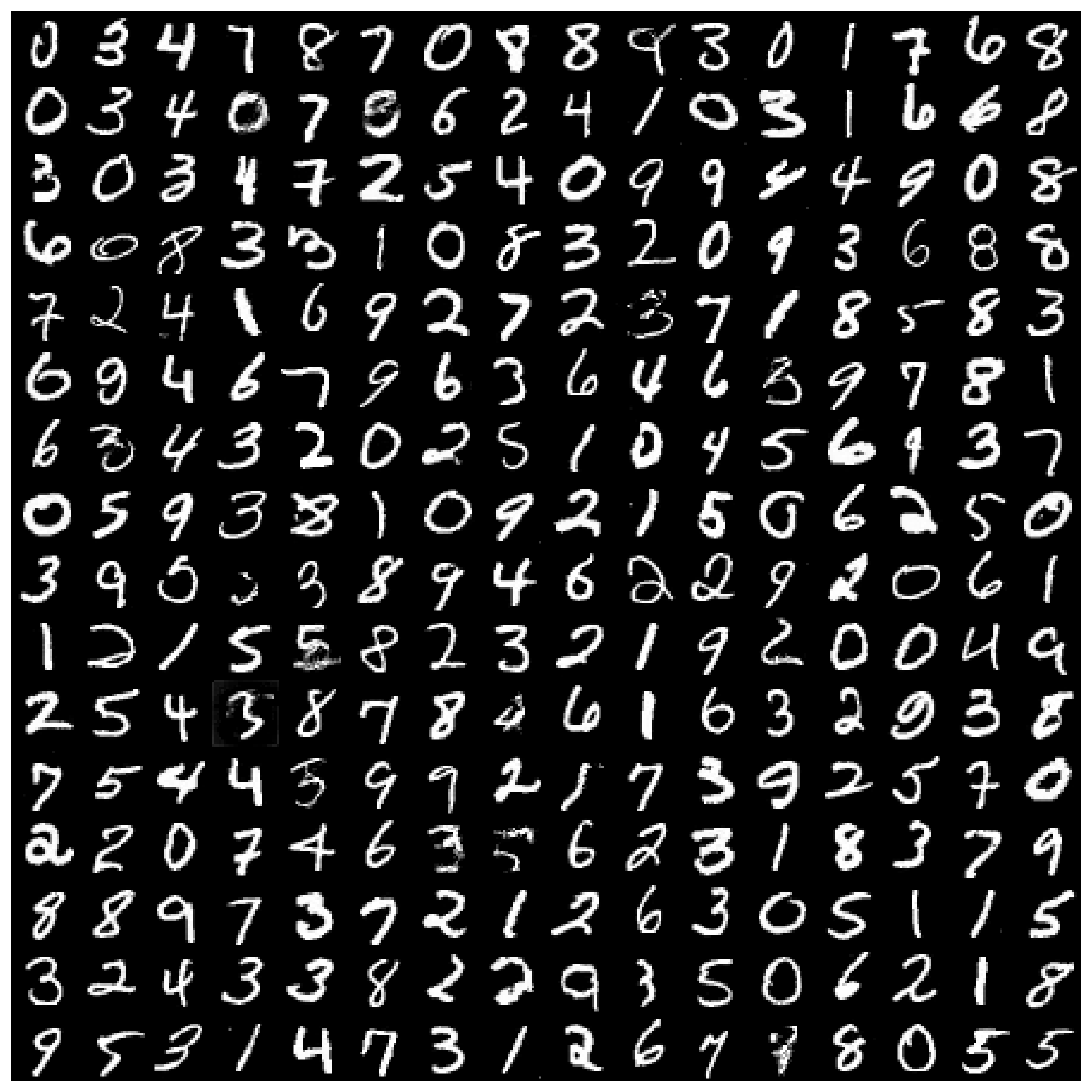}\tabularnewline
(b)\includegraphics[width=7cm]{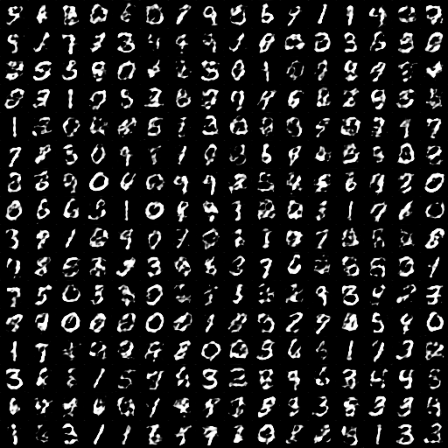}\tabularnewline
\end{tabular}\caption{\label{fig:gacntk-mnist-more}Sample images generated by GA-CNTK (a)
without and (b) with generator on the MNIST dataset of 256 randomly
sampled images.}
\end{figure}

Figures \ref{fig:gacntk-mnist-more}--\ref{fig:gacntkg-celebahq-more}
show more sample images synthesized by GA-CNTK and GA-CNTKg. All these
images are obtained using the settings described in the main paper
and the above.

We can see that the quality of the images synthesized by GA-CNTKg
is worse than that of the images synthesized by GA-CNTK, as discussed
in Section \ref{subsec:Exp-image-quality}. Furthermore, recall from
Table 1 that, without a generator network, the GA-NTK performs better
when the date size increases. However, this is not the case for GA-NTKg
having a generator network. We have resampled training data and rerun
the experiments 5 times with different initial values of $\boldsymbol{Z}^{n}$
but obtained similar results. Therefore, we believe the instability
is due to the sample complexity of the generator network---256 examples
or less are insufficient to train a stable, high-quality generator.
This is evident in Figures 12(b)-15(b) where the generator outputs
unrecognizable images more often. 
\begin{figure}[H]
\centering{}%
\begin{tabular}{c}
(a)\includegraphics[width=9.5cm]{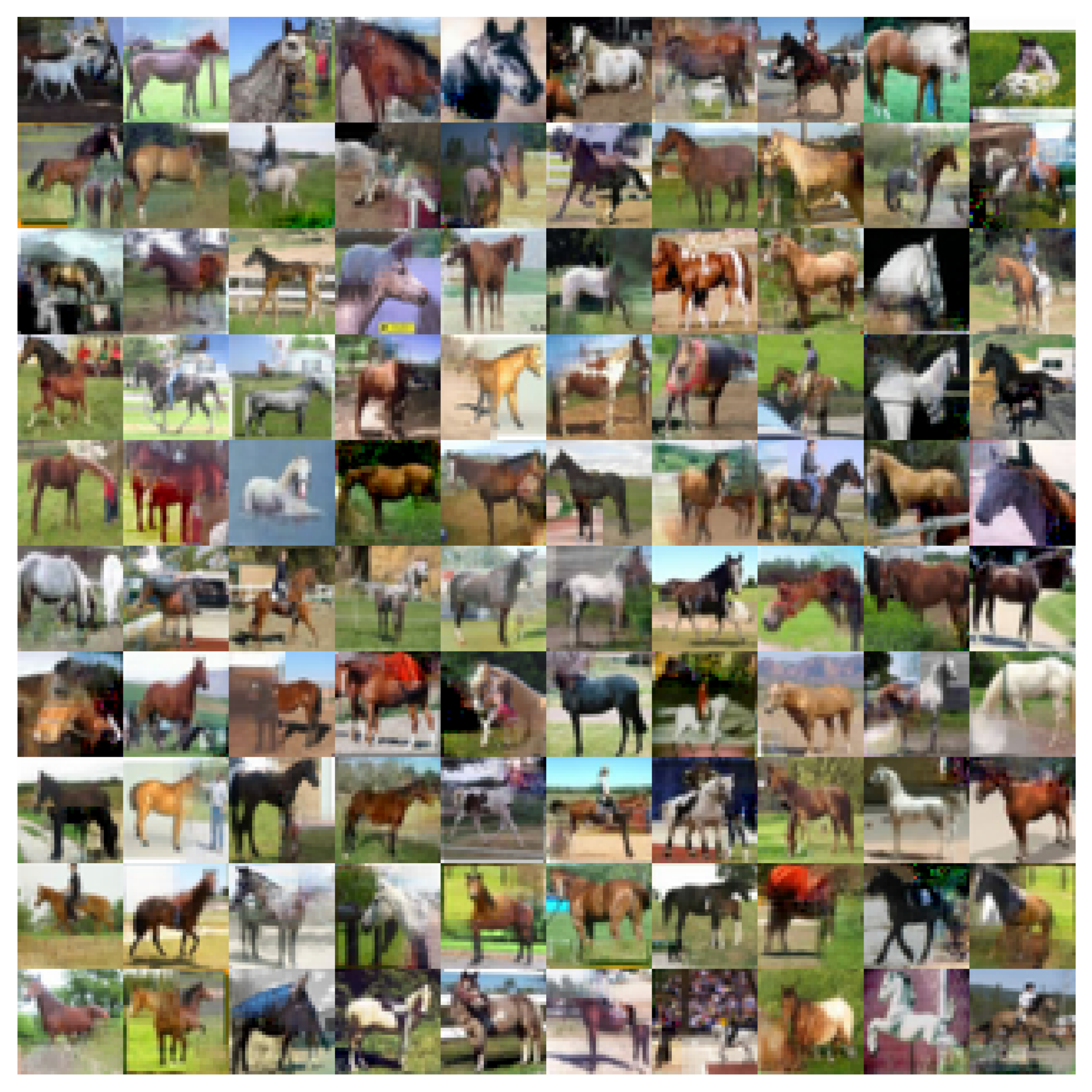}\tabularnewline
(b)\includegraphics[width=10.5cm]{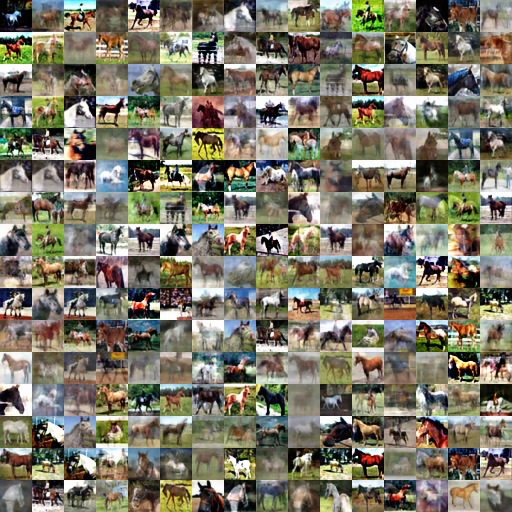}\tabularnewline
\end{tabular}\caption{Sample images generated by GA-CNTK (a)without generator(b)with generator
on the CIFAR-10 dataset of 256 randomly sampled images.}
\end{figure}
\begin{figure}[H]
\centering{}%
\begin{tabular}{c}
(a)\includegraphics[width=9.5cm]{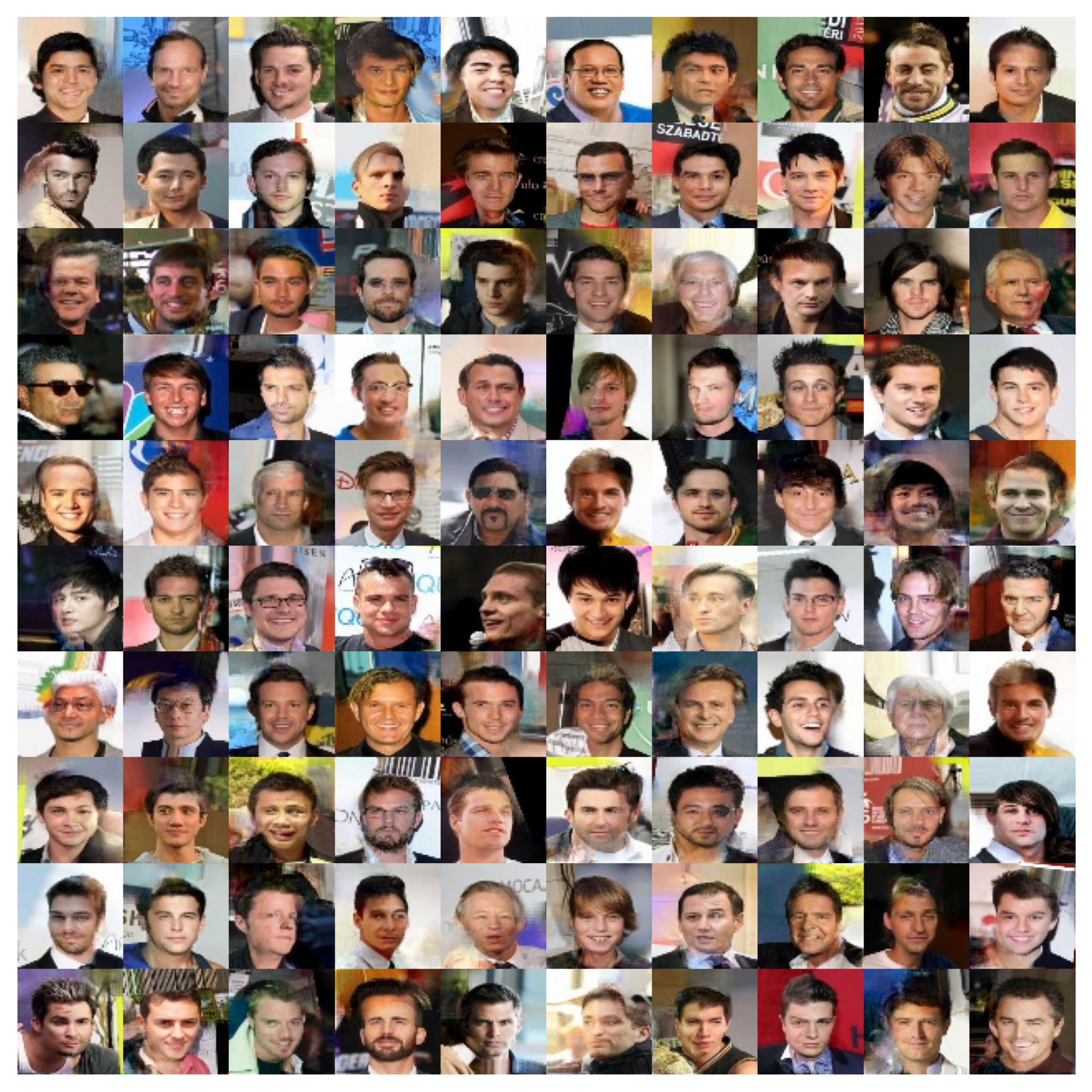}\tabularnewline
(b)\includegraphics[width=10.5cm]{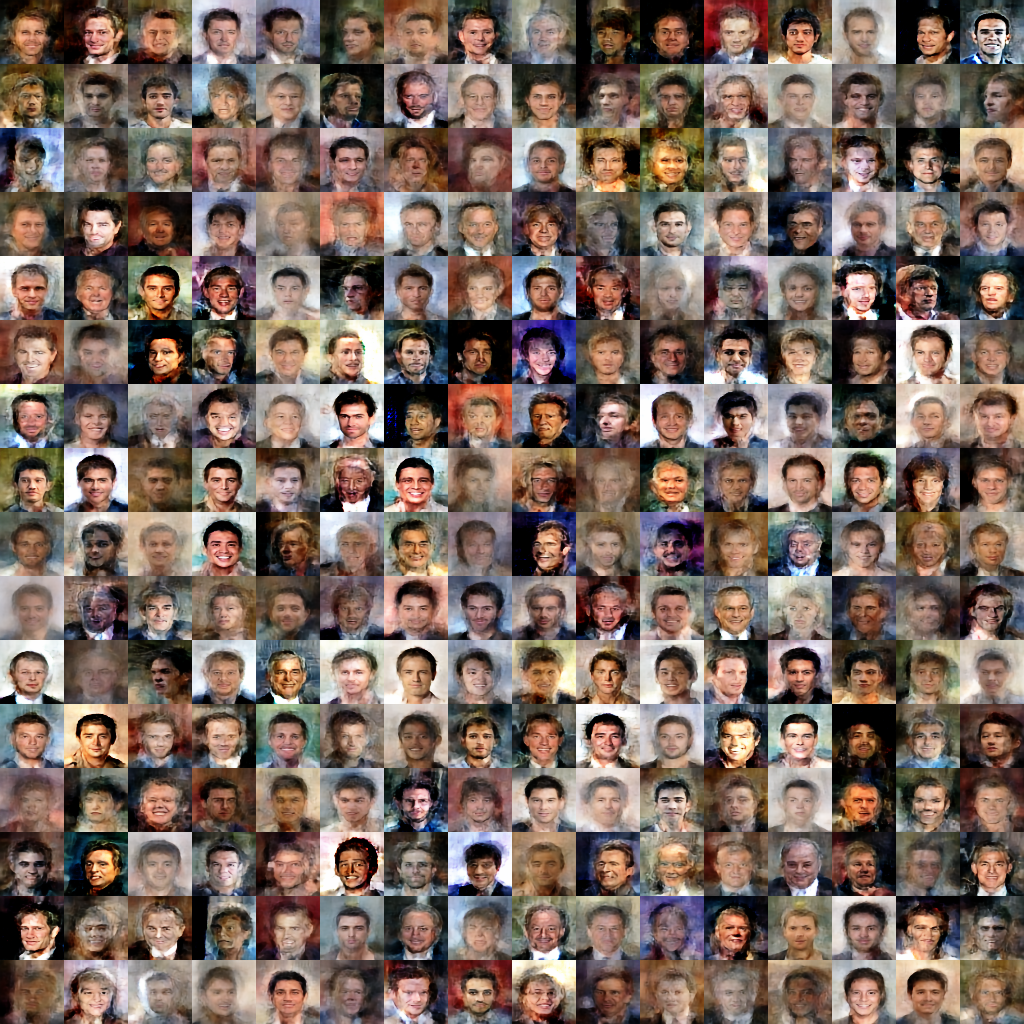}\tabularnewline
\end{tabular}\caption{Sample images generated by GA-CNTK (a) without and (b) with generator
on the CelebA dataset of 256 randomly sampled images.}
\end{figure}

\begin{center}
\begin{figure}[H]
\centering{}%
\begin{tabular}{c}
\includegraphics[width=9.5cm]{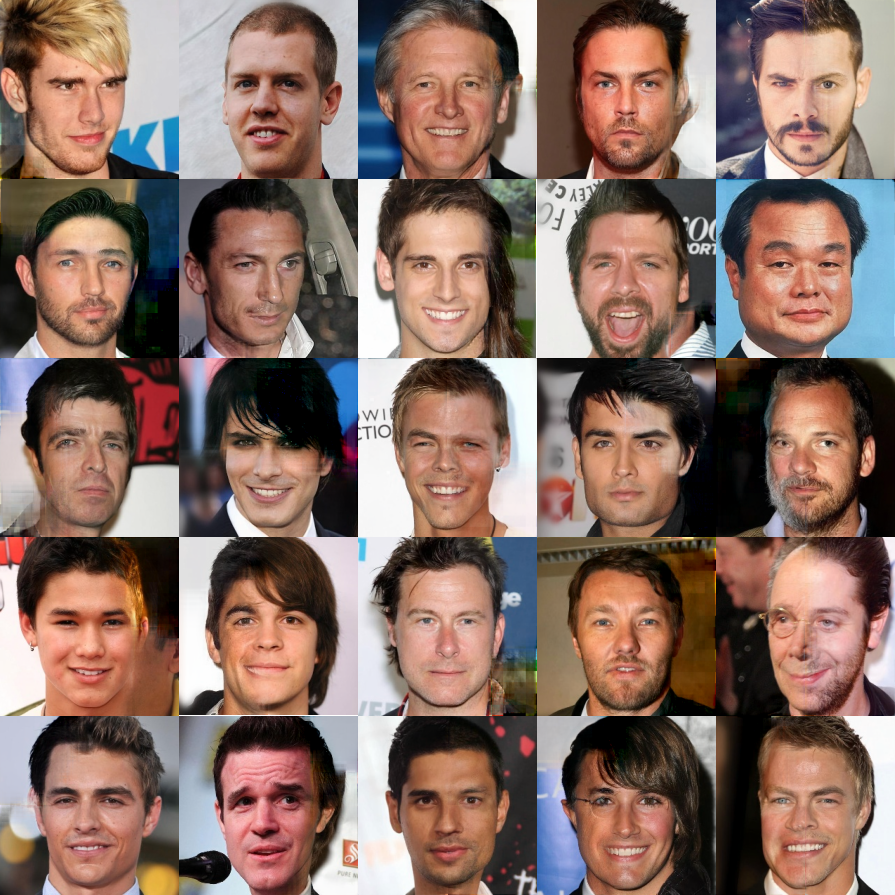}\tabularnewline
\end{tabular}\caption{\label{fig:celeba-hq-multi-res}Sample images generated by multi-resolutional
GA-CNTK on the CelebA-HQ dataset of 256 randomly sampled images.}
\end{figure}
\par\end{center}

\begin{figure}[H]
\noindent \begin{centering}
\includegraphics[viewport=768bp 0bp 2048bp 1280bp,clip,width=9.5cm]{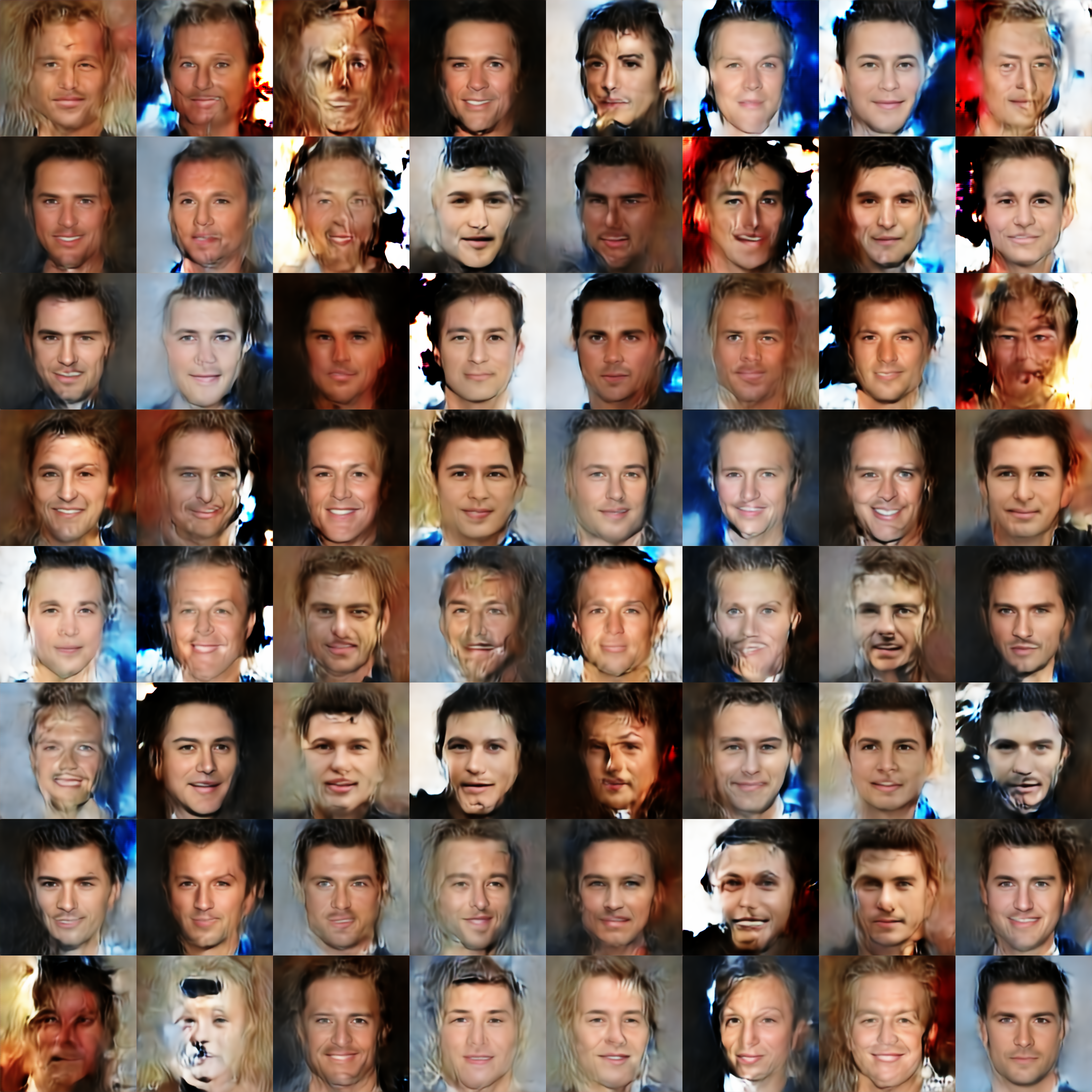}
\par\end{centering}
\noindent \centering{}\caption{\label{fig:gacntkg-celebahq-more}Sample images generated by GA-CNTKg
on the CelebA-HQ dataset of 256 randomly sampled images.}
\end{figure}

\section{Downgrade Images}

\begin{figure}[H]
\centering{}%
\begin{tabular}{cc}
\includegraphics[width=7cm]{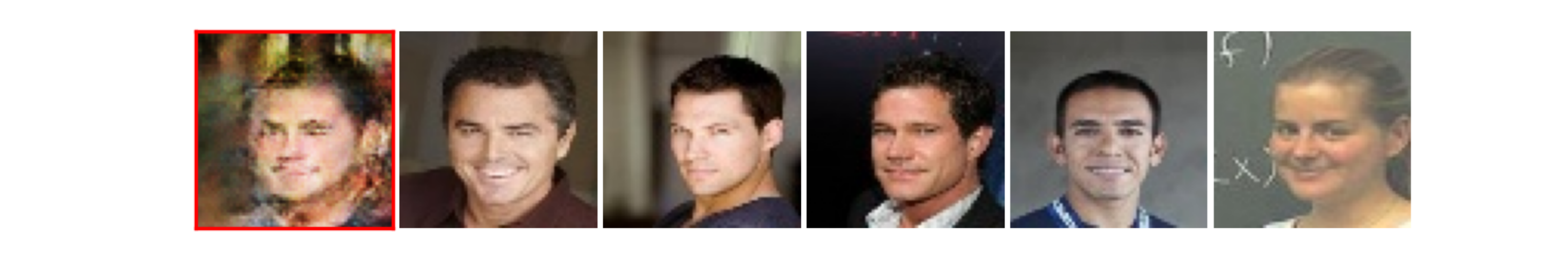} & \includegraphics[width=7cm]{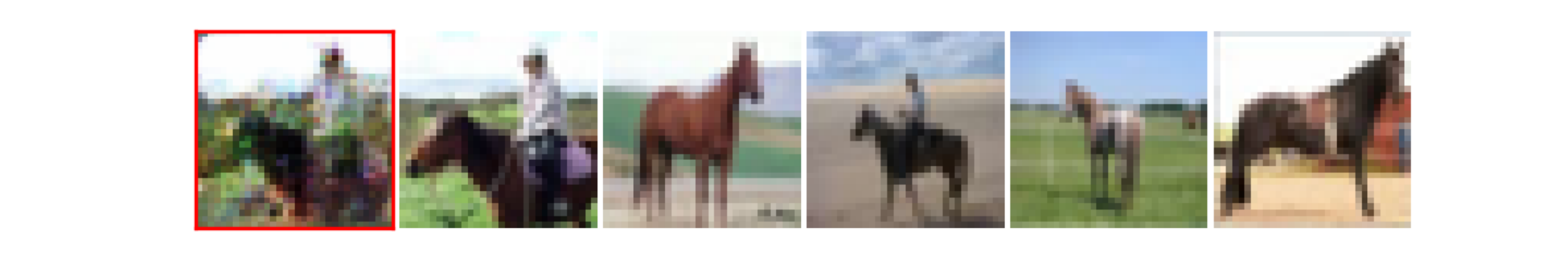}\tabularnewline
\includegraphics[width=7cm]{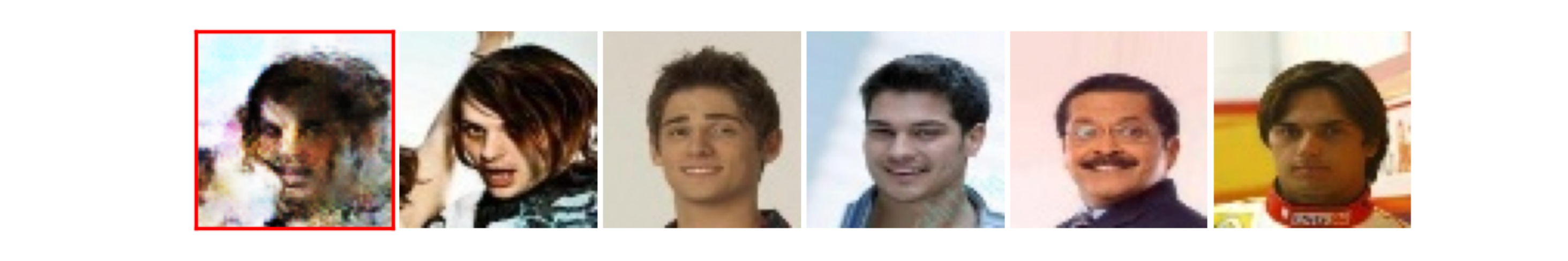} & \includegraphics[width=7cm]{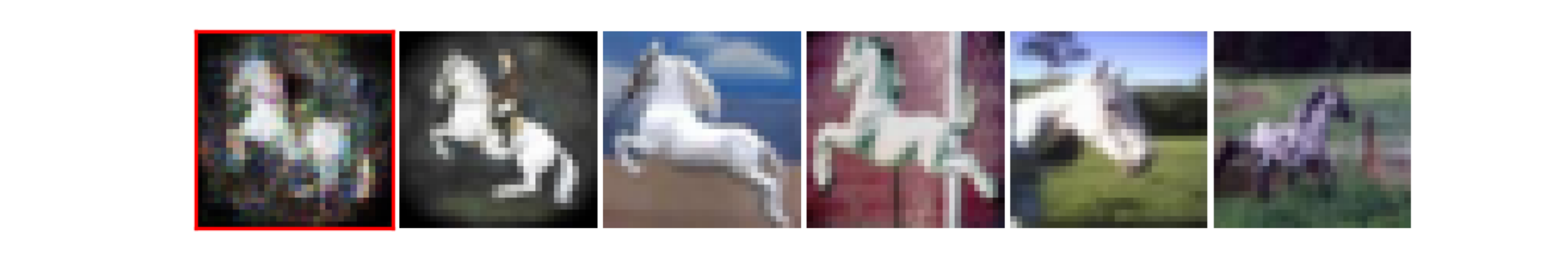}\tabularnewline
\includegraphics[width=7cm]{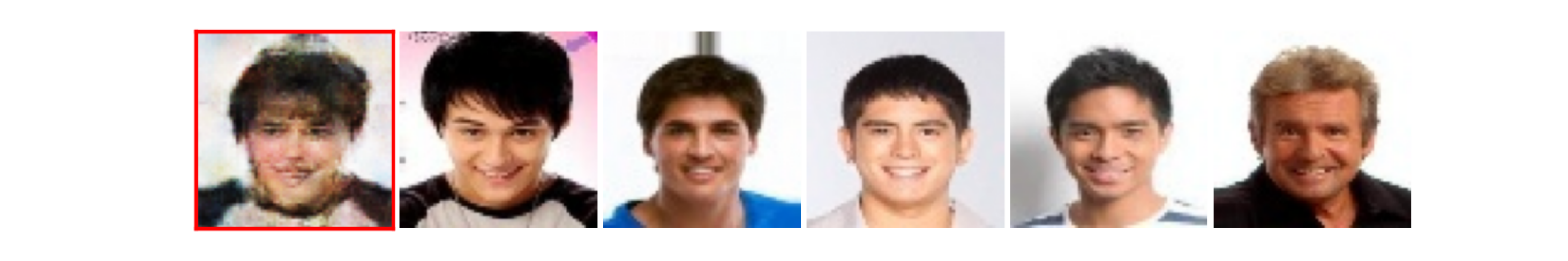} & \includegraphics[width=7cm]{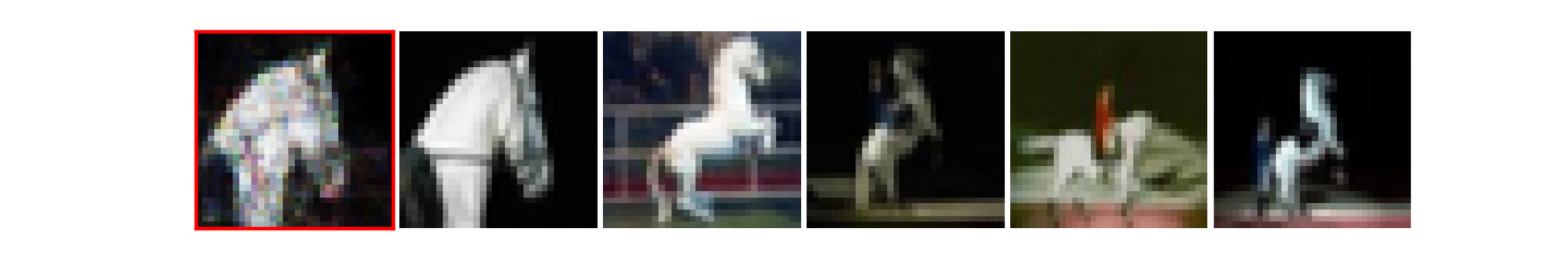}\tabularnewline
\includegraphics[width=7cm]{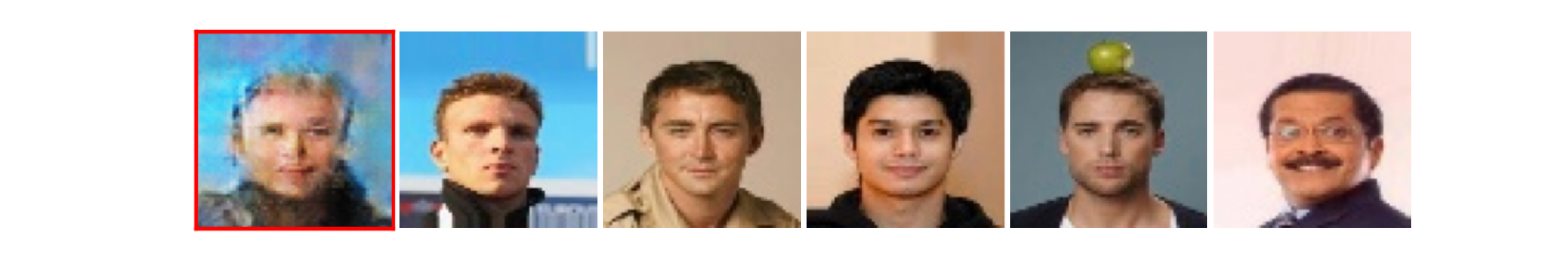} & \includegraphics[width=7cm]{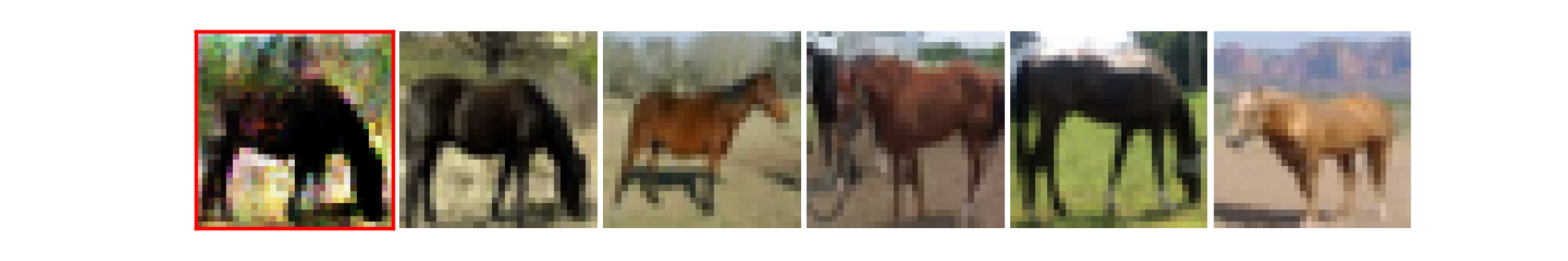}\tabularnewline
\includegraphics[width=7cm]{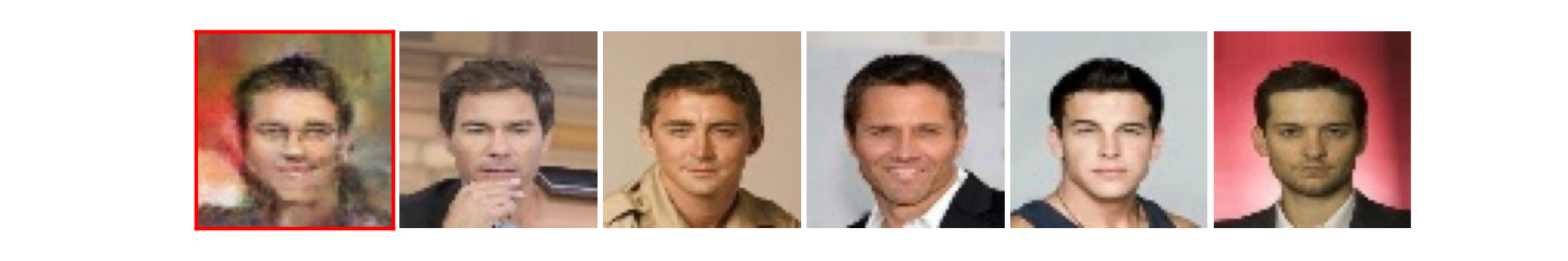} & \includegraphics[width=7cm]{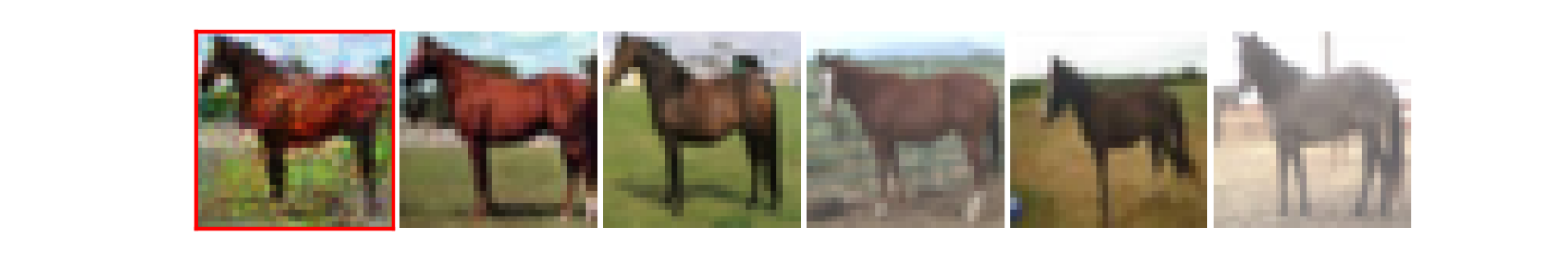}\tabularnewline
\includegraphics[width=7cm]{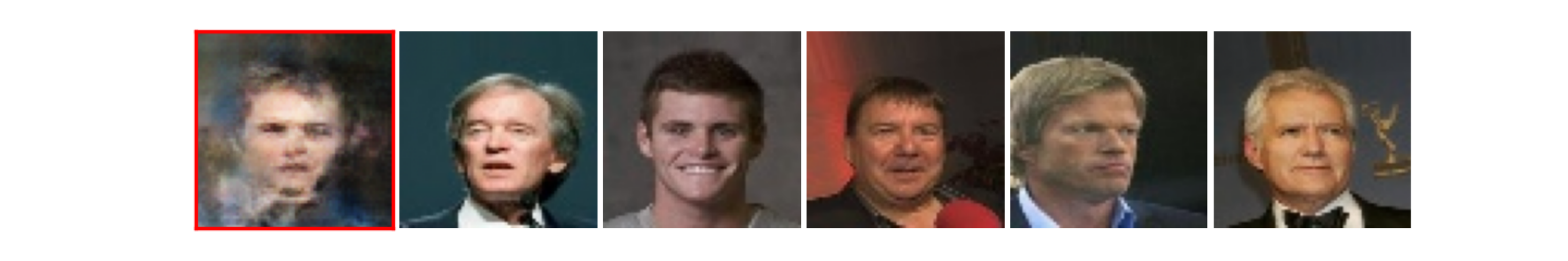} & \includegraphics[width=7cm]{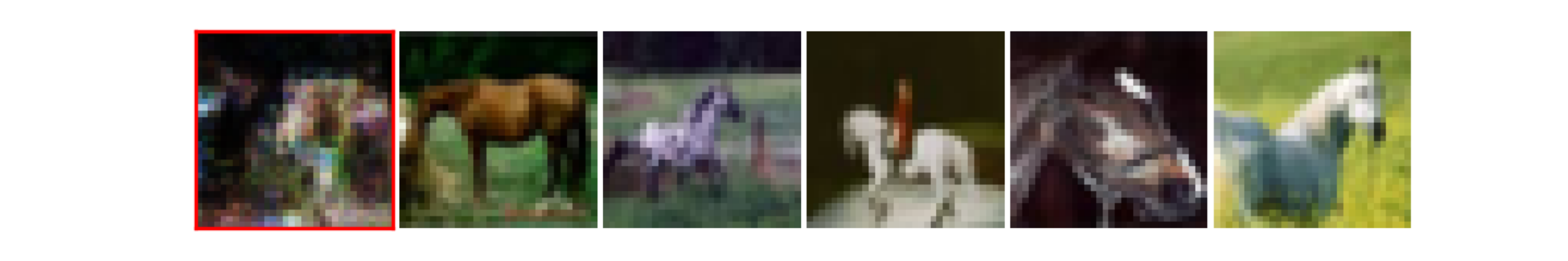}\tabularnewline
\includegraphics[width=7cm]{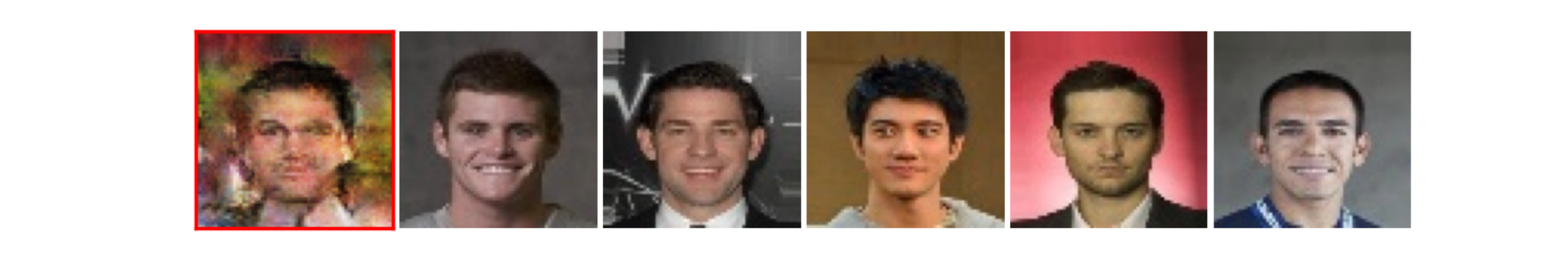} & \includegraphics[width=7cm]{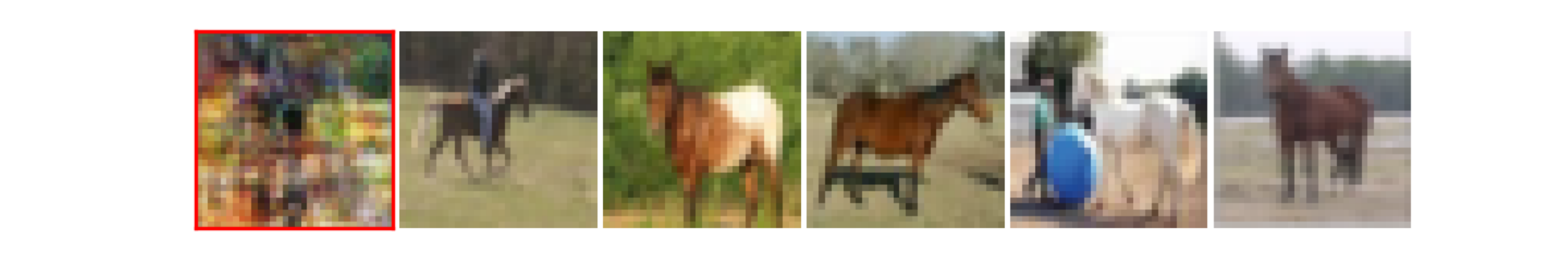}\tabularnewline
\end{tabular}\caption{\label{fig:downgrade-wgangp}Comparison between the images generated
by WGAN-GP trained on 256 images and the nearest neighbors (measured
by SSIM) from the training set. Images with red bounding boxes are
generated images.}
\end{figure}

\begin{figure}[H]
\centering{}%
\begin{tabular}{cc}
\includegraphics[width=7cm]{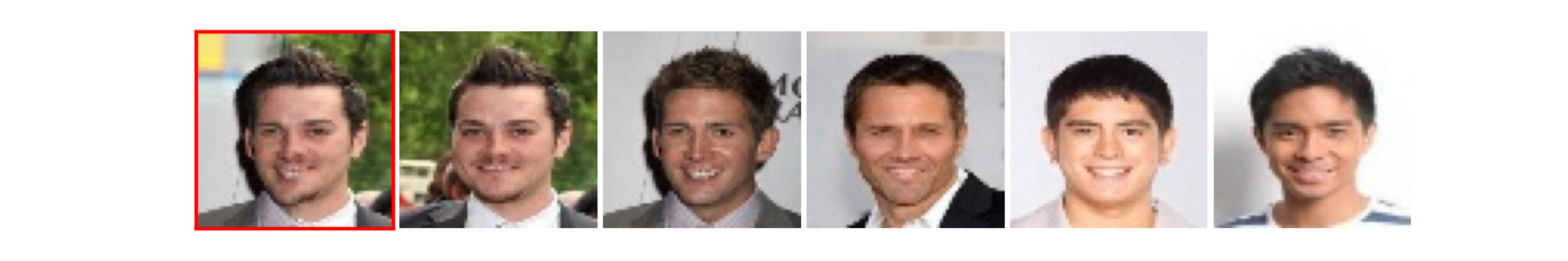} & \includegraphics[width=7cm]{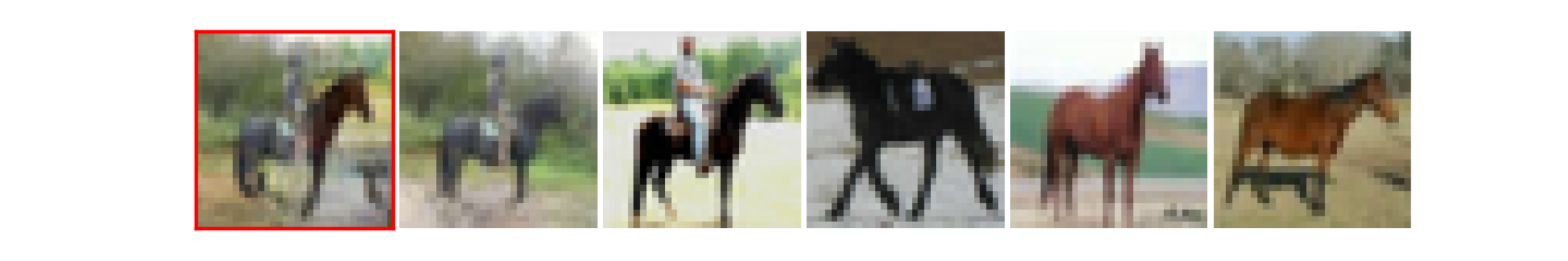}\tabularnewline
\includegraphics[width=7cm]{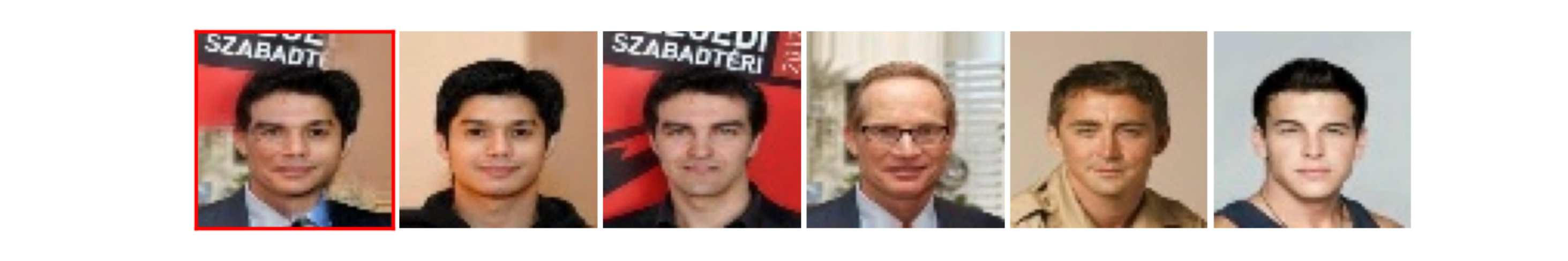} & \includegraphics[width=7cm]{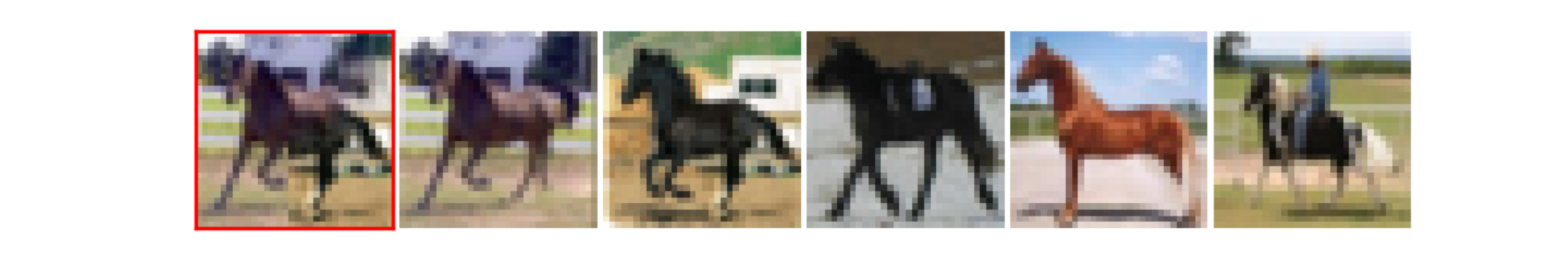}\tabularnewline
\includegraphics[width=7cm]{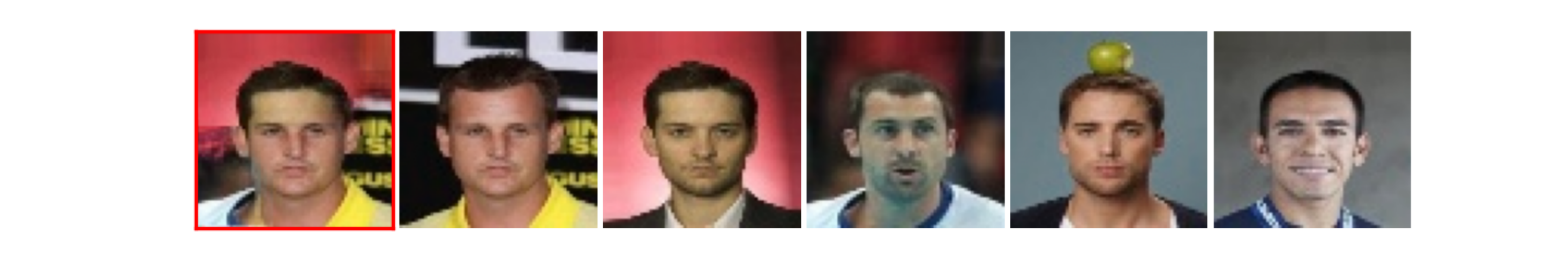} & \includegraphics[width=7cm]{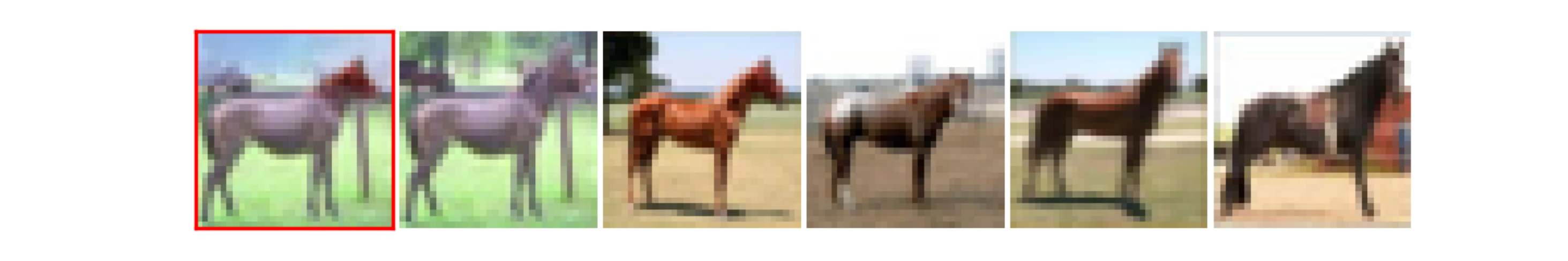}\tabularnewline
\includegraphics[width=7cm]{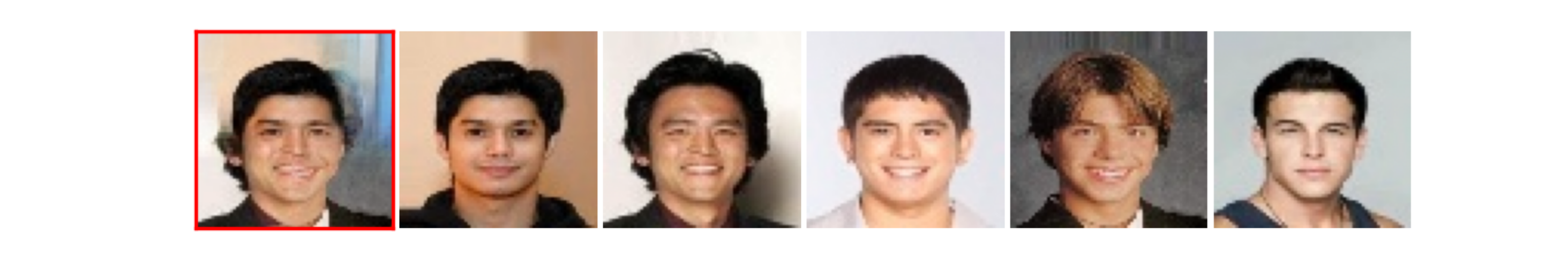} & \includegraphics[width=7cm]{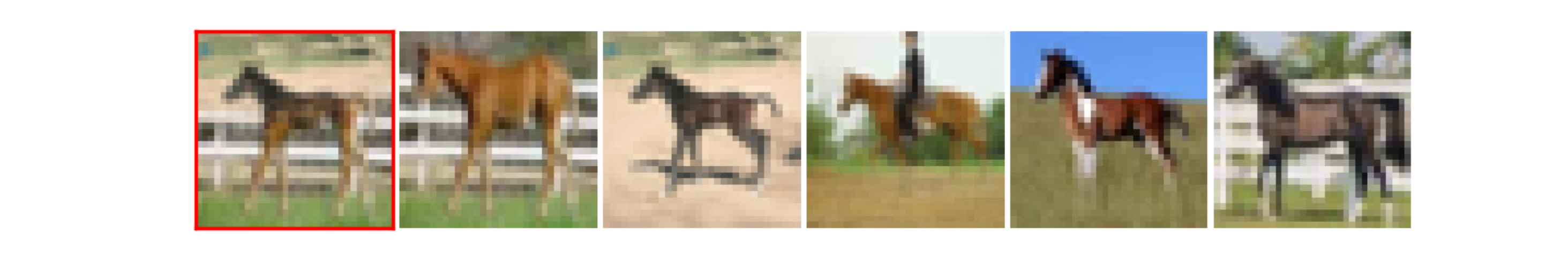}\tabularnewline
\includegraphics[width=7cm]{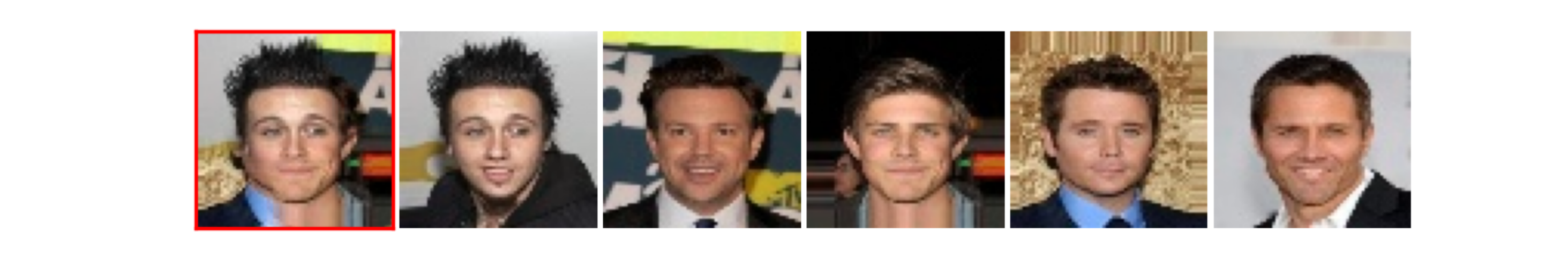} & \includegraphics[width=7cm]{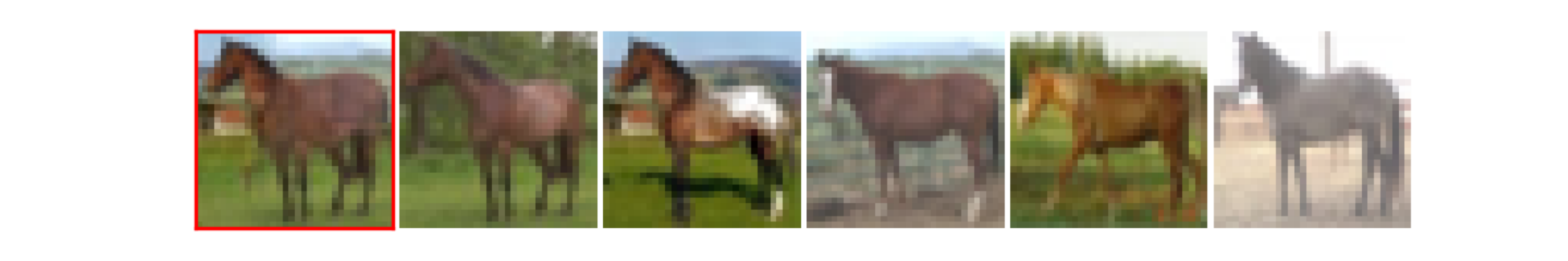}\tabularnewline
\includegraphics[width=7cm]{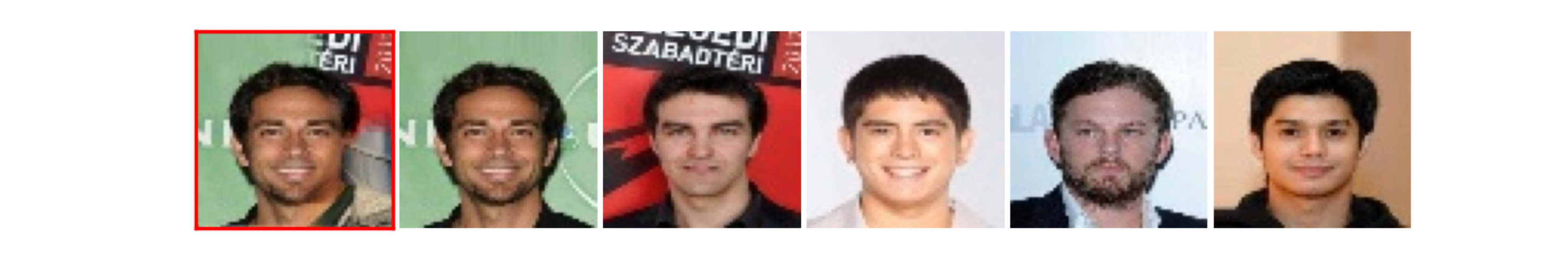} & \includegraphics[width=7cm]{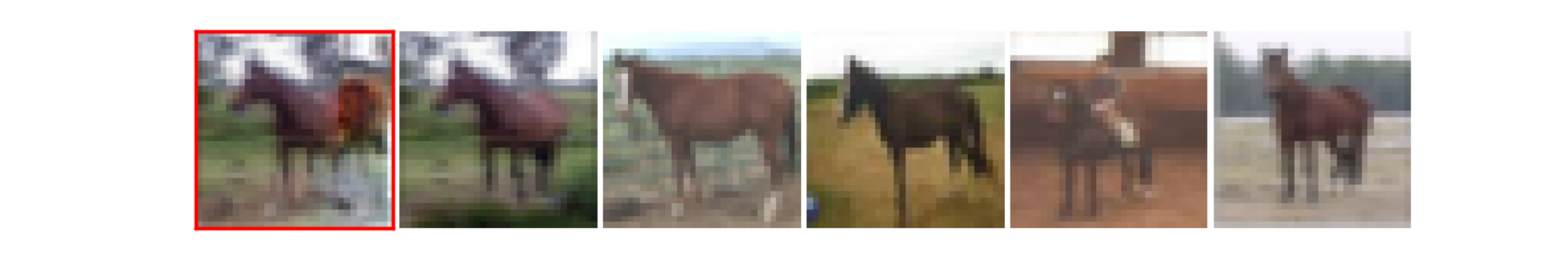}\tabularnewline
\includegraphics[width=7cm]{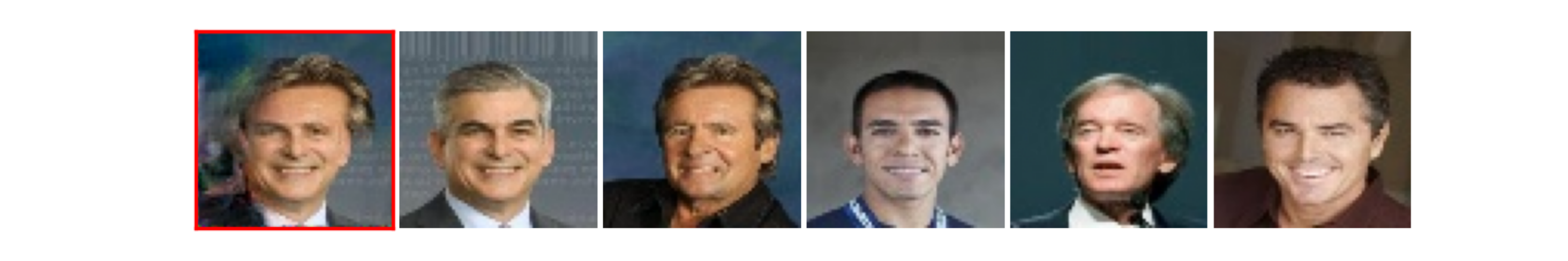} & \includegraphics[width=7cm]{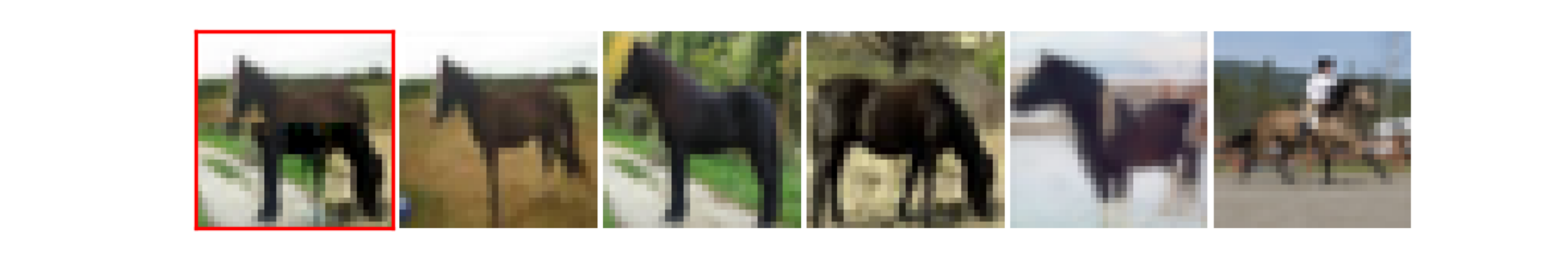}\tabularnewline
\end{tabular}\caption{\label{fig:downgrade-gacntk}Comparison between the images generated
by GA-CNTK trained on 256 images and the nearest neighbors (measured
by SSIM) from the training set. Images with red bounding boxes are
generated images.}
\end{figure}

\begin{figure}[H]
\centering{}%
\begin{tabular}{cc}
\includegraphics[width=7cm]{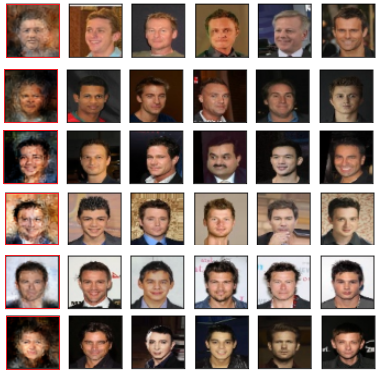} & \includegraphics[width=7cm]{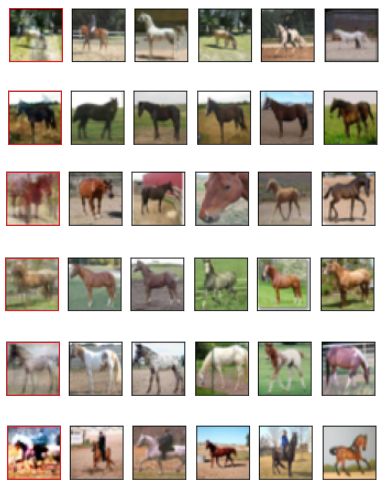}\tabularnewline
\end{tabular}\caption{\label{fig:downgrade-gantk}Comparison between the images generated
by GA-CNTKg trained on 256 images and the nearest neighbors (measured
by SSIM) from the training set. Images with red bounding boxes are
generated images.}
\end{figure}

As discussed in the main paper, we find that, when the size of training
set is small, an image synthesis method may produce downgrade images
that look almost identical to some images in the training set. This
problem is less studied in the literature but important to applications
with limited training data. We investigate this problem by showing
the images from the training set that are the nearest to a generated
image. We use the SSIM \citep{wang2004ssim} as the distance measure.
Figures \ref{fig:downgrade-wgangp}, \ref{fig:downgrade-gacntk},
and \ref{fig:downgrade-gantk} show the results for some randomly
sampled synthesized images. As compared to GANs, both GA-CNTK and
batch-wise GA-CNTK can generate images that look less similar to the
ground-truth images.


\section{Semantics Learned by GA-CNTKg}

\begin{figure}
\begin{centering}
\begin{tabular}{c}
(a) \includegraphics[width=7cm]{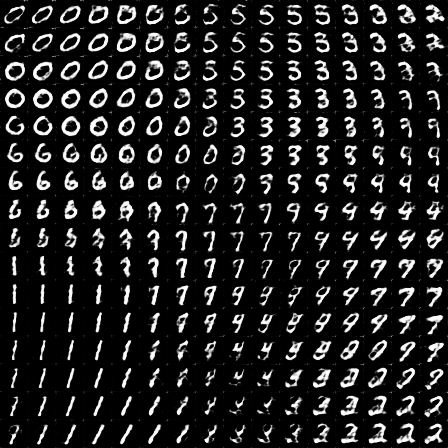}\tabularnewline
\tabularnewline
(b) \includegraphics[width=10.5cm]{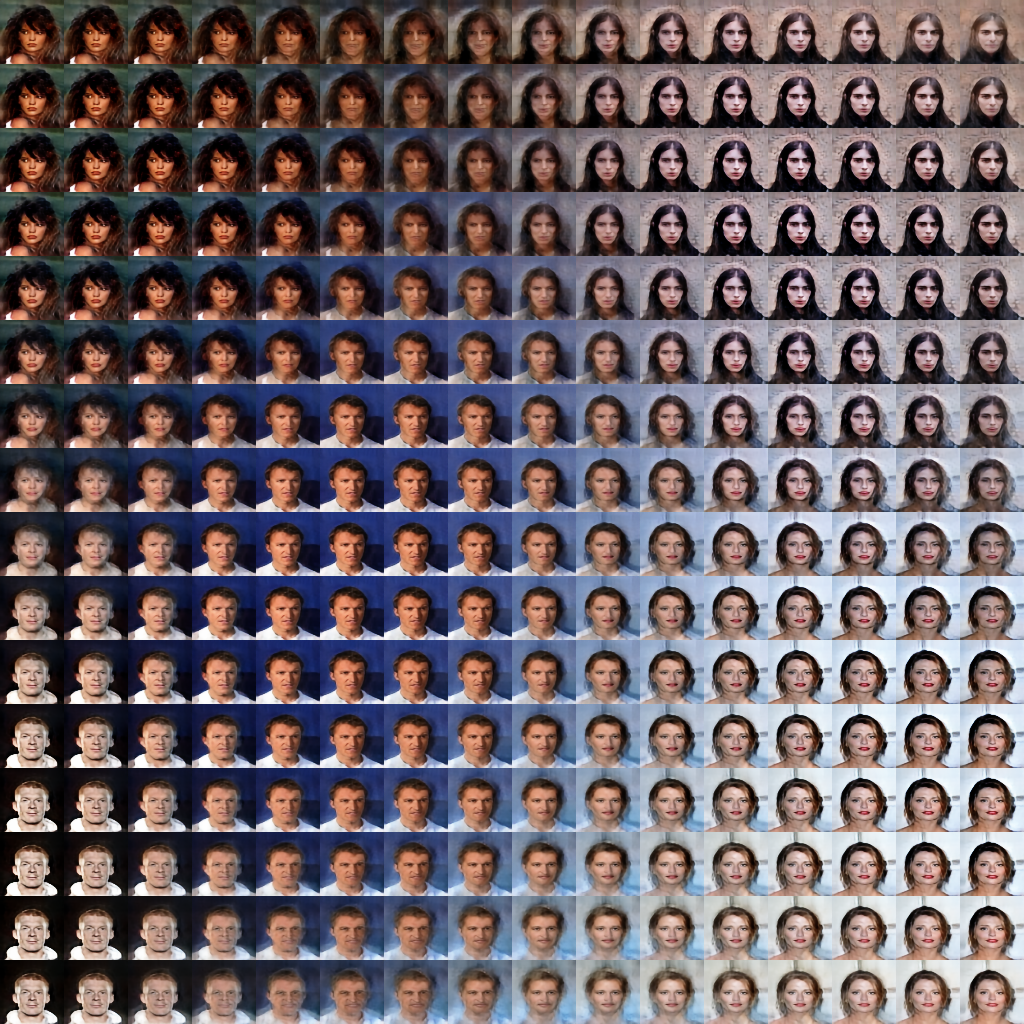}\tabularnewline
\end{tabular}
\par\end{centering}
\caption{\label{fig:manifold}Interpolated images generated by GA-CNTKg, which
is trained on (a) MNIST and (b) CelebA datasets of 256 randomly sampled
examples from all classes. The $\mathcal{G}$ takes 2-dimensional
$\boldsymbol{z}$'s as input. For each dataset, we feed equidistantly
spaced $\boldsymbol{z}$'s along a segment in $\boldsymbol{z}$ space
to $\mathcal{G}$ to get the interpolated images. }

\end{figure}

Here, we investigate whether the features learned by GA-NTK can encode
high-level semantics. We plot ``interpolated'' images output by
the generator $\mathcal{G}$ of GA-CNTKg taking equidistantly spaced
$\boldsymbol{z}$'s along a segment in $\boldsymbol{z}$ space as
the input. For ease of presentation, we consider a 2-dimensional $\boldsymbol{z}$
space and train $\mathcal{G}$ on MNIST and CelebA datasets of 256
examples. Figure \ref{fig:manifold} shows the results, where the
generated patterns transit smoothly across the 2D $\boldsymbol{z}$
space, and neighboring images share similar looks. These similar-looking
images are generated from adjacent but meaningless $\boldsymbol{z}$'s,
suggesting that the learned features encode high-level semantics.

\section{Convergence Speed and Training Time}

In this section, we study the time usage for training GA-NTK variants
and compare it with the training of GANs. We conduct experiments to
investigate the number of iterations and the wall-clock time required
to train different methods on different datasets of 256 randomly sampled
images. We use the batch-wise GA-CNTK and GA-CNTKg and set the batch
size \textbf{$b$} to 64 for all methods. We run the experiments on
a machine with a single NVIDIA Tesla V100 GPU. For DCGAN and LSGAN
whose loss scores do not reflect image quality, we monitor the training
process manually and stop it as long as the generated images contain
recognizable patterns. But these methods do not seem to converge.
For other methods, we use the early-stopping with the patience of
$10000$ steps and delta of $0.05$ to determine convergence. The
results are shown in Table \ref{tab:time}. As we can see, the number
of iterations required by either batch-wise GA-CNTK or GA-CNTKg is
significantly smaller than that used by GANs. This justifies our claims
in Section 1. However, the batch-wise GA-CNTK and GA-CNTKg run fewer
iterations per second than GANs because of the higher computation
cost involved in back-propagating through $\boldsymbol{K}^{b,b}$.
In terms of wall-clock time, the batch-wise GA-CNTK is the fastest
while the GA-CNTKg runs as fast as WGAN-GP. We expect that, with the
continuous optimization of the Neural Tangents library \citep{novak2019neural-tangents}
which our code is based on, the training speed of GA-NTK variants
can be further improved.

\begin{table}
\noindent \begin{centering}
\caption{\label{tab:time}The convergence speed and training time of different
methods on a machine with a single NVIDIA Tesla V100 GPU given different
datasets of 256 randomly sampled images. The GA-CNTK and GA-CNTKg
are batch-wise, and the batch size \textbf{$b$} is set to 64 for
all methods.}
\smallskip{}
\par\end{centering}
\noindent \centering{}{\small{}}%
\begin{tabular}{cc>{\centering}p{1.25cm}>{\centering}p{1.25cm}>{\centering}p{1.25cm}>{\centering}p{1.25cm}>{\centering}p{1.25cm}>{\centering}p{1.25cm}>{\centering}p{1.25cm}}
\toprule 
 & {\small{}Metric} & {\small{}DCGAN} & {\small{}LSGAN} & {\small{}WGAN} & {\small{}WGANGP} & {\small{}SNGAN} & {\small{}GACNTK} & {\small{}GACNTKg}\tabularnewline
\midrule
\midrule 
\multirow{3}{*}{\begin{turn}{90}
{\small{}MNIST}
\end{turn}} & {\small{}Iterations} & {\small{}7400} & {\small{}5100} & {\small{}7000} & {\small{}3400} & {\small{}12800} & {\small{}500} & {\small{}1600}\tabularnewline
\cmidrule{2-9} \cmidrule{3-9} \cmidrule{4-9} \cmidrule{5-9} \cmidrule{6-9} \cmidrule{7-9} \cmidrule{8-9} \cmidrule{9-9} 
 & {\small{}Iter. / sec.} & {\small{}20} & {\small{}19} & {\small{}19} & {\small{}18} & {\small{}18} & {\small{}14} & {\small{}9}\tabularnewline
\cmidrule{2-9} \cmidrule{3-9} \cmidrule{4-9} \cmidrule{5-9} \cmidrule{6-9} \cmidrule{7-9} \cmidrule{8-9} \cmidrule{9-9} 
 & {\small{}Seconds} & {\small{}370} & {\small{}268} & {\small{}368} & {\small{}189} & {\small{}711} & {\small{}35} & {\small{}177}\tabularnewline
\midrule 
\multirow{3}{*}{\begin{turn}{90}
{\small{}CIFAR-10}
\end{turn}} & {\small{}Iterations} & {\small{}N/A} & {\small{}N/A} & {\small{}14000} & {\small{}11100} & {\small{}N/A} & {\small{}600} & {\small{}6200}\tabularnewline
\cmidrule{2-9} \cmidrule{3-9} \cmidrule{4-9} \cmidrule{5-9} \cmidrule{6-9} \cmidrule{7-9} \cmidrule{8-9} \cmidrule{9-9} 
 & {\small{}Iter. / sec.} & {\small{}17} & {\small{}17} & {\small{}16} & {\small{}15} & {\small{}14} & {\small{}13} & {\small{}8}\tabularnewline
\cmidrule{2-9} \cmidrule{3-9} \cmidrule{4-9} \cmidrule{5-9} \cmidrule{6-9} \cmidrule{7-9} \cmidrule{8-9} \cmidrule{9-9} 
 & {\small{}Seconds} & {\small{}N/A} & {\small{}N/A} & {\small{}875} & {\small{}740} & {\small{}N/A} & {\small{}46} & {\small{}775}\tabularnewline
\midrule 
\multirow{3}{*}{\begin{turn}{90}
{\small{}CelebA}
\end{turn}} & {\small{}Iterations} & {\small{}N/A} & {\small{}N/A} & {\small{}18800} & {\small{}11200} & {\small{}N/A} & {\small{}1200} & {\small{}5900}\tabularnewline
\cmidrule{2-9} \cmidrule{3-9} \cmidrule{4-9} \cmidrule{5-9} \cmidrule{6-9} \cmidrule{7-9} \cmidrule{8-9} \cmidrule{9-9} 
 & {\small{}Iter. / sec.} & {\small{}13} & {\small{}12} & {\small{}12} & {\small{}10} & {\small{}9} & {\small{}6} & {\small{}5}\tabularnewline
\cmidrule{2-9} \cmidrule{3-9} \cmidrule{4-9} \cmidrule{5-9} \cmidrule{6-9} \cmidrule{7-9} \cmidrule{8-9} \cmidrule{9-9} 
 & {\small{}Seconds} & {\small{}N/A} & {\small{}N/A} & {\small{}1566} & {\small{}1120} & {\small{}N/A} & {\small{}20} & {\small{}1180}\tabularnewline
\bottomrule
\end{tabular}{\small\par}
\end{table}

\end{document}